\pdfoutput=1

\documentclass[11pt]{article}

\usepackage[final]{acl}

\usepackage{times}
\usepackage{latexsym}

\usepackage[T1]{fontenc}

\usepackage[utf8]{inputenc}

\usepackage{microtype}

\usepackage{dblfloatfix}
\usepackage{graphicx} %
\usepackage{amsmath,amssymb}
\usepackage{multirow}
\usepackage{array}
\usepackage{enumitem}
\usepackage{color, soul}
\usepackage{algorithm}
\usepackage{algpseudocode}
\usepackage{tcolorbox}
\usepackage{float} 
\usepackage{cuted}

\title{An Information-theoretic Approach to Prompt Engineering Without Ground Truth Labels}

\newcommand\blfootnote[1]{%
  \begingroup
  \renewcommand\thefootnote{}\footnote{#1}%
  \addtocounter{footnote}{-1}%
  \endgroup
}

\newcommand\asterix{\textnormal{\normalsize\mbox{*}}}

\author{
Taylor Sorensen\asterix,
Joshua Robinson\asterix,
Christopher Michael Rytting\asterix,\\\textbf{
Alexander Shaw,
Kyle Rogers,
Alexia Delorey, 
Mahmoud Khalil,}\\ \textbf{
Nancy Fulda, 
David Wingate} \\
Computer Science Department, Brigham Young University \\
\texttt{\{tsor13,joshua\_robinson,chrisrytting\}@byu.edu}\\
\texttt{\{nfulda,wingated\}@cs.byu.edu}
}

\begin{document}
\maketitle
\begin{abstract}
Pre-trained language models derive substantial linguistic and factual knowledge from the massive corpora on which they are trained, and prompt engineering seeks to align these models to specific tasks. Unfortunately, existing prompt engineering methods require significant amounts of labeled data, access to model parameters, or both. We introduce a new method for selecting prompt templates \textit{without labeled examples} and \textit{without direct access to the model}. Specifically, over a set of candidate templates, we choose the template that maximizes the mutual information between the input and the corresponding model output. Across 8 datasets representing 7 distinct NLP tasks, we show that when a template has high mutual information, it also has high accuracy on the task. On the largest model, selecting prompts with our method gets 90\% of the way from the average prompt accuracy to the best prompt accuracy and requires no ground truth labels. 
\blfootnote{\mbox{*}Equal Contribution}

\end{abstract}


\section{Introduction}

It is well-known that large pre-trained language models (LMs) learn substantial linguistic \cite{Liu2019, Amrami2018} and factual world knowledge \cite{Petroni2020, Bosselut, Bouraoui, Zuo2018}, achieving state-of-the-art performance on classic NLP tasks like closed-book question-answering, sentiment analysis, and many other tasks \cite{gpt-2, Devlin2019, Raffel2019}. The largest models can do this in a few-shot way--that is, being trained only with generic, semi-supervised objectives and ``taught'' tasks with just instructions and a few examples of the task provided via a natural language ``prompt'' in the context window \cite{Brown2020}. This suggests that pre-training equips them to potentially do many tasks that can be formulated as natural language generation, if only they can be primed in the right way.  

\begin{figure}[t]
\includegraphics[trim=0 0 0 0, clip, width=7.5cm]
{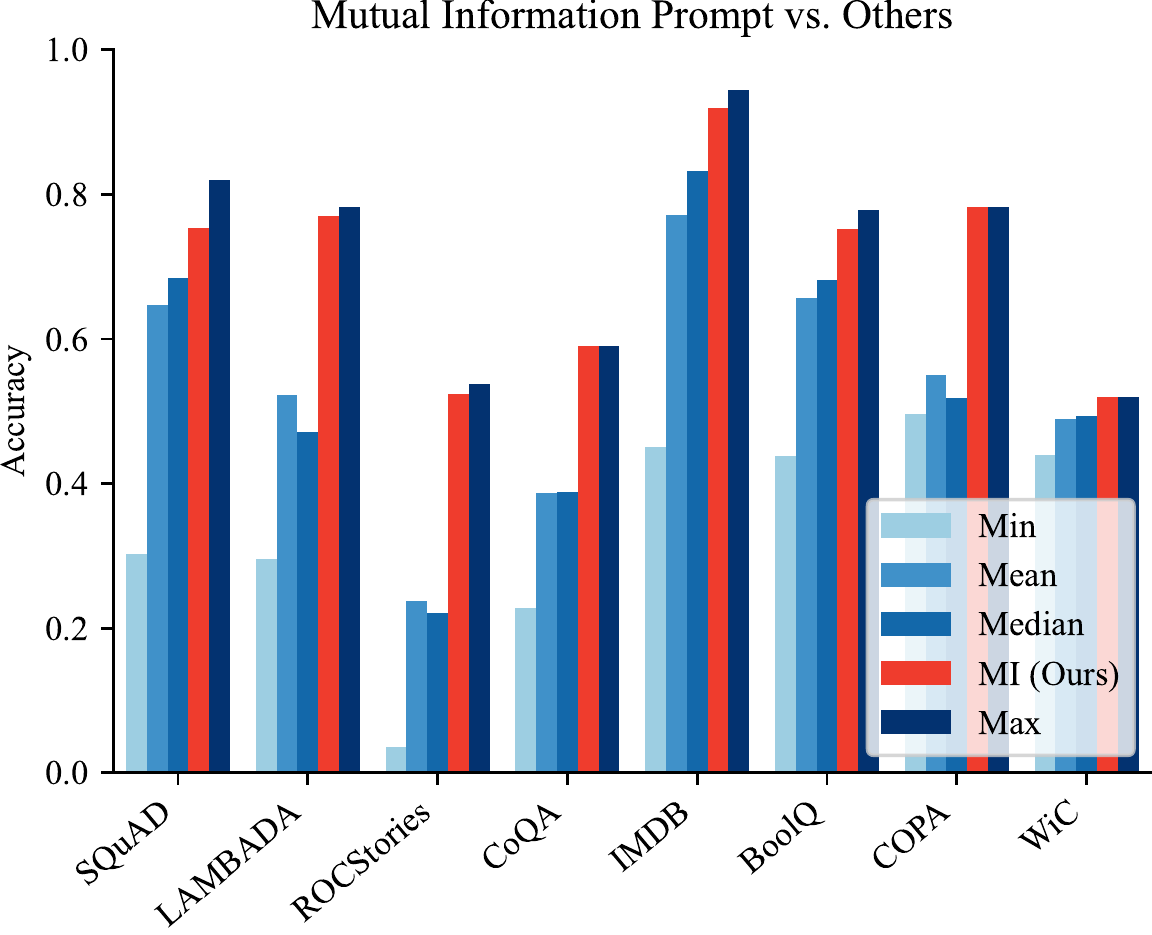}
\centering
\caption{Performance of template selected by our maximum mutual information method (MI) compared to the the worst, mean, median, and best prompt on GPT-3 Davinci (175B). Our method performs at almost oracle levels, without labels or access to model weights.}
\label{fig:cover-plot}
\end{figure}

Such priming is not a trivial task. The few-shot learning breakthrough can give the impression that if the LM is given a sensible prompt, it will ``understand'' what is meant and perform well on the task if it has the capacity. However, LMs can generate substantially different output distributions--and thus text--given two distinct prompts that appear semantically invariant (e.g., alternative orderings, lexical changes like capitalization, and general rephrasing \cite{Zhao2021, Lu2021}). This can lead to surprisingly high variance in performance from prompt to prompt. Clearly, some prompts are better than others for aligning a model to a task. 

Prompt engineering is a nascent field that aims to find aligning prompts \cite{reynolds2021prompt}. While ``prompt'' refers to any language passed to the model via the context window, a \textit{template} refers to a natural language scaffolding filled in with raw data, resulting in a prompt. Thus, prompt engineering includes finding high-quality templates (i.e., those with high test accuracy). Generally, this is done by optimizing for accuracy over a validation set: a template is chosen from a candidate set based on its performance on labeled examples. Such labeled examples can be challenging to procure for some tasks and impossible for others. Some recent methods optimize prompts using backpropagation, which requires access to model weights. In this paper, we propose a new method for selecting prompts by using mutual information, which allows prediction of a prompt's performance without labels or access to model parameters.

Mutual information (MI) is a metric that quantifies the shared information between two random variables (see Section \ref{subsec:MI}). We demonstrate that the mutual information between a prompt and a language model's output can serve as a useful surrogate for the test accuracy of a template. Specifically, for eight popular datasets representing seven classic NLP tasks, we generate a diverse set of 20 templates for each and show that template mutual information and template accuracy are highly correlated. These results are strongest on the largest models we study, for which our method chooses prompts that, on average, get 90\% of the way from mean accuracy to maximum accuracy and even selects the best prompt on three of eight datasets.

This suggests that, across a variety of NLP tasks, mutual information can be used to select one of the best prompts from a set of candidate prompts, even without making use of model weights or ground truth labels. In the following pages, we outline each step of our general method for generating and evaluating templates so that it can easily be ported to any other task. Code is available online.\footnote{\href{https://github.com/BYU-PCCL/information-theoretic-prompts}{github.com/BYU-PCCL/information-theoretic-prompts}}

\section{Related Work}

The promise of language models and the challenge of aligning them has given rise to the field of prompt engineering, which seeks to construct the best prompt given a task and a language model \cite{Liu2021pre}. The best performance on prompt engineering is often achieved using backpropagation in continuous prompt embedding space \cite{Lester2021, Li2021, Gu2021, Liu2021,Zhang2021} in contrast to generating a discrete set of prompts by hand and testing them. While optimizing in continuous prompt space via backprop allows for similar performance to model-tuning (at least at higher model sizes) \cite{Lester2021}, not all models are publicly available. Thus, these methods are only feasible for those who have direct access to the model and can perform backprop on it. Prompts optimized in continuous space are also not interpretable in natural language, making it harder to transfer insights from prompts that work well for one task to another task. Additionally, these methods require labeled examples, while ours does not.

Other selection protocols not based on gradient descent can include cross-validation or minimum description length, as in \cite{Perez2021}. These methods yield prompts that perform marginally better than average in terms of test accuracy. 

Mutual information has been used in n-gram clustering, part-of-speech tagging, probing classifiers, and LM training objective reframing \cite{brown-etal-1992-class, stratos2019mutual, Voita2020, kong2019mutual}. Ours is the first work of which we are aware to apply MI to prompt engineering.
\cite{Lu2021} make use of entropy statistics to determine performant orderings for few-shot examples in prompts. Our work is focused on selecting high quality templates with no special focus on example ordering or need for multiple examples to order (the few-shot case). Our method uses no artificial ``probing set,'' making our prompt selection much cheaper, and we also explore open-ended tasks. While the GlobalE and LocalE statistics they use are similar (and in the case of LocalE identical) to the two parts of our MI calculation (see \ref{subsec:MI}), we use the two statistics jointly and choose prompts by minimizing, rather than maximizing, LocalE.


\section{Methods}
\label{sec:methods}
\begin{figure*}[t]
\includegraphics[trim=0 0 0 0, clip, width=16cm]{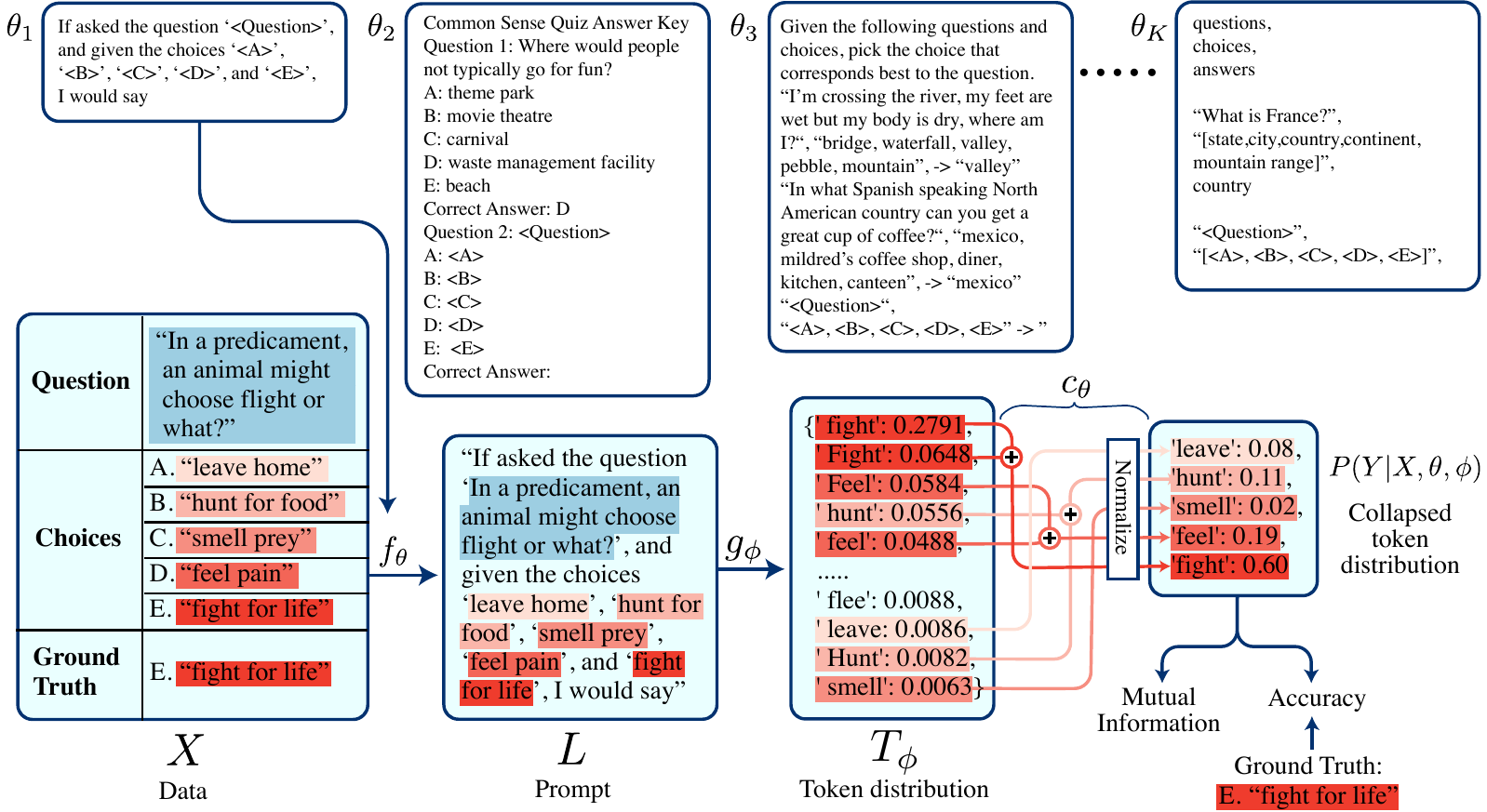}
\centering
\caption{We choose $\theta \in \{\theta_i\}_{i=1}^K$ and templatize a sampled instance from the dataset $X$. We pass this prompt through the language model via $g_\phi$, yielding a probability distribution over the model's tokens $T_\phi$. The collapsing function $c_\theta$ sums the weight given to each token corresponding to each possible answer $y\in Y$  and normalizes, giving a probability distribution $P(Y|\mathbf{x}_i)$, which we can use to estimate mutual information or obtain a guess for $y_i$.}
\label{fig:templatizing-diagram}
\end{figure*}


At the most abstract, our method is as follows (see Appendix \ref{subsec:promptengineeringprocess} for a more thorough description):

\begin{tcolorbox}[right=0.2in,
                  left=0in,
                  top=0.1in,
                  bottom=0.1in,
                  ]
                  
    \begin{enumerate}[itemsep=0pt]
    \item Generate a set of $K$ prompt templatizing functions.
    \item Playground a couple of examples to ensure that templates give roughly expected output.
    \item Estimate mutual information for each template given a set of inputs $\mathbf{x}_1, \mathbf{x}_2, ... ,\mathbf{x}_N$ where $\mathbf{x}_i \sim X , \forall i$.
    \item Choose template(s) based on mutual information and perform inference.
    \end{enumerate}
\end{tcolorbox}




We find it useful to unify all the tasks we study within a single framework, which we describe in Section \ref{subsec:otr}. We also justify our use of mutual information as a surrogate for prompt quality and specify how we estimate it in Section \ref{subsec:MI}.


\subsection{Task Definition}
\label{subsec:otr}

In order to demonstrate our method's widespread applicability and general effectiveness, we validate it across many datasets and tasks. This requires us to estimate MI and accuracy, and this is most straightforward in the case where, given a context, a language model produces just one probability distribution $P(\mathbf{t}_n|\text{context}=\mathbf{t}_1, \mathbf{t}_2,...,\mathbf{t}_{n-1})$. This is in contrast to other experimental setups that use multi-token sampling methods (e.g., beam search), although our method is easily tractable in such setups.\footnote{The only difference: For each considered answer, simply calculate its unnormalized probability by multiplying the probabilities of the decisions taken at each branch in the sequence of tokens, then normalize the resulting probability scores.} Any NLP task is tractable in this framework so long as the output space consists of a set of options that each start with a unique token. In this case, the language model can ``give'' an answer by assigning probability to tokens that begin giving each of these answers
(invariant to lexical variation like capitalization and leading/trailing spaces). While, for open-ended tasks, this method might artificially inflate accuracy if the model starts to give a wrong answer that happens to start with the same token as the correct one, we find that this difference is small and does not affect our results.\footnote{Our open-ended datasets are SQuAD, LAMBADA, and ROCStories, and none of these seemed more likely than ROCStories to exhibit this issue. We reran our experiment on ROCStories by sampling with temperature 0 until reaching a space, and only counted responses as accurate if they exactly matched the corresponding ground truth labels. Results were virtually unchanged: accuracy decreased by only 0.03 on average, and the correlation between mutual information and test accuracy increased by 0.04, from 0.68 to 0.72.}
Irrelevant tokens (with which none of the desired answers begin) are ignored, and the resulting collapsed probabilities are normalized. We term this approach \textit{One-token Response} (OTR). Although our method isn't limited to OTR tasks, we choose tasks that can be cast as OTR tasks for simplicity and to reduce computational expense. Many NLP tasks fit within this framework, although a few do not (e.g., machine translation and summarization). This basic approach is in common use \cite{Brown2020}, but we formalize it for clarity below.

Generally, the OTR framework casts a natural language task as a classification problem with raw data input $\mathbf{x}_i \in X$ and output $P(Y|\mathbf{x}_i)$, a probability distribution over targets. In order to use a language model ${\phi}$ for this task, a templatizing function $f_\theta: X \rightarrow L$ is needed to map raw data into natural language prompts. $g_\phi: L \rightarrow T_\phi$ maps prompts to a probability distribution over $T_\phi$, the token set represented by the model tokenizer. Finally, a collapsing function $c_{\theta}: T_\phi \rightarrow P(Y|\mathbf{x}, \theta, \phi)$ 
(see Appendix \ref{subsec:promptengineeringprocess}) yields an estimate of $P(Y|X)$: 

\vspace{-15pt}

\begin{equation}
\label{equation:otr}
P(Y|\mathbf{x}, \theta, \phi) = c_\theta(g_\phi(f_\theta(\mathbf{x}))), \mathbf{x} \in X
\end{equation} We also refer to $P(Y|\mathbf{x}, \theta, \phi)$ as $P(Y|f_{\theta}(\mathbf{x}))$.

The above pipeline can be specified in many ways using different $\theta$ and $\phi$ (see Figure \ref{fig:templatizing-diagram}), which will result in different accuracies. Our ultimate aim is to select the best $\theta$ given $\phi$. Whereas past prompt engineering methods rely on scores calculated by comparing model answers and ground truth, our method selects $\theta$ by maximizing mutual information, which requires no ground truth labels.

\subsection{Mutual Information}
\label{subsec:MI}

Mutual information is a measure of the amount of shared information between two random variables \cite{Cover2006}; in other words, it is the reduction in entropy that is observed in one random variable when the other random variable is known.

We expect MI to serve as a good criterion for comparing prompts. Previous work has shown that large networks trained with cross-entropy loss are calibrated (e.g., a 60\% confidence corresponds to a 60\% chance of the model being correct) when in the early-stopped ($\sim 1$ epoch) regime \cite{ji2021earlystopped}, but become miscalibrated in the overfit regime \cite{DBLP:journals/corr/abs-2009-08092}. According to \cite{Brown2020}, GPT-3 was trained for a different number of epochs on each corpus in its training data. We calculate it was trained for an average of 1.57 epochs, so we have reason to believe that GPT-3 is generally well-calibrated. Thus, we postulate that a prompt that elicits a very confident response (high MI) from the language model is more likely than a less confident prompt to score well.

We denote the mutual information between random variables $X$ and $Y$ as $I(X;Y)$ and the entropy of $X$ as $H(X)=-\int_{\mathbf{x}\in X} P(\mathbf{x}) \text{ log}(P(\mathbf{x})) d\mathbf{x}$. The mutual information between $X$ and $Y$ is defined as $D_{\text{KL}}(P_{(X,Y)}||P_X \otimes P_Y)$, and can be rewritten as $H(Y) - H(Y|X)$ (the reduction in entropy in $Y$ given knowledge of $X$).

Using the OTR framework, we fix a model $\phi$ and generate a diverse set of $K$ prompt templatizing functions $f_{\theta_1}, f_{\theta_2}, ..., f_{\theta_K}$ along with their corresponding collapsing functions $c_{\theta_k}$
(see Appendix \ref{subsec:promptengineeringprocess}). Treating $f_\theta(X):=\{f_\theta(\mathbf{x}), \mathbf{x} \in X\}$ as a random variable, we can calculate $I(f_\theta(X); Y)$ and use it as a criterion for selecting prompt templatizing functions with which to do inference.

We hypothesize that a $\theta_i$ with higher mutual information will align a language model to a task better than a $\theta_j$ with lower mutual information.  Formally, we select $\hat{\theta} = \mathrm{argmax}_\theta \{I(f_\theta(X); Y)\}$. 

Mutual information is estimated as:
\vspace{-2pt}
\begin{equation}
\label{equation:mi}
I\left(f_\theta(X); Y) = H(Y) - H(Y | f_\theta(X)\right)
\end{equation} 
where each term is estimated in expectation using draws $\mathbf{x}_i \sim X$ and Equation \ref{equation:otr} as follows:
\vspace{-3pt}
\begin{equation}
\label{equation:marginal-entropy}
H(Y) \approx H\left( \frac{1}{N} \sum_{i=1}^N P(Y|f_\theta(\mathbf{x}_i))\right)
\end{equation}
\vspace{-14pt} 
\begin{equation}
\label{equation:conditional-entropy}
H(Y|f_\theta(X)) \approx \frac{1}{N} \sum_{i=1}^N H(P(Y|f_\theta(\mathbf{x}_i))))
\vspace{-6pt} 
\end{equation}
The marginal entropy $H(Y)$ is the entropy of the mean of the conditional distributions, and the conditional entropy $H(Y|f_\theta(X))$ is the mean of entropies of the individual conditional distributions.


This definition gives us another reason to expect that mutual information will work well. Since mutual information is the marginal entropy minus the conditional entropy, maximizing mutual information is equivalent to maximizing marginal entropy and minimizing conditional entropy. Thus, MI is high for templates that are, on average, less biased towards any given answer (high marginal entropy) and templates with outputs the model is confident about (low conditional entropy). These attributes are desirable in constructing prompts, and we postulate that maximizing mutual information will yield a well-aligned template.

Looking at it another way, by the data processing inequality \cite{Cover2006},  $I(f_\theta(X); Y) \leq I(X;Y)$. Thus, $I(f_\theta(X);Y)$ gives a lower bound for $I(X;Y)$, and the highest mutual information is the tightest lower bound. The prompt corresponding to this lower bound preserves the most information between $X$ and $Y$.

\section{Experimental Setup}
\subsection{Datasets}

\newcolumntype{M}[1]{>{\centering\arraybackslash}m{#1}}


\begin{table}
\centering
\small
\begin{tabular}
{ 
|M{1.5cm}|M{2.4cm}|M{0.4cm}|M{.7cm}| M{.4cm} |
}
\hline
Dataset & Task & $|Y|$ & Base Acc. & Size $N_{\mathrm{all}}$\\
\hline
\hline
SQuAD & Open Book QA & $|T_\phi|$ & $\sim 0$ & 16K\\
\hline
LAMBADA & Cloze & $|T_\phi|$ & $\sim 0$ & 5K\\
\hline
ROCStories & Cloze & $|T_\phi|$ & $\sim 0$ & 52K\\
\hline
CoQA & Closed Book QA & 5 & 0.2 & 9K\\
\hline
IMDB & Sentiment Analysis & 2 & 0.5 & 50K\\
\hline
BoolQ & Reading Comprehension & 2 & 0.5 & 16K\\
\hline
COPA & Choice of Positive Alternatives& 2 & 0.5 & 1K\\
\hline
WiC & Word in Context & 2 & 0.5 & 5K\\
\hline
\end{tabular}
\caption{All datasets used in our experiments. $|Y|$ is the size of the label space and $N_{\mathrm{all}}$ is the size of the dataset we sample from (after any modifications).}

\label{tab:datasets}
\end{table}

\begin{figure*}[t]
\includegraphics[trim=0 0 0 0, clip, width=16cm]
{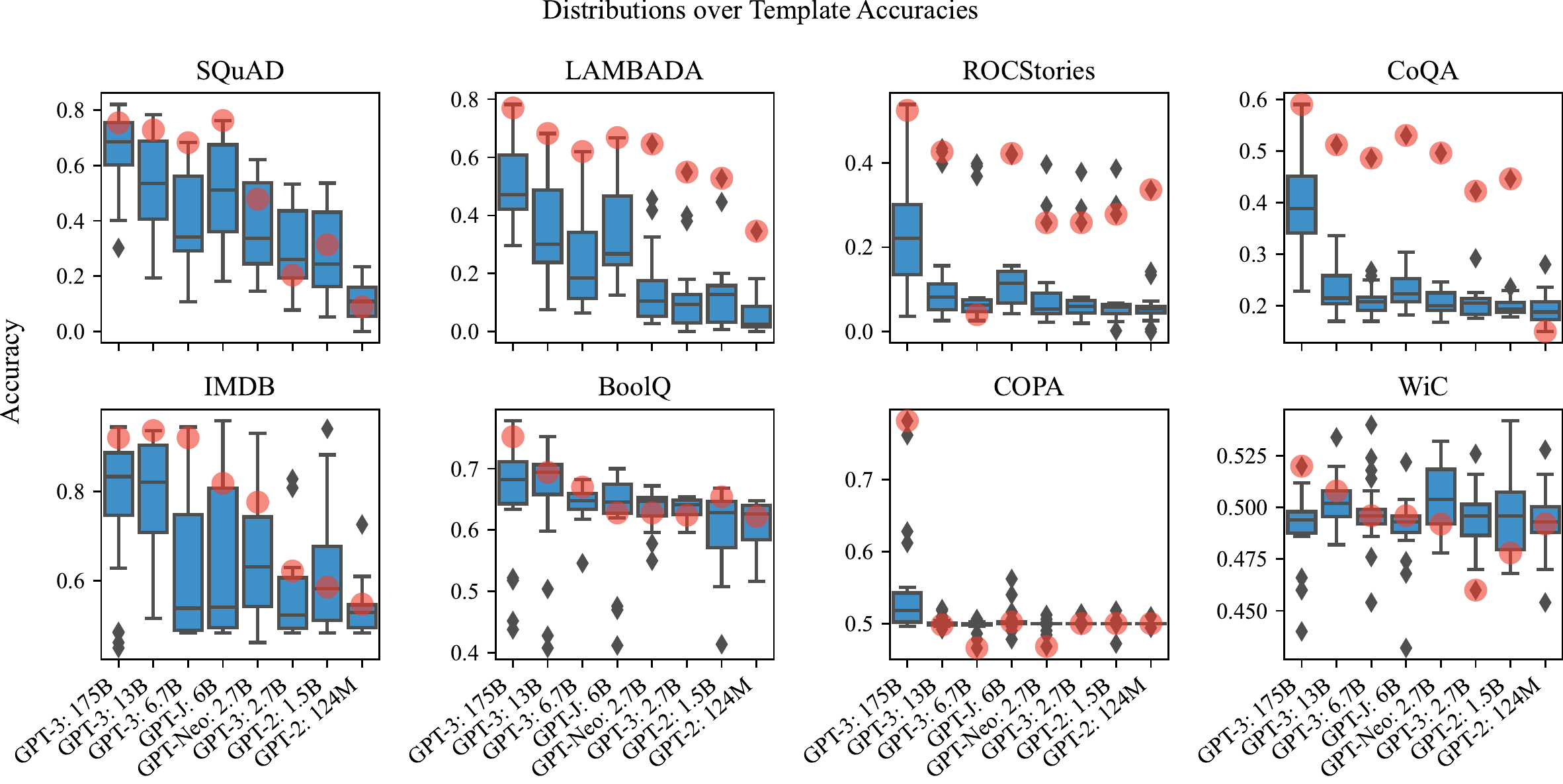}
\centering
\caption{Distributions of accuracies over $K=20$ templates for each model/dataset pair, compared to the prompts selected with MI (translucent red dots).
}
\label{fig:box-whisker}
\end{figure*}

We validate the efficacy of our prompt engineering method with experiments on eight well-known NLP datasets\footnote{Datasets are listed in descending order here and throughout the paper, first by $|Y|$, and then by method performance.}--SQuAD2.0 \cite{Rajpurkar2018}, LAMBADA \cite{paperno2016}, ROCStories \cite{Mostafazadeh16}, CommonsenseQA (CoQA) \cite{Talmor18}, IMDB \cite{Maas11}, BoolQ \cite{Clark19}, COPA \cite{Gordon12}, and WiC \cite{Pilehvar18})--that span seven unique NLP tasks (see Table \ref{tab:datasets}). We used a random sample of $N=500$ samples from each dataset for our experiments.\footnote{We sampled from the train sets of CoQA and SQuAD; the train and validation sets of WIC, COPA, and BoolQ; the full datasets of ROCStories and IMDB; and the test set for LAMBADA.} For ROCStories, which consists of a set of five sentence stories, we randomly masked a word from each story in order to use the data for masked word prediction (cloze).

We made minor changes to two of the datasets in order to cast the associated tasks into OTR. For the SQuAD dataset, we dropped all questions that did not have a one word answer. For the CoQA dataset we dropped all questions with answer choices that started with a shared first word (e.g, the dog, the cat, the monkey). 
Both changes were to decrease ambiguity about which option the model was choosing given its output distribution for a single token.

\subsection{Models}
We assess our method on eight models ranging from 124 million to 175 billion parameters
: These include GPT-2 124M \& 1.5B \cite{gpt-2}, GPT-Neo 2.7B \cite{gpt-neo}, GPT-J (6B) \cite{gpt-j}, and (Ada, Babbage, Curie, \& Davinci) GPT-3 \cite{Brown2020}. We assume (per \cite{Perez2021}) these models to correspond, respectively, to the 2.7B, 6.7B, 13B, and 175B models in \cite{Brown2020}. 
Each is a causal language model, and although we do not include masked language models, this is a promising area for future work.

\section{Results}
\label{sec:results}
In this section, we analyze our experiments. First, we look at our method's ability to select high-accuracy prompts across models and datasets (Section \ref{subsec:MI-performance}). Next, we correlate template mutual information and accuracy in Section \ref{subsec:corr}. After that, we compare our method and template selection using labeled examples in Section \ref{subsec:baseline}. In Section \ref{subsec:ensemble}, we explore the robustness of MI and use ensembling to improve it. Finally, we compare the tranferability of prompt templates selected with MI from model to model in Section \ref{subsec:transfer}.

\newcommand{\resultsubsec}{\subsection}


\resultsubsec{Template Selection Performance}
\label{subsec:MI-performance}

\begin{figure}[t]
\includegraphics[trim=0 0 0 0, clip, height=6.5cm]
{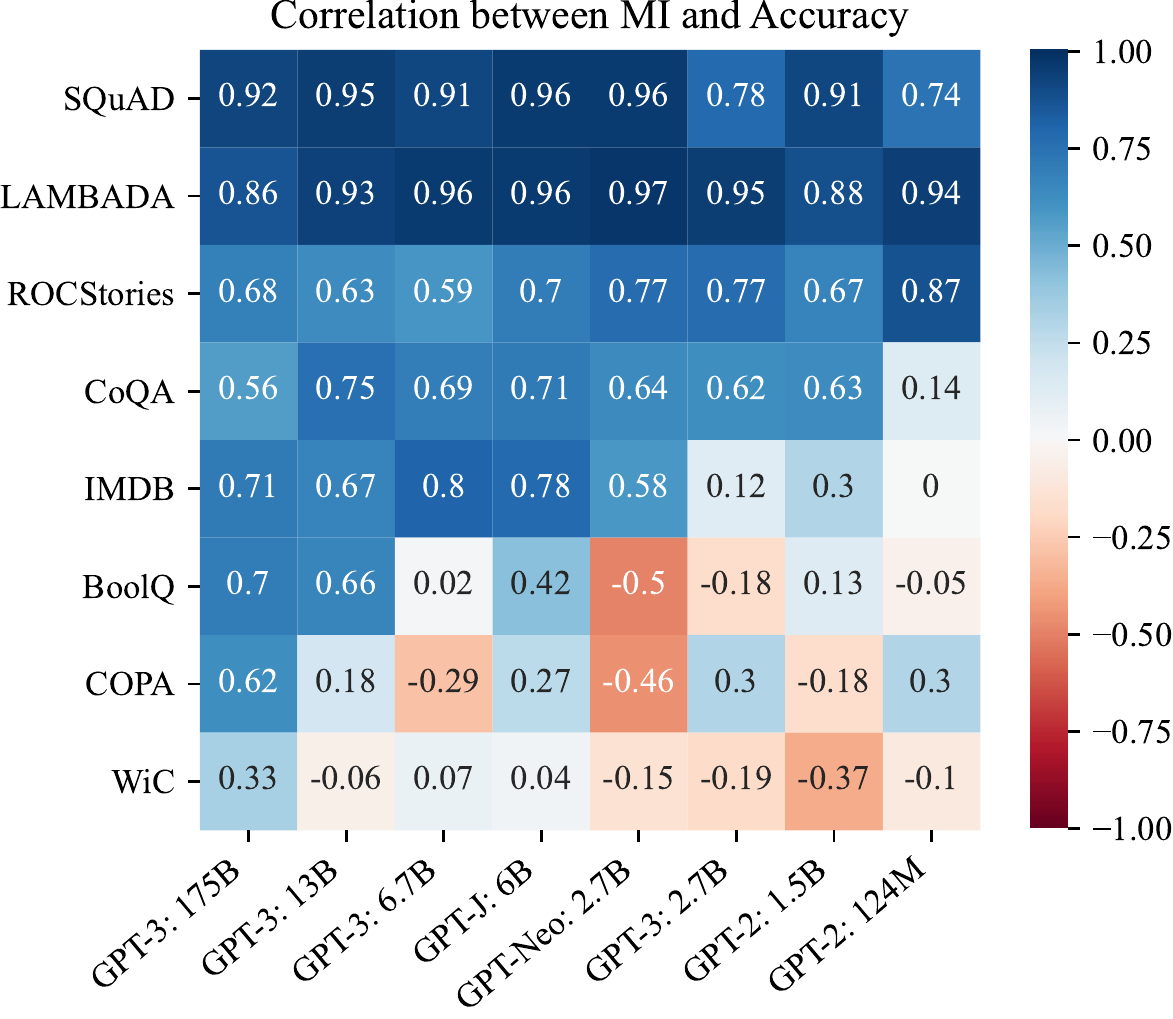}
\centering
\caption{Correlations are more consistently high across all tasks for the largest models, suggesting that our method is most useful at those model sizes.}
\label{fig:corr-conc}
\end{figure}

\begin{figure}[t]
\includegraphics[trim=0 0 0 0, clip, height=11cm]
{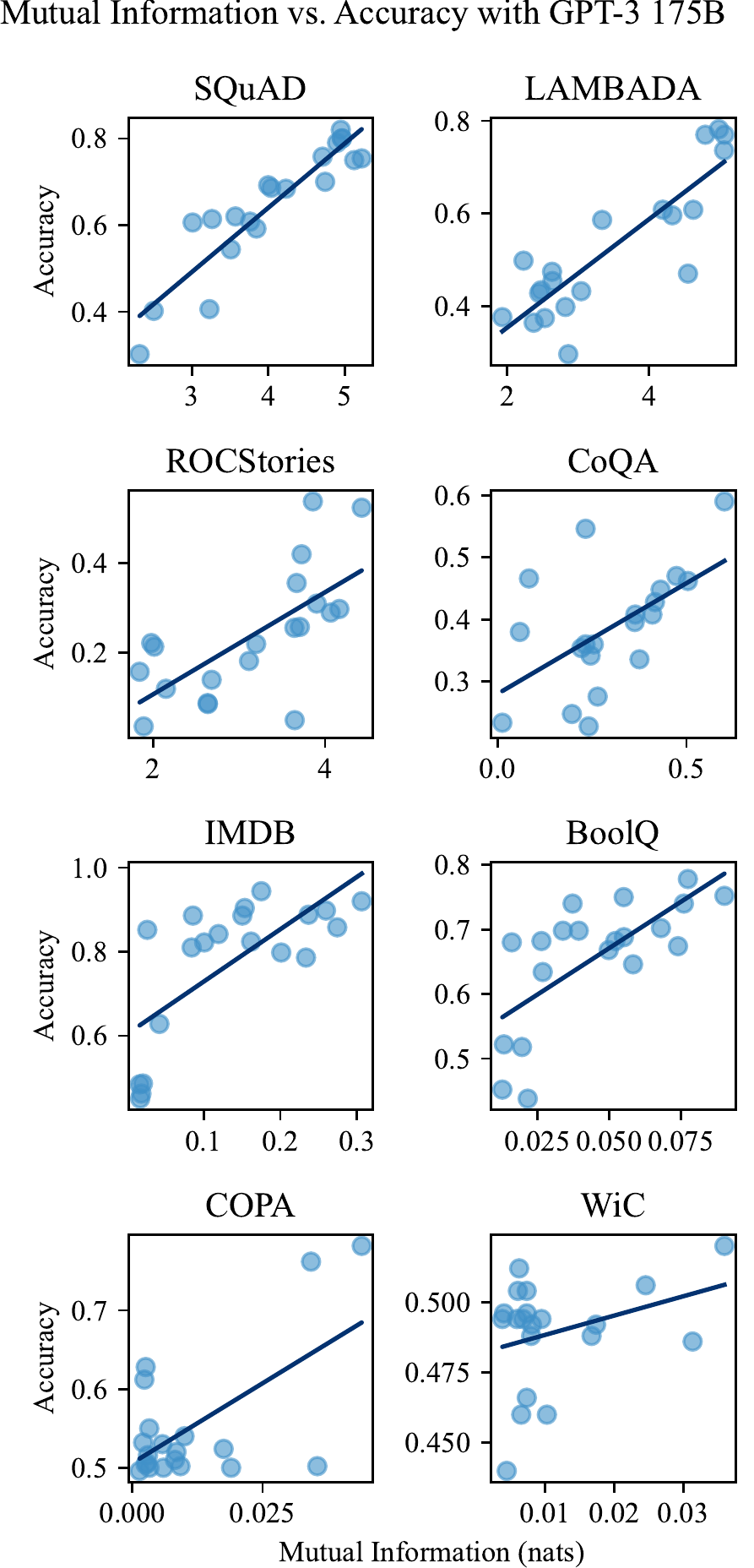}
\centering
\caption{Each dot represents a template and its average mutual information and accuracy over $N=500$ task instances. Linear best fit (by mean standard error) lines are included to show overall trends.}
\label{fig:davinci-scatter}
\end{figure}


We first define baselines against which we compare our approach. Other prompt engineering methods generally require either access to model weights, labeled data (validation set selection), or both (backprop/continuous prompt embedding methods). Our method does not require these, so we instead compare to random and oracle baselines. A random template selection method would give us the average accuracy of our template set (in expectation), while an oracle selection method would give us the best accuracy every time. To understand how our MI method compares to these two baselines for each dataset, refer to Figure \ref{fig:cover-plot}, where we analyze performance on GPT-3 175B. On each of the eight datasets, mutual information selects a prompt template that outperforms both the mean and median accuracies (random baseline performance). In three of the eight datasets, mutual information selects the best (highest accuracy) template from the 20 proposed (equivalent to oracle performance). 
\begin{figure*}[t]
\definecolor{kde_blue}{RGB}{157, 206, 226}
\definecolor{avg_blue}{RGB}{64, 145, 201}
\definecolor{ens_20_blue}{RGB}{3, 50, 112}
\definecolor{ens_mi_red}{RGB}{239, 60, 45}
\definecolor{dark_red}{RGB}{169, 20, 10}
\includegraphics[trim=0 0 0 0, clip, width=16cm]
{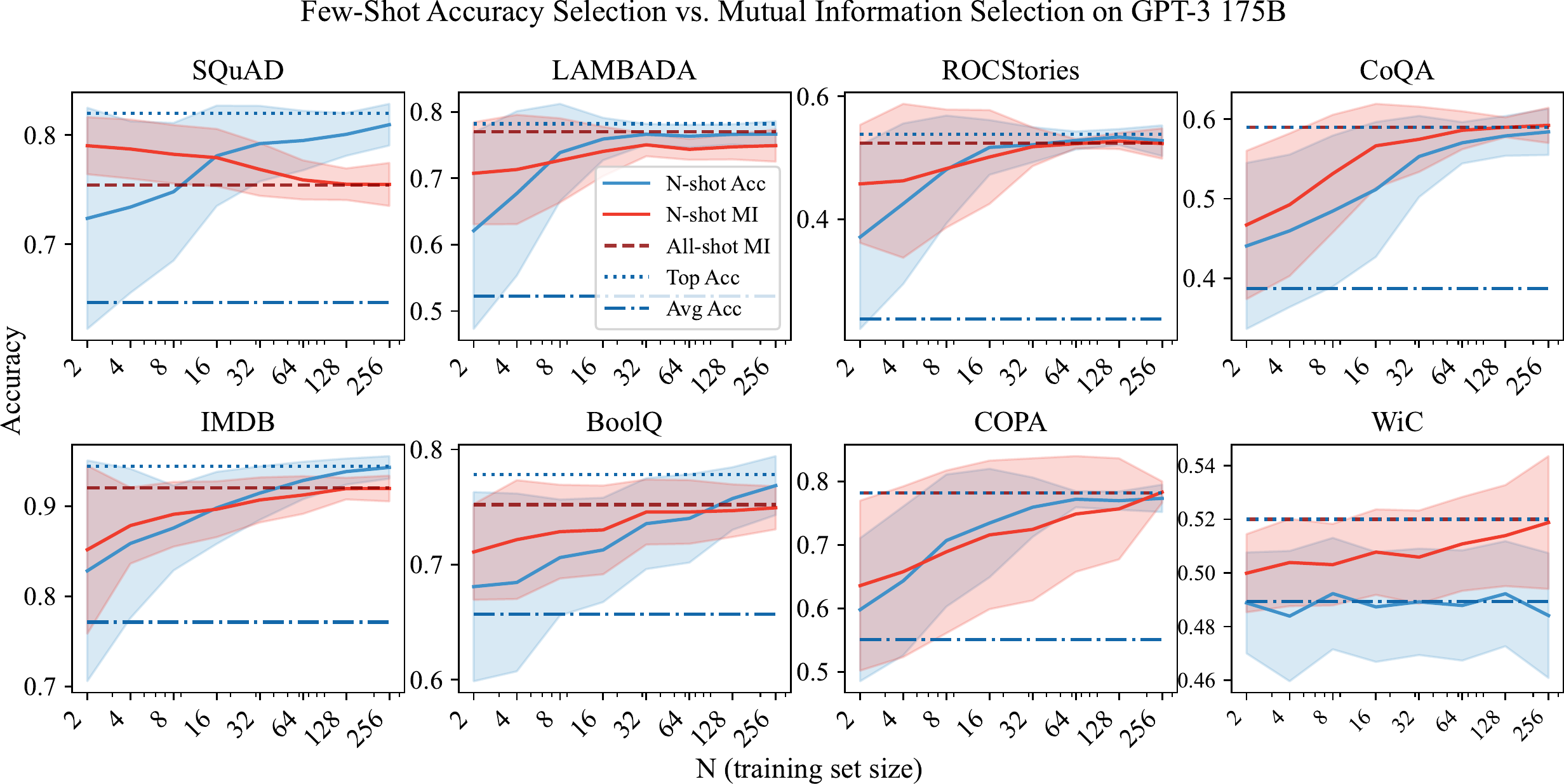}
\centering
\caption{For $P=100$ random train/test set partitions for each training size $N=2, 4, 8, ..., 256$, we select a template based on accuracy (\textcolor{avg_blue}{N-shot Acc}) and based on mutual information based on just those $N$ examples (\textcolor{ens_mi_red}{N-shot MI}). Then, we report accuracy of that template on the test set (size: $500 - N$). Error bars ($\pm \sigma$) are reported across the $P=100$ partitions. For reference, the \textcolor{ens_20_blue}{highest}, \textcolor{ens_20_blue}{average}, and \textcolor{dark_red}{full-dataset MI template} accuracy is also reported.}
\label{fig:baseline}
\end{figure*}

\begin{figure*}[t]
\definecolor{kde_blue}{RGB}{157, 206, 226}
\definecolor{avg_blue}{RGB}{64, 145, 201}
\definecolor{ens_20_blue}{RGB}{3, 50, 112}
\definecolor{ens_mi_red}{RGB}{239, 60, 45}
\title{Accuracy given different ensembling techniques}
\vspace{1 mm}
\includegraphics[trim=0 0 0 0, clip, width=16cm]
{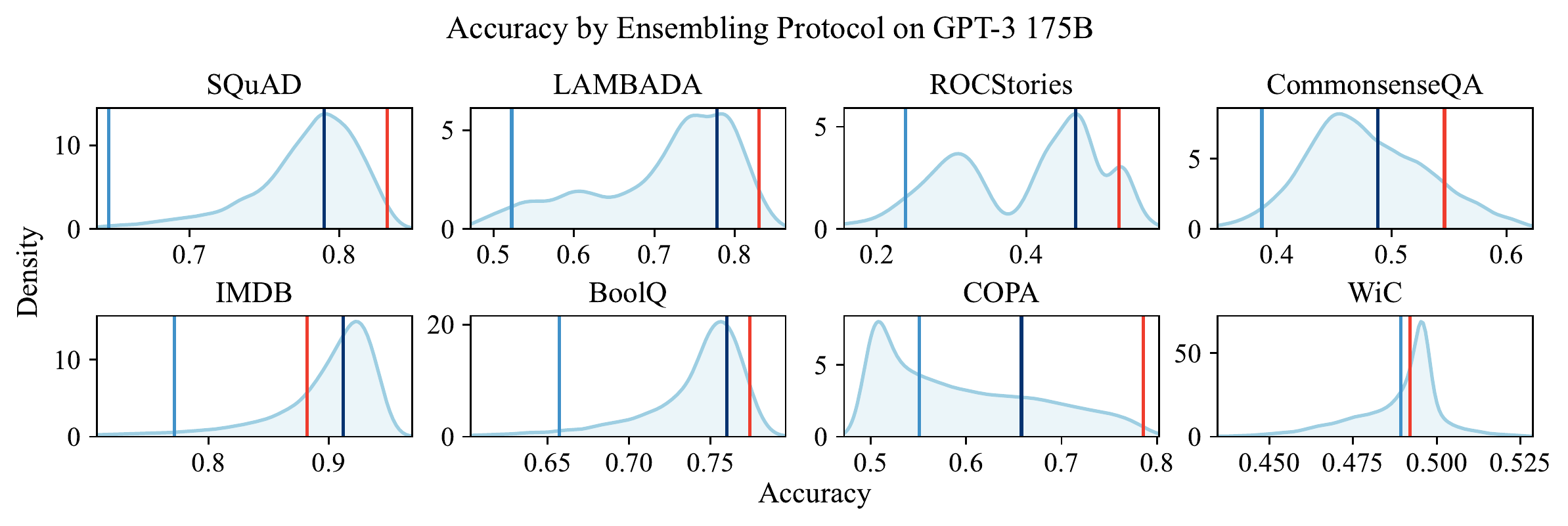}
\centering
\caption{For each dataset the \textcolor{kde_blue}{KDE plot} represents accuracy over each of the $20 \choose 5$ ensembles of 5 templates from the 20 templates associated with the dataset. Each plot also includes lines representing the \textcolor{avg_blue}{average accuracy of all single templates for the dataset}, the \textcolor{ens_20_blue}{accuracy of the ensemble of all 20 templates}, and the  \textcolor{ens_mi_red}{accuracy of the ensemble of the top 5 templates chosen by MI}. 
In only one case does \textcolor{ens_20_blue}{all-20} beat \textcolor{ens_mi_red}{top-5-MI}, and it does so at 4$\times$ the cost.}
\label{fig:ensembling-kde}
\end{figure*}

Given our method's promising performance with GPT-3 175B, it is natural to ask how it performs with smaller models. Figure \ref{fig:box-whisker} shows the accuracy distributions over prompt templates for each dataset/model pair. With every model, MI gives above-average performance on several datasets. Although MI is more likely to select a high accuracy template for larger models, it is a good criterion even for smaller models on all but two datasets, COPA and WiC. Note that, for these two datasets, none of the templates do significantly better than chance ($\sim$50\%) besides the largest model on COPA, which is in line with previous work.\footnote{Our template's best accuracy is 54\% for WiC, and 78.2\% for COPA, which is similar to previous work (WiC: \cite{Brown2020} - 49.4\%, \cite{Perez2021} - 54.1\%; COPA: \cite{Brown2020} - 92.0\%, \cite{Perez2021} - 84.8\%).} Thus, we observe that mutual information performs best when there is a high-signal prompt to select, and worse when all prompts are low-signal.
%

When considering all other datasets, MI selects an above average prompt 83\% of the time for all models; for the largest two models, MI selects an above average template 100\% of the time.

\begin{figure*}[t]
\includegraphics[trim=0 0 0 0, clip, width=13cm]
{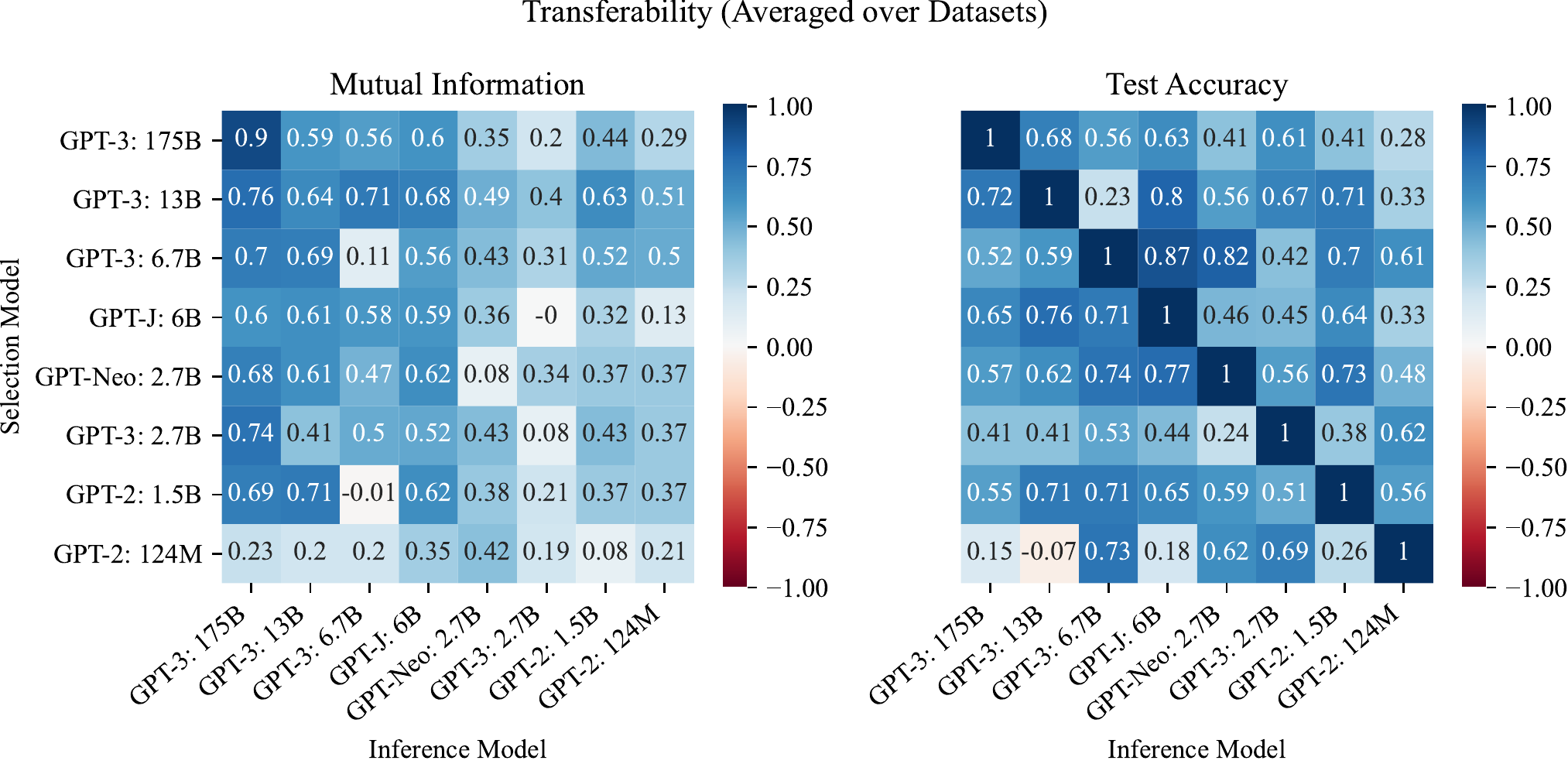}
\centering
\caption{For each model/dataset pair, accuracies are normalized linearly so that 0 is the average prompt accuracy and 1 is the highest test accuracy. Using the prompt chosen by either MI or test accuracy on each selection model, average performance across datasets is reported for each inference model.}
\label{fig:transfer-heatmap}
\end{figure*}
\resultsubsec{Correlation between Template Mutual Information and Accuracy}
\label{subsec:corr}
In Section \ref{subsec:MI-performance}, we see how the mutual information selected template does in terms of accuracy compared to all other templates. We have not discussed, however, how generally MI and accuracy are correlated, except that the highest MI template tends to have anomalously high accuracy. Here, we establish that their correlation is high across all templates for the largest LMs. 
Each of the $K=20$ templates has two corresponding measures: average accuracy and average MI. We can use these pairs to correlate MI and accuracy via Pearson's R.


We see in Figure \ref{fig:corr-conc} that the correlations are surprisingly high for the majority of models and datasets. For SQuAD, LAMBADA, ROCStories, and CoQA, this pattern holds across all model sizes; for the remainder, results are good on larger models and are much less reliable on smaller models. Overall, this is evidence that as mutual information increases, so does accuracy. In other words, mutual information can be used to make an educated guess about accuracy without having to use any ground truth labels, especially on larger models.
\resultsubsec{Compared to Few Labeled Examples}
\label{subsec:baseline}

Next, we ask: How does our method compare to selecting a template based on the accuracy of a few-labeled examples? Also, how many unlabeled examples does MI need to be able to perform well?

Results with the largest model are reported in Figure \ref{fig:baseline}. Note that with as few as $N=2$ instances, MI selects a far better than average template, allowing performance gains even in the low-data, unlabeled regime. Additionally, for low $N$ and across all eight datasets, MI even selects a better template on average than selecting based on labeled train set accuracy. This suggests that, even with labeled examples, selecting based off of MI may be preferable to test accuracy with few examples. Selecting by labeled train set accuracy often begins to perform better at higher $N$, but at the cost of requiring labeled data, while our method needs no labels.


\resultsubsec{Method Robustness and Ensembling}
\label{subsec:ensemble}
To explore our method's robustness we consider the question: what if we had included a different subset of templates, especially not including the top MI template? Figure \ref{fig:davinci-scatter} shows average MI/accuracy data for all $K=20$ prompt templates on GPT-3 175B (similar plots for other models are found in Appendix \ref{subsec:all-scatter}). For six of eight datasets, the results are robust; the top few prompt templates (by MI) are all high performers. The performance for COPA and WiC is more brittle; excluding the top-MI template would have resulted in a large drop in accuracy. This attests to the utility of generating a diverse slate of templates as recommended in Appendix \ref{subsec:promptengineeringprocess} and also to the risk that outliers could compromise our method's effectiveness.

A comprehensive discussion of remedies for outliers is beyond the scope of this paper, but it is an important concern. Considering the strength of MI/accuracy correlations, one simple approach is to ensemble the top 5 MI templates. 

To compare this principled top-5 ensemble to other possible ensembles of templates, we take all ${20 \choose 5}$ subsets of 5 templates from all 20 templates and calculate the accuracy of each ensemble. For each dataset, we plot this distribution's kernel density estimate, which models the p.d.e. of the random variable ``accuracy of 5 random templates ensembled together''. We then compare the top-5 MI ensemble to other possible ensembles. The results are shown in Figure  \ref{fig:ensembling-kde}.

We found that the top-5 MI ensemble does at least as well as the top-20 ensemble in all but one case. Two reasons to use MI are, then, that 1) the MI ensemble gets as good or better a result as ensembling all prompt templates and 2) at a fourth of the experimental cost.  
In short, ensembling by MI is a cheap and effective way to guard against anomalous high MI/low accuracy templates.

\resultsubsec{Transferability across Models}
\label{subsec:transfer}

Finally, we explore how well-chosen templates generalize between models.  Concretely, we choose templates by maximizing either test accuracy (oracle) or mutual information (our method) using a selection model $\phi_s$, and then calculate test accuracy using a different inference model $\phi_i$. We calculate absolute test accuracy and then normalize it such that 0 and 100 correspond to the average and maximum scores across templates for a model/dataset pair. 
We average our results across datasets and present the results in Figure \ref{fig:transfer-heatmap}. Prompt transfer for each dataset can be found in Appendix \ref{app:transfer}.

MI performance is best when the largest model (GPT-3 175B) is used as both the selection and inference model: on average, MI scores 90\% on this normalized scale. Additionally, performance is most consistently high when the largest models are used either for selection or inference. But almost all transfer scores are well above 0 (only one negative average gain out of 64 transfer permutations), suggesting that transfer is often effective. 

Overall, we have observed that prompt selection by mutual information is surprisingly effective across a variety of datasets and model sizes. This method works best on larger models and for tasks that the LM is capable of performing. Given the high diversity of tasks that we have explored, we expect this method to transfer well to many other NLP tasks, including regimes with little labeled data.


\section{Conclusion}


In this paper, we introduce a method for selecting prompts that effectively align language models to NLP tasks. Over a set of candidate prompts, our method selects the template that maximizes the mutual information between the input and the model output. We demonstrate that 1) mutual information is highly correlated with test accuracy and 2) selecting a prompt based on mutual information leads to significant accuracy gains over random choice, approaching oracle performance on GPT-3 175B, and it does so across model sizes and tasks. 

Whereas other methods rely on ground truth labels and/or direct model access, ours requires neither. 
Many applications characterized by lack of computational resources, limited model access (e.g., inference only), and lack of ground truth data prohibiting testing of candidate prompts become feasible with our method.


\section{Ethics}

There are many ways to prompt a language model poorly, and there still seem to be NLP tasks which are beyond alignment regardless of model size or prompt quality. This method cannot align a LM to a task if the entire set of prompts is poor or, obviously, if the model cannot be aligned. High mutual information does not necessarily imply high accuracy despite the strong correlation we found. Thus, our method should only be employed on a task if there is some understanding of how high MI needs to be on a domain or set of templates to imply a sufficiently high accuracy for safe use.

Otherwise, we introduce no model, dataset, or other contribution that might warrant ethical concern.

\section*{Acknowledgements}
We thank the anonymous reviewers for their helpful feedback. This material is based upon work supported by the National Science Foundation under Grant No. RI 2141680.


\newpage
\bibliographystyle{acl_natbib}
\bibliography{custom}

\newpage

\appendix

\section{Prompt Engineering Process}
\label{subsec:promptengineeringprocess}

In this section, we step through our method in detail. Again, note that this method uses no ground truth labels and does not require gradient updates or model parameter access. Given a task that can be represented in natural language with the OTR framework, the only requirements for our approach are a) several candidate prompt templates and b) some instances ($X$) on which to do inference.

\begin{enumerate}[wide, labelwidth=!, labelindent=0pt]
\item \textbf{Generate a set of $K$ prompt templatizing functions with corresponding collapsing functions.} Each prompt template function $f_{\theta_k}$ should take in an input from the dataset and output a prompt ready for processing by the language model.
We chose to generate our template functions by hand, i.e., a human writes a sensible, custom natural language scaffolding that can be filled with input data (see examples in \ref{appendix:template-examples}).

Each template must also have a collapsing function $c_{\theta_k}$ that takes the language model output logprobs, exponentiates and sums ``equivalent'' logprobs, and normalizes the resulting probabilities to produce a distribution over targets. Equivalent logprobs are those that indicate the same answer. For example, 
a template might be designed for a question-answering task with possible answers ``Yes'' and ``No''. We consider all logits corresponding to possible lexical variants of each of these answers to be equivalent. For example, what logits should count toward the answer ``Yes''? Not just the exact token ``Yes'', since ``ye'', `` yes'', and ``YES'' are all lexical variants of the same answer or the beginning of it, just with surrounding white space and alternative capitalization. The collapsing function lower-cases and strips white space from all logits, and if the lower-cased answer begins with a token, that token's probability (the exponentiated logprob) is added to the sum of probability for that answer. Finally, the sums of probabilities for all individual answers are normalized. Prompt template functions should be chosen to be as diverse as possible to increase the probability of finding high-quality prompts. For example, we use templates that frame input from datasets as test questions, back and forth dialogue between friends, Python code, test answer banks, etc. A sample of the prompt templates used in this work is provided in Appendix \ref{appendix:template-examples}. A good resource for coming up with prompt template function ideas is the \href{https://beta.openai.com/examples}{OpenAI API examples collection}\footnote{\href{https://beta.openai.com/examples}{beta.openai.com/examples}}. While we aimed for as diverse a set of prompts as possible in this work, additional dimensions of variation in prompt templates could be explored in future work (e.g., ordering of few-shot examples).

\item \textbf{Playground.} For each chosen $f_{\theta_k}$, calculate $g_{\phi}(f_{\theta_k}(x))$ for a few dataset samples. Do not look at associated ground truth labels for these samples. Simply check to ensure that $g_{\phi}$ puts high probability on the tokens one would expect given $f_{\theta_k}$ that could be reasonably collapsed by $c_{\theta_k}$ into $P(Y)$. For example, on the BoolQ reading comprehension task, the language model predicts the answer to a yes/no question with a corresponding passage. Given this task, we would expect the highest probability to be on tokens like ``Yes'' or ``No''. A poor prompt template, though, might put the highest probability on unrelated tokens like ``I'', ``think'', or ``\textbackslash n''. Revise or replace any template that fails to put high probability mass on the tokens expected.

\item \textbf{Estimate mutual information for each  template $f_{\theta_k}$.} Choose how many data points $N$ to use for estimating mutual information for each template function. A higher $N$ will allow for estimation of mutual information based on a more representative sample of the dataset at the cost of more LM computation. Sample $N$ samples from your dataset. Since we do not require any $Y$ labels, one could even choose the $X$'s on which you desire to do inference (as we do). Then, for each sample $x$ and each template $f_{\theta_k}$, calculate $P(Y|f_\theta(x))$ using Equation \ref{equation:otr}. Use the output to estimate MI for each prompt template with Equation \ref{equation:mi}.

For all of our experiments, $c_\theta$ takes in a distribution of tokens $g_{\phi}(f_{\theta_k}(x))$ and a mapping between the set of possible ground truth labels for $f_{\theta_k}(x)$ and model vocabulary $T_\phi$. For a sentiment analysis task, that mapping would be from the ground truth labels ``positive'' and ``negative'' to the expected tokens ``positive'' and ``negative'' respectively. If a prompt template for the task was phrased as a yes/no question, the mapping for it would be from ``positive'' and ``negative'' to ``yes'' and ``no'' respectively. Our $c$ function returns a probability over $Y$ (target label space), and the highest probability label is treated as the prediction. To keep things simple, the values in our map are always single tokens. See examples in Appendix \ref{appendix:template-examples}. 

\item \textbf{Choose prompt template(s) to use for inference based on mutual information.}
For choosing a single prompt template to use for inference, select the template with highest estimated mutual information. With an increased computational budget, one could also ensemble the top $p$ prompt templates, as we describe in Section \ref{subsec:ensemble}.

\item \textbf{Use chosen prompt template(s) to perform inference}
Use chosen prompt template(s) $f_{\hat{\theta}}$ to calculate $c_{\hat{\theta}}(g_{\phi}(f_{\hat{\theta}}(x))$ for each dataset sample. Inference can be done with the language model used for estimating mutual information or a smaller model if cost is prohibitive (for information on performance statistics with this approach, see Figure \ref{fig:transfer-heatmap}).
\end{enumerate}

\clearpage

\section{Additional Figures}

\subsection{Mutual Information vs. Accuracy}
\begin{figure*}[t]
\includegraphics[trim=0 0 0 0, clip, width=16.25cm]{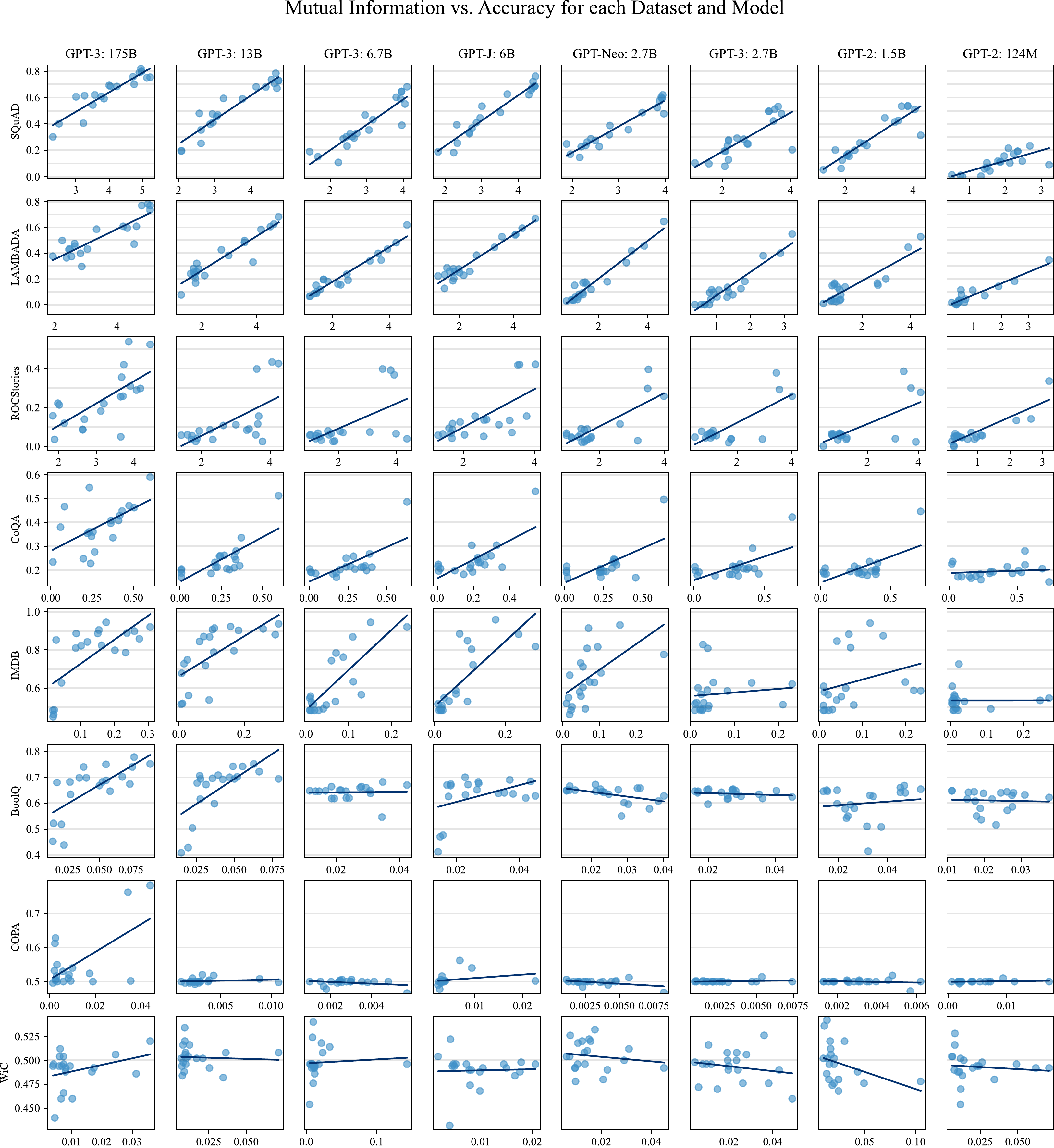}
\caption{Mutual information plotted against accuracy per prompt for each dataset using GPT-3 175B with linear best fit (by MSE) lines to show overall trends}

\label{fig:big-scatter}
\end{figure*}

See Figure \ref{fig:big-scatter}.
\label{subsec:all-scatter}



\subsection{Per Dataset Transfer Heatmaps}
\label{app:transfer}
See Figures 10-17.

\begin{figure*}[t]
\includegraphics[trim=0 0 0 0, clip, width=15cm]
{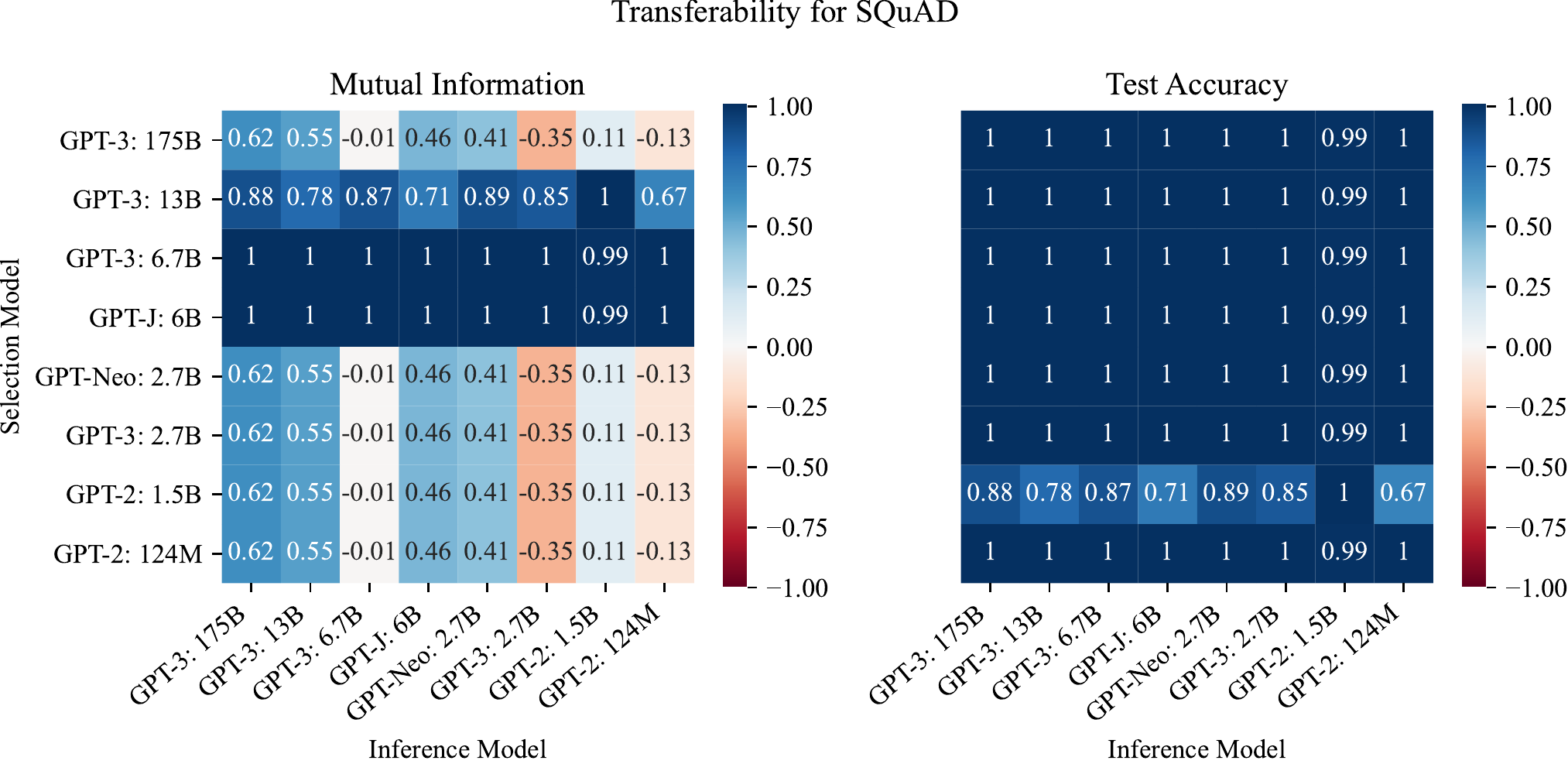}
\centering
\caption{Prompt transfer performance for SQuAD}
\label{fig:transfer-squad}
\end{figure*}

\begin{figure*}[t]
\includegraphics[trim=0 0 0 0, clip, width=15cm]
{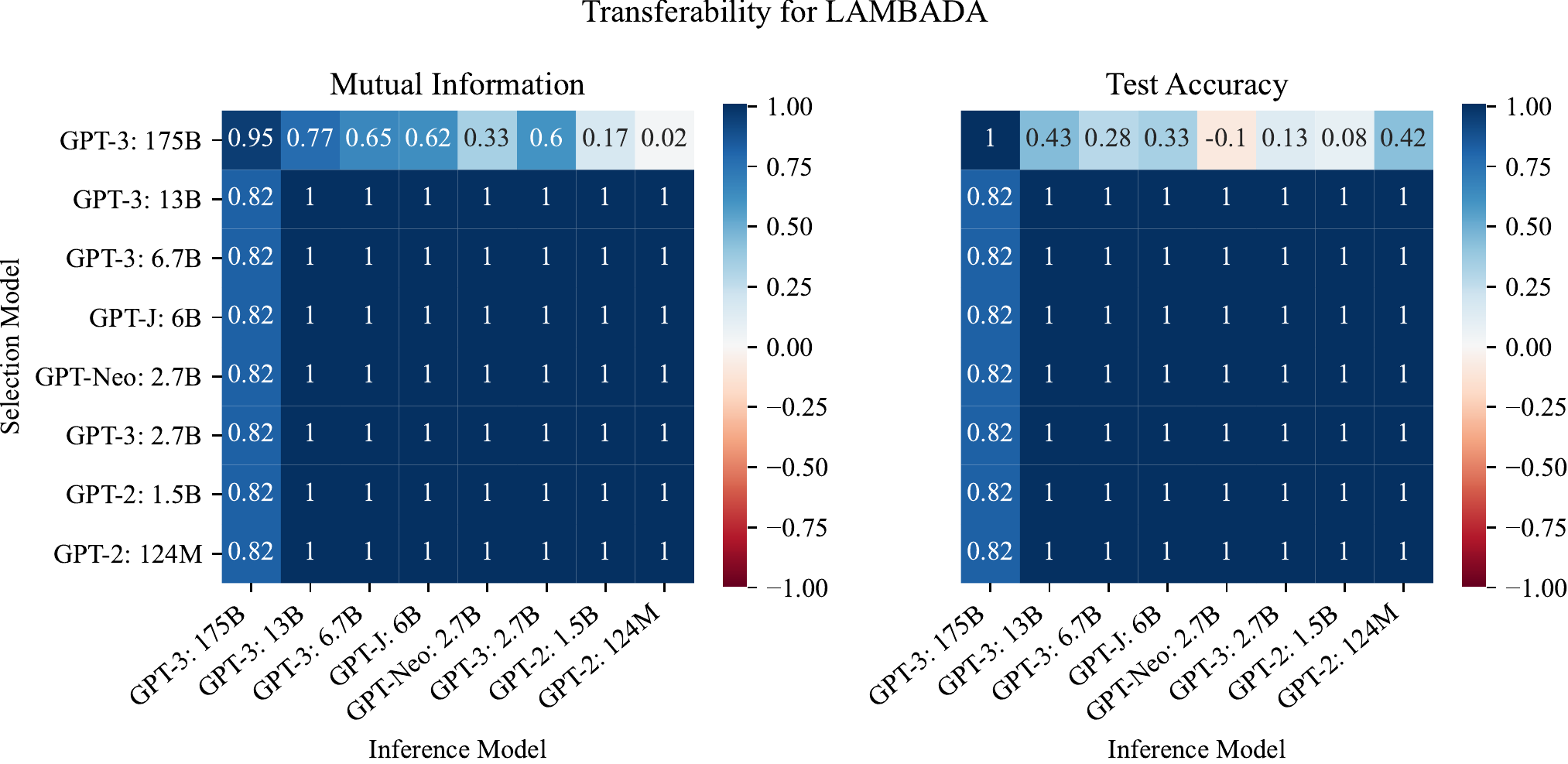}
\centering
\caption{Prompt transfer performance for LAMBADA}
\label{fig:transfer-squad}
\end{figure*}

\begin{figure*}[t]
\includegraphics[trim=0 0 0 0, clip, width=15cm]
{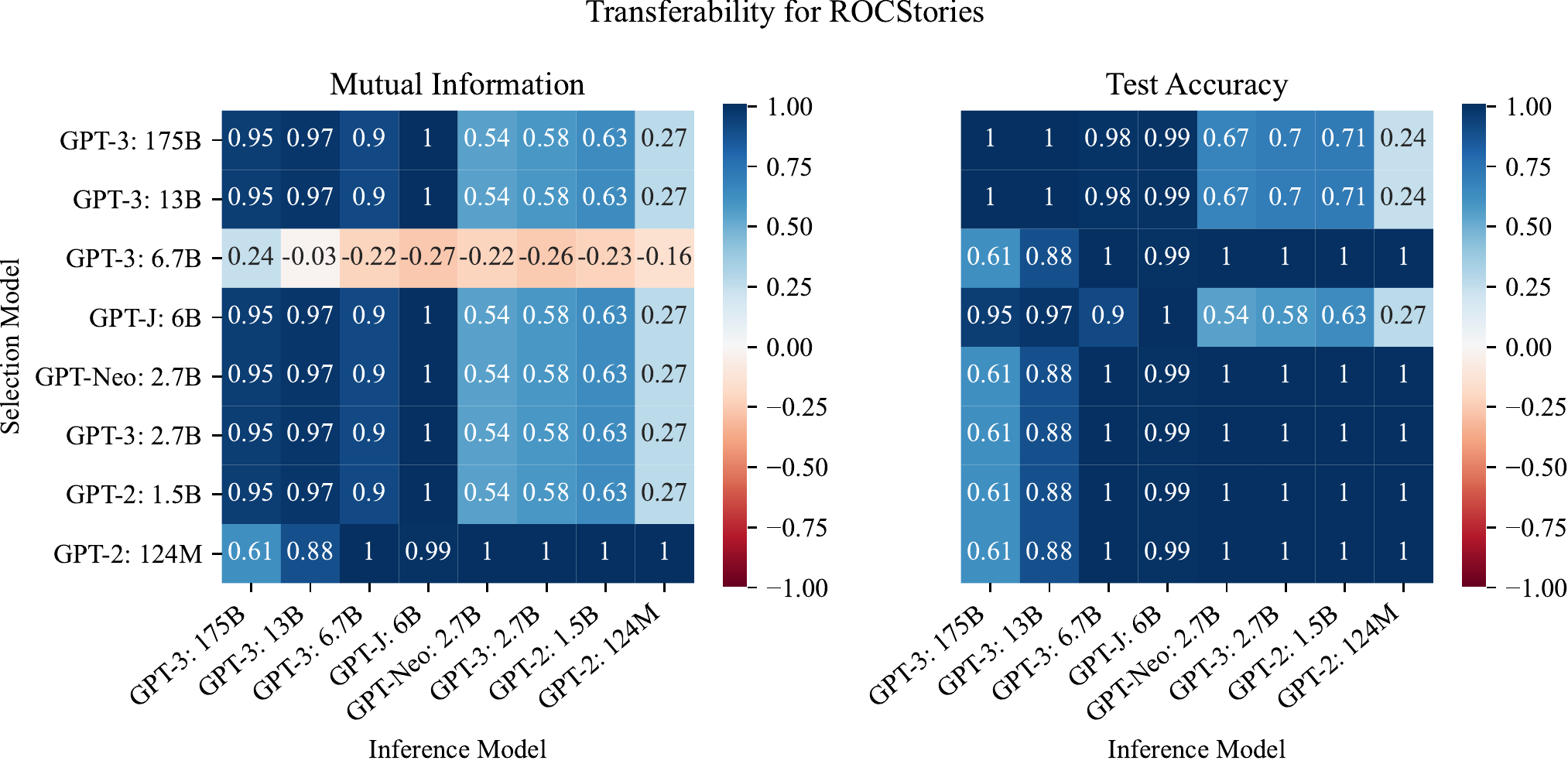}
\centering
\caption{Prompt transfer performance for ROCStories}
\label{fig:transfer-squad}
\end{figure*}

\begin{figure*}[t]
\includegraphics[trim=0 0 0 0, clip, width=15cm]
{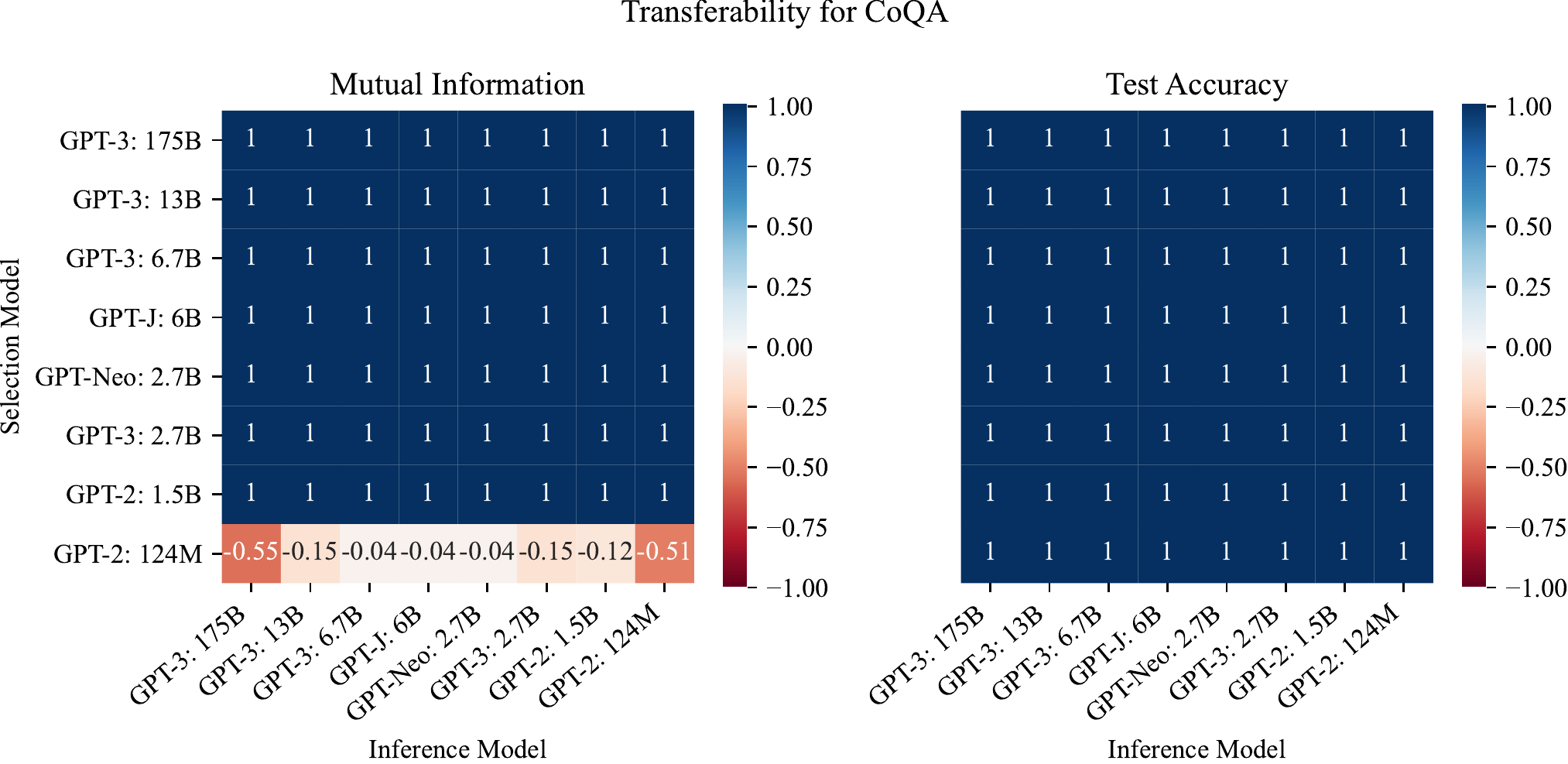}
\centering
\caption{Prompt transfer performance for CoQA}
\label{fig:transfer-squad}
\end{figure*}

\begin{figure*}[t]
\includegraphics[trim=0 0 0 0, clip, width=15cm]
{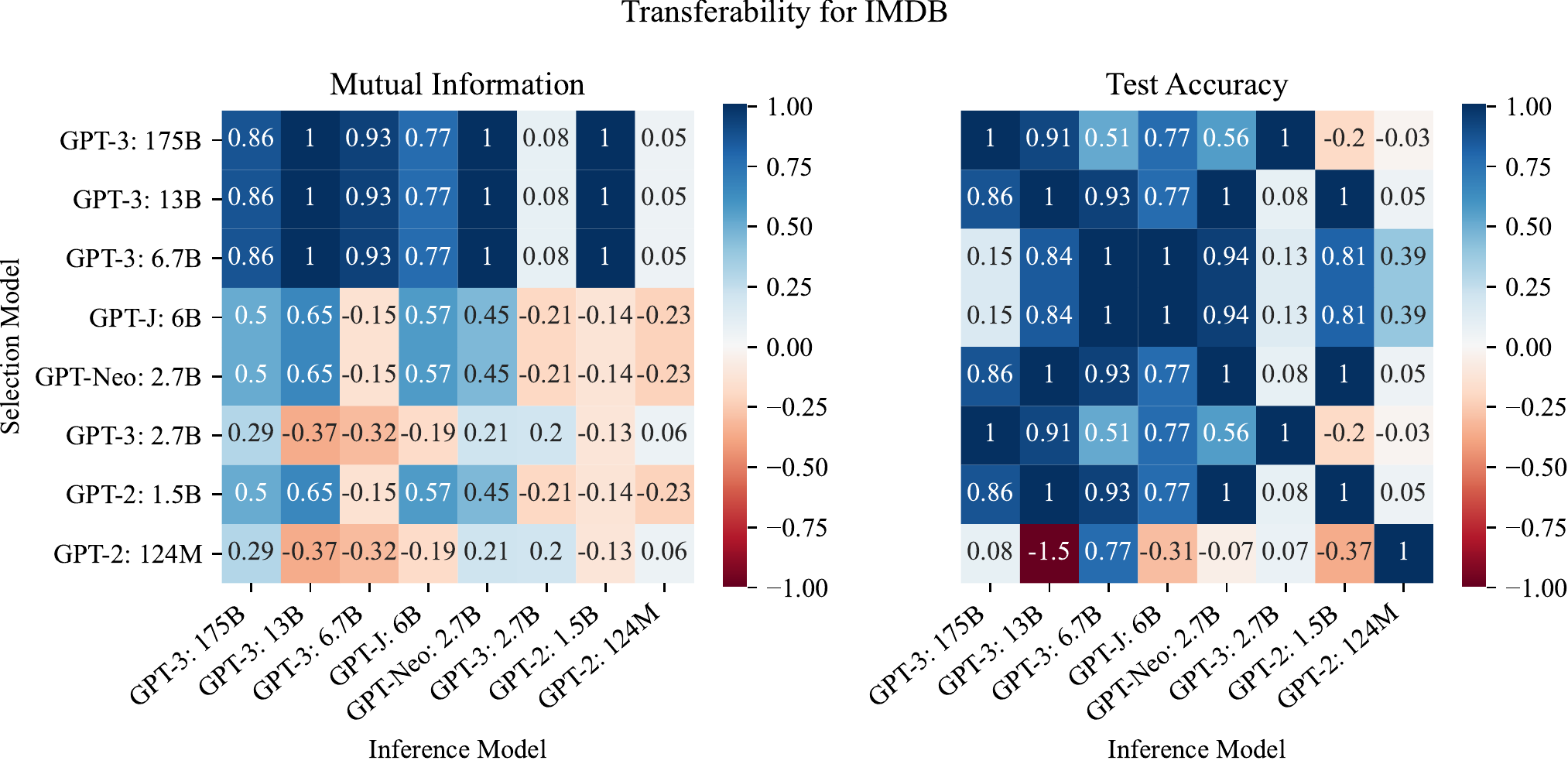}
\centering
\caption{Prompt transfer performance for IMDB}
\label{fig:transfer-squad}
\end{figure*}

\begin{figure*}[t]
\includegraphics[trim=0 0 0 0, clip, width=15cm]
{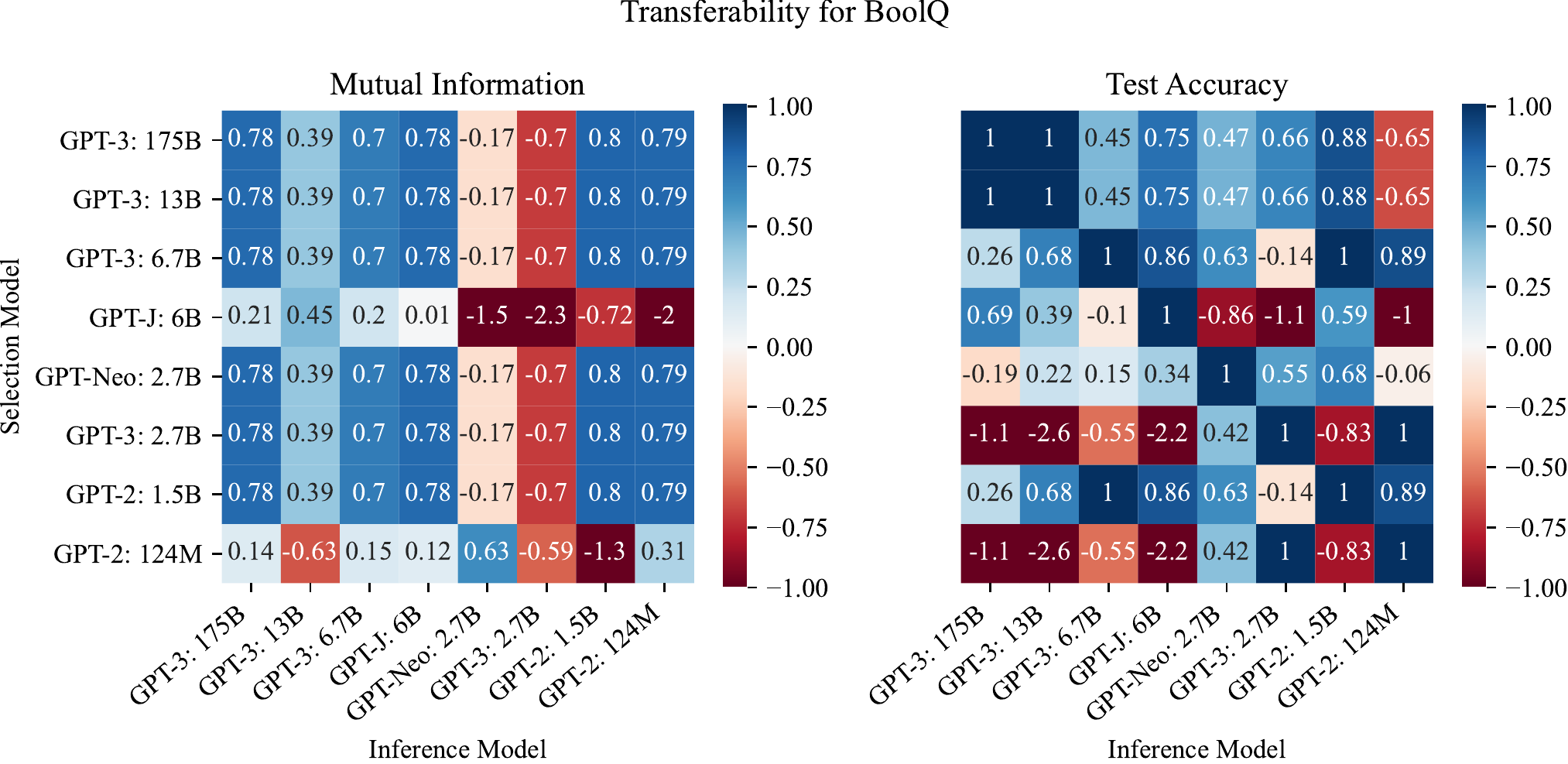}
\centering
\caption{Prompt transfer performance for BoolQ}
\label{fig:transfer-squad}
\end{figure*}

\begin{figure*}[t]
\includegraphics[trim=0 0 0 0, clip, width=15cm]
{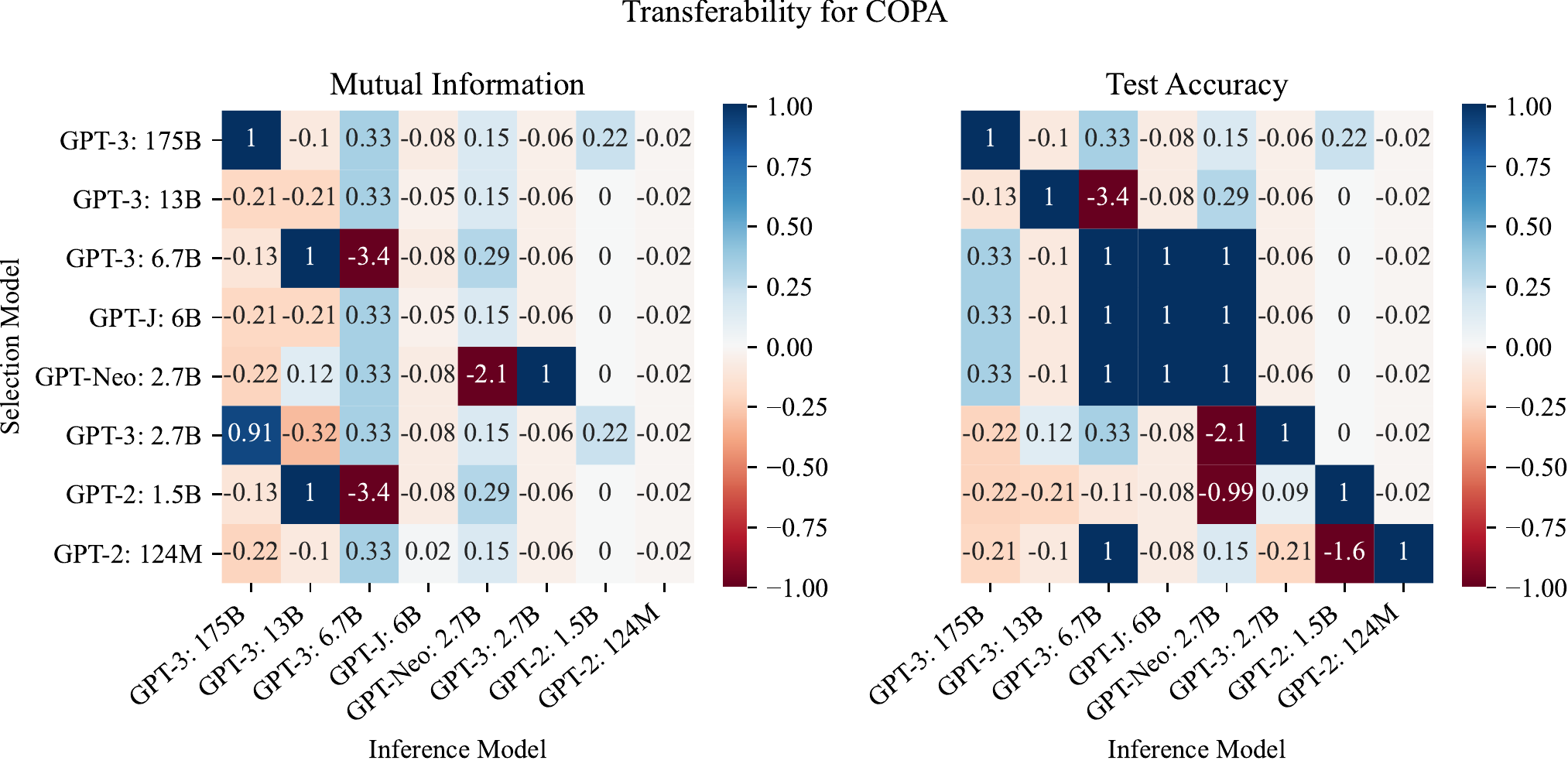}
\centering
\caption{Prompt transfer performance for COPA}
\label{fig:transfer-squad}
\end{figure*}

\begin{figure*}[t]
\includegraphics[trim=0 0 0 0, clip, width=15cm]
{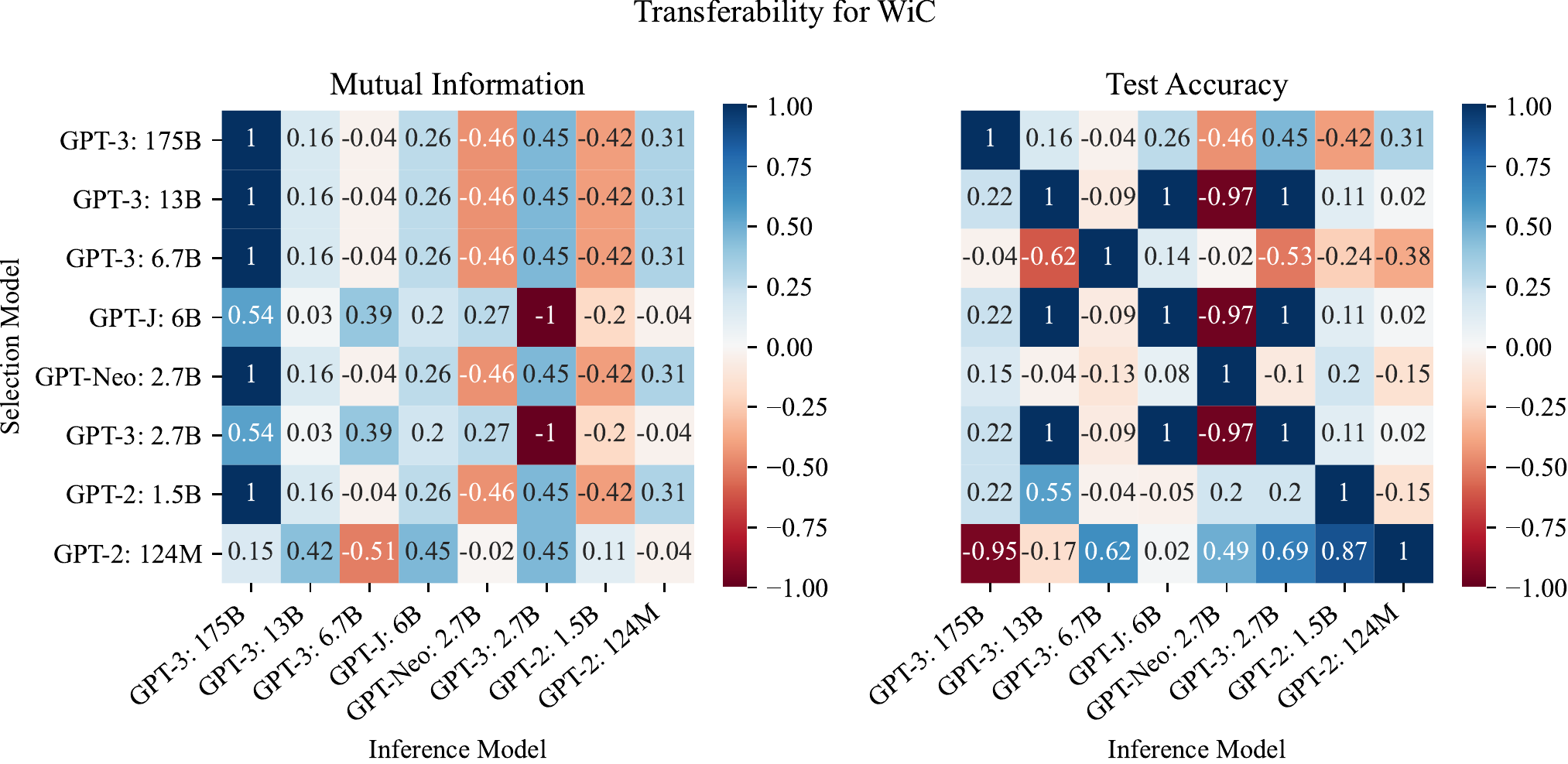}
\centering
\caption{Prompt transfer performance for WiC}
\label{fig:transfer-squad}
\end{figure*}

\clearpage

\section{Template Examples}
\label{appendix:template-examples}

\definecolor{themered}{RGB}{239, 60, 45}
\definecolor{themeblue}{RGB}{157, 206, 226}

\definecolor{myblue}{RGB}{76, 146, 195}
\definecolor{mygreen}{RGB}{86, 179, 86}
\definecolor{myred}{RGB}{222, 82, 23}
\definecolor{mypurple}{RGB}{169, 133, 202}
\definecolor{myorange}{RGB}{254, 153, 62}
\newcommand{\colorx}{\textcolor{myblue}}
\newcommand{\colorenu}{\textcolor{myorange}}
\newcommand{\colorexe}{\textcolor{mygreen}}
\newcommand{\colorcla}{\textcolor{myred}}

\newcommand{\figurefont}{8}

\DeclareRobustCommand{\hlx}[1]{{\sethlcolor{themeblue}\hl{#1}}}
\DeclareRobustCommand{\hly}[1]{{\sethlcolor{themered}\hl{#1}}}

The following are example template $f_\theta$s provided for each dataset. We include all used templates, ordered by accuracy. \hlx{In blue}, we highlight the data that is filled in from $X$; \hly{in red}, we highlight the area where we ask the model to predict the next token; everything that is not highlighted is static from instance to instance. We also include the token sets used in the collapsing functions.

\subsection{SQuAD}
\begin{minipage}{.95\linewidth}
 \textbf{Prompt 1 (MI: 4.950, Acc: 0.820):}
\fbox{ \parbox{\textwidth}{ \fontsize{\figurefont}{\figurefont}\selectfont
TASK: Answer the questions below using the phrasing from the context.\\\\CONTEXT:\\\hlx{As of the census of 2000, there were 197,790 people, 84,549 households, and 43,627 families residing in the city. The population density was 3,292.6 people per square mile (1,271.3/km). There were 92,282 housing units at an average density of 1,536.2 per square mile (593.1/km). The racial makeup of the city was 38.3\% White, 57.2\% African American, 0.2\% Native American, 1.3\% Asian, 0.1\% Pacific Islander, 1.5\% from other races, and 1.5\% from two or more races. Hispanic or Latino of any race were 2.6\% of the population.}\\\\QUESTIONS:\\1) \hlx{In 2000, how many families lived in Richmond?}\\Answer: "\hlx{43,627}"\\\\2) \hlx{What percentage of the Richmond population of 2000 was Pacific Islander?}\\Answer: "\hly{ }
}}\end{minipage} \parbox{\textwidth}{\textbf{Collapsing token sets:} None, all tokens are \newline considered\newline}
\newpage
\noindent
\begin{minipage}{.95\linewidth}
\textbf{Prompt 2 (MI: 4.965, Acc: 0.800):}
\fbox{ \parbox{\textwidth}{ \fontsize{\figurefont}{\figurefont}\selectfont
Given the following passages and questions, provide a brief, correct answer from the text.\\\\"BYU students arrive with superb preparation. The entering class has an average high school GPA of 3.71 (on a 4.0 scale) and an average ACT score that ranks in the 89th percentile nationally. The University consistently places in the top 20 for enrollment of National Merit Scholars.", "What high school GPA for BYU freshmen have on average?" -> "3.71""BYU students arrive with superb preparation. The entering class has an average high school GPA of 3.71 (on a 4.0 scale) and an average ACT score that ranks in the 89th percentile nationally. The University consistently places in the top 20 for enrollment of National Merit Scholars.", "What high school GPA for BYU freshmen have on average?" -> "3.71"\\"In meteorology, precipitation is any product of the condensation of atmospheric water vapor that falls under gravity. The main forms of precipitation include drizzle, rain, sleed, snow, graupel, and hail... Precipitation forms as smaller droplets coalesce via collision with other rain drops or ice crystals within a cloud. Short, intense periods of rain in scattered locations are called"showers".", "What causes precipitation to fall?" -> "gravity"\\"\hlx{As of the census of 2000, there were 197,790 people, 84,549 households, and 43,627 families residing in the city. The population density was 3,292.6 people per square mile (1,271.3/km). There were 92,282 housing units at an average density of 1,536.2 per square mile (593.1/km). The racial makeup of the city was 38.3\% White, 57.2\% African American, 0.2\% Native American, 1.3\% Asian, 0.1\% Pacific Islander, 1.5\% from other races, and 1.5\% from two or more races. Hispanic or Latino of any race were 2.6\% of the population.}", "\hlx{What percentage of the Richmond population of 2000 was Pacific Islander?}" -> "\hly{ }
}}\end{minipage} \parbox{\textwidth}{\textbf{Collapsing token sets:} None, all tokens are \newline considered\newline}
\begin{minipage}{.95\linewidth}
\textbf{Prompt 3 (MI: 4.965, Acc: 0.800):}
\fbox{ \parbox{\textwidth}{ \fontsize{\figurefont}{\figurefont}\selectfont
Given the following passages and questions, provide a brief, correct answer from the text.\\\\"BYU students arrive with superb preparation. The entering class has an average high school GPA of 3.71 (on a 4.0 scale) and an average ACT score that ranks in the 89th percentile nationally. The University consistently places in the top 20 for enrollment of National Merit Scholars.", "What high school GPA for BYU freshmen have on average?" -> "3.71""BYU students arrive with superb preparation. The entering class has an average high school GPA of 3.71 (on a 4.0 scale) and an average ACT score that ranks in the 89th percentile nationally. The University consistently places in the top 20 for enrollment of National Merit Scholars.", "What high school GPA for BYU freshmen have on average?" -> "3.71"\\"In meteorology, precipitation is any product of the condensation of atmospheric water vapor that falls under gravity. The main forms of precipitation include drizzle, rain, sleed, snow, graupel, and hail... Precipitation forms as smaller droplets coalesce via collision with other rain drops or ice crystals within a cloud. Short, intense periods of rain in scattered locations are called"showers".", "What causes precipitation to fall?" -> "gravity"\\"\hlx{As of the census of 2000, there were 197,790 people, 84,549 households, and 43,627 families residing in the city. The population density was 3,292.6 people per square mile (1,271.3/km). There were 92,282 housing units at an average density of 1,536.2 per square mile (593.1/km). The racial makeup of the city was 38.3\% White, 57.2\% African American, 0.2\% Native American, 1.3\% Asian, 0.1\% Pacific Islander, 1.5\% from other races, and 1.5\% from two or more races. Hispanic or Latino of any race were 2.6\% of the population.}", "\hlx{What percentage of the Richmond population of 2000 was Pacific Islander?}" -> "\hly{ }
}}\end{minipage} \parbox{\textwidth}{\textbf{Collapsing token sets:} None, all tokens are \newline considered\newline}
\begin{minipage}{.95\linewidth}
\textbf{Prompt 4 (MI: 4.901, Acc: 0.790):}
\fbox{ \parbox{\textwidth}{ \fontsize{\figurefont}{\figurefont}\selectfont
TASK: Answer the questions below using the phrasing from the context.\\\\CONTEXT:\\BYU students arrive with superb preparation. The entering class has an average high school GPA of 3.71 (on a 4.0 scale) and an average ACT score that ranks in the 89th percentile nationally. The University consistently places in the top 20 for enrollment of National Merit Scholars.\\QUESTIONS:\\1) What high school GPA for BYU freshmen have on average?\\Answer: "3.71"\\\\\\CONTEXT:\\\hlx{As of the census of 2000, there were 197,790 people, 84,549 households, and 43,627 families residing in the city. The population density was 3,292.6 people per square mile (1,271.3/km). There were 92,282 housing units at an average density of 1,536.2 per square mile (593.1/km). The racial makeup of the city was 38.3\% White, 57.2\% African American, 0.2\% Native American, 1.3\% Asian, 0.1\% Pacific Islander, 1.5\% from other races, and 1.5\% from two or more races. Hispanic or Latino of any race were 2.6\% of the population.}\\\\QUESTIONS:\\1) \hlx{What percentage of the Richmond population of 2000 was Pacific Islander?}\\Answer: "\hly{ }
}}\end{minipage} \parbox{\textwidth}{\textbf{Collapsing token sets:} None, all tokens are \newline considered\newline}
\begin{minipage}{.95\linewidth}
\textbf{Prompt 5 (MI: 4.711, Acc: 0.758):}
\fbox{ \parbox{\textwidth}{ \fontsize{\figurefont}{\figurefont}\selectfont
P1: \hlx{As of the census of 2000, there were 197,790 people, 84,549 households, and 43,627 families residing in the city. The population density was 3,292.6 people per square mile (1,271.3/km). There were 92,282 housing units at an average density of 1,536.2 per square mile (593.1/km). The racial makeup of the city was 38.3\% White, 57.2\% African American, 0.2\% Native American, 1.3\% Asian, 0.1\% Pacific Islander, 1.5\% from other races, and 1.5\% from two or more races. Hispanic or Latino of any race were 2.6\% of the population.}\\P2: \hlx{In 2000, how many families lived in Richmond?}\\P1: \hlx{43,627}\\P2: \hlx{What percentage of the Richmond population of 2000 was Pacific Islander?}\\P1:\hly{ }
}}\end{minipage} \parbox{\textwidth}{\textbf{Collapsing token sets:} None, all tokens are \newline considered\newline}
\newpage
\noindent
\begin{minipage}{.95\linewidth}
\textbf{Prompt 6 (MI: 5.224, Acc: 0.754):}
\fbox{ \parbox{\textwidth}{ \fontsize{\figurefont}{\figurefont}\selectfont
CHAPTER QUIZ\\\\PASSAGE:\\\hlx{As of the census of 2000, there were 197,790 people, 84,549 households, and 43,627 families residing in the city. The population density was 3,292.6 people per square mile (1,271.3/km). There were 92,282 housing units at an average density of 1,536.2 per square mile (593.1/km). The racial makeup of the city was 38.3\% White, 57.2\% African American, 0.2\% Native American, 1.3\% Asian, 0.1\% Pacific Islander, 1.5\% from other races, and 1.5\% from two or more races. Hispanic or Latino of any race were 2.6\% of the population.}\\\\QUESTIONS:\\1) \hlx{In 2000, how many families lived in Richmond?}\\2) \hlx{What percentage of the Richmond population of 2000 was Pacific Islander?}\\\\ANSWER KEY:\\1) \hlx{43,627}\\2)\hly{ }
}}\end{minipage} \parbox{\textwidth}{\textbf{Collapsing token sets:} None, all tokens are \newline considered\newline}
\begin{minipage}{.95\linewidth}
\textbf{Prompt 7 (MI: 5.126, Acc: 0.750):}
\fbox{ \parbox{\textwidth}{ \fontsize{\figurefont}{\figurefont}\selectfont
CHAPTER QUIZ\\PASSAGE: BYU students arrive with superb preparation. The entering class has an average high school GPA of 3.71 (on a 4.0 scale) and an average ACT score that ranks in the 89th percentile nationally. The University consistently places in the top 20 for enrollment of National Merit Scholars.\\QUESTIONS:\\1) What high school GPA for BYU freshmen have on average?\\\\ANSWER KEY: \\1) 3.71\\\\CHAPTER QUIZ\\\\PASSAGE:\\\hlx{As of the census of 2000, there were 197,790 people, 84,549 households, and 43,627 families residing in the city. The population density was 3,292.6 people per square mile (1,271.3/km). There were 92,282 housing units at an average density of 1,536.2 per square mile (593.1/km). The racial makeup of the city was 38.3\% White, 57.2\% African American, 0.2\% Native American, 1.3\% Asian, 0.1\% Pacific Islander, 1.5\% from other races, and 1.5\% from two or more races. Hispanic or Latino of any race were 2.6\% of the population.}\\\\QUESTIONS:\\1) \hlx{What percentage of the Richmond population of 2000 was Pacific Islander?}\\\\ANSWER KEY:\\1)\hly{ }
}}\end{minipage} \parbox{\textwidth}{\textbf{Collapsing token sets:} None, all tokens are \newline considered\newline}
\begin{minipage}{.95\linewidth}
\textbf{Prompt 8 (MI: 4.745, Acc: 0.700):}
\fbox{ \parbox{\textwidth}{ \fontsize{\figurefont}{\figurefont}\selectfont
P1: BYU students arrive with superb preparation. The entering class has an average high school GPA of 3.71 (on a 4.0 scale) and an average ACT score that ranks in the 89th percentile nationally. The University consistently places in the top 20 for enrollment of National Merit Scholars.\\P2: What high school GPA for BYU freshmen have on average?\\P1: 3.71\\\\\\P1: \hlx{As of the census of 2000, there were 197,790 people, 84,549 households, and 43,627 families residing in the city. The population density was 3,292.6 people per square mile (1,271.3/km). There were 92,282 housing units at an average density of 1,536.2 per square mile (593.1/km). The racial makeup of the city was 38.3\% White, 57.2\% African American, 0.2\% Native American, 1.3\% Asian, 0.1\% Pacific Islander, 1.5\% from other races, and 1.5\% from two or more races. Hispanic or Latino of any race were 2.6\% of the population.}\\P2: \hlx{What percentage of the Richmond population of 2000 was Pacific Islander?}\\P1:\hly{ }
}}\end{minipage} \parbox{\textwidth}{\textbf{Collapsing token sets:} None, all tokens are \newline considered\newline}
\begin{minipage}{.95\linewidth}
\textbf{Prompt 9 (MI: 3.998, Acc: 0.692):}
\fbox{ \parbox{\textwidth}{ \fontsize{\figurefont}{\figurefont}\selectfont
CHAPTER QUIZ\\\\PASSAGE:\\\hlx{As of the census of 2000, there were 197,790 people, 84,549 households, and 43,627 families residing in the city. The population density was 3,292.6 people per square mile (1,271.3/km). There were 92,282 housing units at an average density of 1,536.2 per square mile (593.1/km). The racial makeup of the city was 38.3\% White, 57.2\% African American, 0.2\% Native American, 1.3\% Asian, 0.1\% Pacific Islander, 1.5\% from other races, and 1.5\% from two or more races. Hispanic or Latino of any race were 2.6\% of the population.}\\\\QUESTIONS:\\1) \hlx{What percentage of the Richmond population of 2000 was Pacific Islander?}\\\\ANSWER KEY:\\1)\hly{ }
}}\end{minipage} \parbox{\textwidth}{\textbf{Collapsing token sets:} None, all tokens are \newline considered\newline}
\begin{minipage}{.95\linewidth}
\textbf{Prompt 10 (MI: 4.037, Acc: 0.686):}
\fbox{ \parbox{\textwidth}{ \fontsize{\figurefont}{\figurefont}\selectfont
TASK: Using words from the CONTEXT, answer the below QUESTIONS.\\\\CONTEXT:\\\hlx{As of the census of 2000, there were 197,790 people, 84,549 households, and 43,627 families residing in the city. The population density was 3,292.6 people per square mile (1,271.3/km). There were 92,282 housing units at an average density of 1,536.2 per square mile (593.1/km). The racial makeup of the city was 38.3\% White, 57.2\% African American, 0.2\% Native American, 1.3\% Asian, 0.1\% Pacific Islander, 1.5\% from other races, and 1.5\% from two or more races. Hispanic or Latino of any race were 2.6\% of the population.}\\\\QUESTIONS:\\1) \hlx{What percentage of the Richmond population of 2000 was Pacific Islander?}\\Answer: "\hly{ }
}}\end{minipage} \parbox{\textwidth}{\textbf{Collapsing token sets:} None, all tokens are \newline considered\newline}
\begin{minipage}{.95\linewidth}
\textbf{Prompt 11 (MI: 4.231, Acc: 0.684):}
\fbox{ \parbox{\textwidth}{ \fontsize{\figurefont}{\figurefont}\selectfont
P1 tells P2 some information, P2 asks comprehension questions, and P1 answers.\\\\P1: \hlx{As of the census of 2000, there were 197,790 people, 84,549 households, and 43,627 families residing in the city. The population density was 3,292.6 people per square mile (1,271.3/km). There were 92,282 housing units at an average density of 1,536.2 per square mile (593.1/km). The racial makeup of the city was 38.3\% White, 57.2\% African American, 0.2\% Native American, 1.3\% Asian, 0.1\% Pacific Islander, 1.5\% from other races, and 1.5\% from two or more races. Hispanic or Latino of any race were 2.6\% of the population.}\\P2: \hlx{What percentage of the Richmond population of 2000 was Pacific Islander?}\\P1: The answer is "\hly{ }
}}\end{minipage} \parbox{\textwidth}{\textbf{Collapsing token sets:} None, all tokens are \newline considered\newline}
\begin{minipage}{.95\linewidth}
\textbf{Prompt 12 (MI: 3.568, Acc: 0.620):}
\fbox{ \parbox{\textwidth}{ \fontsize{\figurefont}{\figurefont}\selectfont
P1: \hlx{As of the census of 2000, there were 197,790 people, 84,549 households, and 43,627 families residing in the city. The population density was 3,292.6 people per square mile (1,271.3/km). There were 92,282 housing units at an average density of 1,536.2 per square mile (593.1/km). The racial makeup of the city was 38.3\% White, 57.2\% African American, 0.2\% Native American, 1.3\% Asian, 0.1\% Pacific Islander, 1.5\% from other races, and 1.5\% from two or more races. Hispanic or Latino of any race were 2.6\% of the population.}\\P2: \hlx{What percentage of the Richmond population of 2000 was Pacific Islander?}\\P1: The answer is "\hly{ }
}}\end{minipage} \parbox{\textwidth}{\textbf{Collapsing token sets:} None, all tokens are \newline considered\newline}
\begin{minipage}{.95\linewidth}
\textbf{Prompt 13 (MI: 3.261, Acc: 0.614):}
\fbox{ \parbox{\textwidth}{ \fontsize{\figurefont}{\figurefont}\selectfont
Given the following passages and questions, provide a brief, correct answer from the text.\\"\hlx{As of the census of 2000, there were 197,790 people, 84,549 households, and 43,627 families residing in the city. The population density was 3,292.6 people per square mile (1,271.3/km). There were 92,282 housing units at an average density of 1,536.2 per square mile (593.1/km). The racial makeup of the city was 38.3\% White, 57.2\% African American, 0.2\% Native American, 1.3\% Asian, 0.1\% Pacific Islander, 1.5\% from other races, and 1.5\% from two or more races. Hispanic or Latino of any race were 2.6\% of the population.}", "\hlx{What percentage of the Richmond population of 2000 was Pacific Islander?}" -> "\hly{ }
}}\end{minipage} \parbox{\textwidth}{\textbf{Collapsing token sets:} None, all tokens are \newline considered\newline}
\newpage
\noindent
\begin{minipage}{.95\linewidth}
\textbf{Prompt 14 (MI: 3.760, Acc: 0.608):}
\fbox{ \parbox{\textwidth}{ \fontsize{\figurefont}{\figurefont}\selectfont
\hlx{As of the census of 2000, there were 197,790 people, 84,549 households, and 43,627 families residing in the city. The population density was 3,292.6 people per square mile (1,271.3/km). There were 92,282 housing units at an average density of 1,536.2 per square mile (593.1/km). The racial makeup of the city was 38.3\% White, 57.2\% African American, 0.2\% Native American, 1.3\% Asian, 0.1\% Pacific Islander, 1.5\% from other races, and 1.5\% from two or more races. Hispanic or Latino of any race were 2.6\% of the population.}\\\\\hlx{What percentage of the Richmond population of 2000 was Pacific Islander?}\\The correct answer is:\hly{ }
}}\end{minipage} \parbox{\textwidth}{\textbf{Collapsing token sets:} None, all tokens are \newline considered\newline}
\begin{minipage}{.95\linewidth}
\textbf{Prompt 15 (MI: 3.006, Acc: 0.606):}
\fbox{ \parbox{\textwidth}{ \fontsize{\figurefont}{\figurefont}\selectfont
I read this in a book today:\\\hlx{As of the census of 2000, there were 197,790 people, 84,549 households, and 43,627 families residing in the city. The population density was 3,292.6 people per square mile (1,271.3/km). There were 92,282 housing units at an average density of 1,536.2 per square mile (593.1/km). The racial makeup of the city was 38.3\% White, 57.2\% African American, 0.2\% Native American, 1.3\% Asian, 0.1\% Pacific Islander, 1.5\% from other races, and 1.5\% from two or more races. Hispanic or Latino of any race were 2.6\% of the population.}\\From that context, did you catch \hlx{What percentage of the Richmond population of 2000 was Pacific Islander?}\\Yes, the answer is\hly{ }
}}\end{minipage} \parbox{\textwidth}{\textbf{Collapsing token sets:} None, all tokens are \newline considered\newline}
\begin{minipage}{.95\linewidth}
\textbf{Prompt 16 (MI: 3.843, Acc: 0.592):}
\fbox{ \parbox{\textwidth}{ \fontsize{\figurefont}{\figurefont}\selectfont
TASK: Answer the questions below using the phrasing from the context.\\\\CONTEXT:\\\hlx{As of the census of 2000, there were 197,790 people, 84,549 households, and 43,627 families residing in the city. The population density was 3,292.6 people per square mile (1,271.3/km). There were 92,282 housing units at an average density of 1,536.2 per square mile (593.1/km). The racial makeup of the city was 38.3\% White, 57.2\% African American, 0.2\% Native American, 1.3\% Asian, 0.1\% Pacific Islander, 1.5\% from other races, and 1.5\% from two or more races. Hispanic or Latino of any race were 2.6\% of the population.}\\\\QUESTIONS:\\1) \hlx{What percentage of the Richmond population of 2000 was Pacific Islander?}\\Answer: "\hly{ }
}}\end{minipage} \parbox{\textwidth}{\textbf{Collapsing token sets:} None, all tokens are \newline considered\newline}
\begin{minipage}{.95\linewidth}
\textbf{Prompt 17 (MI: 3.508, Acc: 0.544):}
\fbox{ \parbox{\textwidth}{ \fontsize{\figurefont}{\figurefont}\selectfont
I read this in a book today:\\\hlx{As of the census of 2000, there were 197,790 people, 84,549 households, and 43,627 families residing in the city. The population density was 3,292.6 people per square mile (1,271.3/km). There were 92,282 housing units at an average density of 1,536.2 per square mile (593.1/km). The racial makeup of the city was 38.3\% White, 57.2\% African American, 0.2\% Native American, 1.3\% Asian, 0.1\% Pacific Islander, 1.5\% from other races, and 1.5\% from two or more races. Hispanic or Latino of any race were 2.6\% of the population.}\\\hlx{What percentage of the Richmond population of 2000 was Pacific Islander?}\\Answer:\hly{ }
}}\end{minipage} \parbox{\textwidth}{\textbf{Collapsing token sets:} None, all tokens are \newline considered\newline}
\begin{minipage}{.95\linewidth}
\textbf{Prompt 18 (MI: 3.227, Acc: 0.406):}
\fbox{ \parbox{\textwidth}{ \fontsize{\figurefont}{\figurefont}\selectfont
Context: \hlx{As of the census of 2000, there were 197,790 people, 84,549 households, and 43,627 families residing in the city. The population density was 3,292.6 people per square mile (1,271.3/km). There were 92,282 housing units at an average density of 1,536.2 per square mile (593.1/km). The racial makeup of the city was 38.3\% White, 57.2\% African American, 0.2\% Native American, 1.3\% Asian, 0.1\% Pacific Islander, 1.5\% from other races, and 1.5\% from two or more races. Hispanic or Latino of any race were 2.6\% of the population.}\\\\Q: \hlx{What percentage of the Richmond population of 2000 was Pacific Islander?}\\\\A:\hly{ }
}}\end{minipage} \parbox{\textwidth}{\textbf{Collapsing token sets:} None, all tokens are \newline considered\newline}
\begin{minipage}{.95\linewidth}
\textbf{Prompt 19 (MI: 2.497, Acc: 0.402):}
\fbox{ \parbox{\textwidth}{ \fontsize{\figurefont}{\figurefont}\selectfont
A friend of mine told me this:\\\hlx{As of the census of 2000, there were 197,790 people, 84,549 households, and 43,627 families residing in the city. The population density was 3,292.6 people per square mile (1,271.3/km). There were 92,282 housing units at an average density of 1,536.2 per square mile (593.1/km). The racial makeup of the city was 38.3\% White, 57.2\% African American, 0.2\% Native American, 1.3\% Asian, 0.1\% Pacific Islander, 1.5\% from other races, and 1.5\% from two or more races. Hispanic or Latino of any race were 2.6\% of the population.}\\My friend then asked: \hlx{What percentage of the Richmond population of 2000 was Pacific Islander?}\\I answered:\hly{ }
}}\end{minipage} \parbox{\textwidth}{\textbf{Collapsing token sets:} None, all tokens are \newline considered\newline}
\newpage
\noindent
\begin{minipage}{.95\linewidth}
\textbf{Prompt 20 (MI: 2.312, Acc: 0.302):}
\fbox{ \parbox{\textwidth}{ \fontsize{\figurefont}{\figurefont}\selectfont
ANSWER KEY:\\\\QUESTION1:\\"\hlx{As of the census of 2000, there were 197,790 people, 84,549 households, and 43,627 families residing in the city. The population density was 3,292.6 people per square mile (1,271.3/km). There were 92,282 housing units at an average density of 1,536.2 per square mile (593.1/km). The racial makeup of the city was 38.3\% White, 57.2\% African American, 0.2\% Native American, 1.3\% Asian, 0.1\% Pacific Islander, 1.5\% from other races, and 1.5\% from two or more races. Hispanic or Latino of any race were 2.6\% of the population.}" \hlx{What percentage of the Richmond population of 2000 was Pacific Islander?}\\ANSWER1:\hly{ }
}}\end{minipage} \parbox{\textwidth}{\textbf{Collapsing token sets:} None, all tokens are \newline considered\newline}
\\ \\\subsection{LAMBADA}
\begin{minipage}{.95\linewidth}
 \textbf{Prompt 1 (MI: 4.984, Acc: 0.782):}
\fbox{ \parbox{\textwidth}{ \fontsize{\figurefont}{\figurefont}\selectfont
Fill in blank:\\\\Alice was friends with Bob. Alice went to visit her friend \_\_\_\_. -> Bob\\\hlx{"I would speak to you privately," Bowen said, casting a glance around at the others milling about.\\\\The worry in her eyes deepened, but she nodded hesitantly and awaited Bowen's directive.\\\\He led her through the great hall, annoyance biting at him when he saw no place where people weren't congregated. He stepped outside the back of the keep, where, finally, he spied an area near the bathhouses, where it was quiet and} \_\_\_\_. ->\hly{ }
}}\end{minipage} \parbox{\textwidth}{\textbf{Collapsing token sets:} None, all tokens are \newline considered\newline}
\begin{minipage}{.95\linewidth}
\textbf{Prompt 2 (MI: 4.793, Acc: 0.770):}
\fbox{ \parbox{\textwidth}{ \fontsize{\figurefont}{\figurefont}\selectfont
Fill in blank:\\\\She held the torch in front of her.\\\\She caught her breath.\\\\"Chris? There's a step."\\\\"What?"\\\\"A step. Cut in the rock. About fifty feet ahead." She moved faster. They both moved faster. "In fact," she said, raising the torch higher, "there's more than a \_\_\_\_. -> step\\\\\hlx{"I would speak to you privately," Bowen said, casting a glance around at the others milling about.\\\\The worry in her eyes deepened, but she nodded hesitantly and awaited Bowen's directive.\\\\He led her through the great hall, annoyance biting at him when he saw no place where people weren't congregated. He stepped outside the back of the keep, where, finally, he spied an area near the bathhouses, where it was quiet and} \_\_\_\_. ->\hly{ }
}}\end{minipage} \parbox{\textwidth}{\textbf{Collapsing token sets:} None, all tokens are \newline considered\newline}
\newline
\newline
\newline
\newline
\begin{minipage}{.95\linewidth}
\textbf{Prompt 3 (MI: 5.062, Acc: 0.770):}
\fbox{ \parbox{\textwidth}{ \fontsize{\figurefont}{\figurefont}\selectfont
Fill in blank:\\\\Alice was friends with Bob. Alice went to visit her friend \_\_\_\_. -> Bob\\George bought some baseball equipment, a ball, a glove, and a \_\_\_\_. -> bat\\\hlx{"I would speak to you privately," Bowen said, casting a glance around at the others milling about.\\\\The worry in her eyes deepened, but she nodded hesitantly and awaited Bowen's directive.\\\\He led her through the great hall, annoyance biting at him when he saw no place where people weren't congregated. He stepped outside the back of the keep, where, finally, he spied an area near the bathhouses, where it was quiet and} \_\_\_\_. ->\hly{ }
}}\end{minipage} \parbox{\textwidth}{\textbf{Collapsing token sets:} None, all tokens are \newline considered\newline}
\begin{minipage}{.95\linewidth}
\textbf{Prompt 4 (MI: 5.058, Acc: 0.736):}
\fbox{ \parbox{\textwidth}{ \fontsize{\figurefont}{\figurefont}\selectfont
\hlx{"I would speak to you privately," Bowen said, casting a glance around at the others milling about.\\\\The worry in her eyes deepened, but she nodded hesitantly and awaited Bowen's directive.\\\\He led her through the great hall, annoyance biting at him when he saw no place where people weren't congregated. He stepped outside the back of the keep, where, finally, he spied an area near the bathhouses, where it was quiet and}\hly{ }
}}\end{minipage} \parbox{\textwidth}{\textbf{Collapsing token sets:} None, all tokens are \newline considered\newline}
\begin{minipage}{.95\linewidth}
\textbf{Prompt 5 (MI: 4.194, Acc: 0.608):}
\fbox{ \parbox{\textwidth}{ \fontsize{\figurefont}{\figurefont}\selectfont
P1: I'm going to tell you a story, but leave a word out. Once I'm done telling the story, pick the word that best fits in the blank. \\I like to eat peanut butter and jelly \_\_\_\_.\\P2: sandwiches\\P1: I'm going to tell you a story, but leave a word out. Once I'm done telling the story, pick the word that best fits in the blank. \\\hlx{"I would speak to you privately," Bowen said, casting a glance around at the others milling about.\\\\The worry in her eyes deepened, but she nodded hesitantly and awaited Bowen's directive.\\\\He led her through the great hall, annoyance biting at him when he saw no place where people weren't congregated. He stepped outside the back of the keep, where, finally, he spied an area near the bathhouses, where it was quiet and} \_\_\_\_. \\P2:\hly{ }
}}\end{minipage} \parbox{\textwidth}{\textbf{Collapsing token sets:} None, all tokens are \newline considered\newline}
\begin{minipage}{.95\linewidth}
\textbf{Prompt 6 (MI: 4.623, Acc: 0.608):}
\fbox{ \parbox{\textwidth}{ \fontsize{\figurefont}{\figurefont}\selectfont
Fill in the blank for the following sentences.\\\\"It was a cold night. The wind was whistling around the courtyard as I stepped out of the car and into the \_\_\_\_." -> "It was a cold night. The wind was whistling around the courtyard as I stepped out of the car and into the darkness."\\"\hlx{"I would speak to you privately," Bowen said, casting a glance around at the others milling about.\\\\The worry in her eyes deepened, but she nodded hesitantly and awaited Bowen's directive.\\\\He led her through the great hall, annoyance biting at him when he saw no place where people weren't congregated. He stepped outside the back of the keep, where, finally, he spied an area near the bathhouses, where it was quiet and} \_\_\_\_." -> "\hlx{"I would speak to you privately," Bowen said, casting a glance around at the others milling about.\\\\The worry in her eyes deepened, but she nodded hesitantly and awaited Bowen's directive.\\\\He led her through the great hall, annoyance biting at him when he saw no place where people weren't congregated. He stepped outside the back of the keep, where, finally, he spied an area near the bathhouses, where it was quiet and}\hly{ }
}}\end{minipage} \parbox{\textwidth}{\textbf{Collapsing token sets:} None, all tokens are \newline considered\newline}
\begin{minipage}{.95\linewidth}
\textbf{Prompt 7 (MI: 4.328, Acc: 0.596):}
\fbox{ \parbox{\textwidth}{ \fontsize{\figurefont}{\figurefont}\selectfont
P1: I'm going to tell you a story, but leave a word out. Once I'm done telling the story, pick the word that best fits in the blank. \\It was a cold night. The wind was \_\_\_\_ around the courtyard as I stepped out of the car and into the darkness.\\P2: whistling\\P1: I'm going to tell you a story, but leave a word out. Once I'm done telling the story, pick the word that best fits in the blank. \\\hlx{"I would speak to you privately," Bowen said, casting a glance around at the others milling about.\\\\The worry in her eyes deepened, but she nodded hesitantly and awaited Bowen's directive.\\\\He led her through the great hall, annoyance biting at him when he saw no place where people weren't congregated. He stepped outside the back of the keep, where, finally, he spied an area near the bathhouses, where it was quiet and} \_\_\_\_. \\P2:\hly{ }
}}\end{minipage} \parbox{\textwidth}{\textbf{Collapsing token sets:} None, all tokens are \newline considered\newline}
\begin{minipage}{.95\linewidth}
\textbf{Prompt 8 (MI: 3.338, Acc: 0.586):}
\fbox{ \parbox{\textwidth}{ \fontsize{\figurefont}{\figurefont}\selectfont
Fill in the blank with the missing word to complete the sentence.\\\\Passage: I like to eat peanut butter and jelly \_\_\_\_.\\Missing Word: sandwiches\\\\Passage: \hlx{"I would speak to you privately," Bowen said, casting a glance around at the others milling about.\\\\The worry in her eyes deepened, but she nodded hesitantly and awaited Bowen's directive.\\\\He led her through the great hall, annoyance biting at him when he saw no place where people weren't congregated. He stepped outside the back of the keep, where, finally, he spied an area near the bathhouses, where it was quiet and} \_\_\_\_.\\Missing Word: "\hly{ }
}}\end{minipage} \parbox{\textwidth}{\textbf{Collapsing token sets:} None, all tokens are \newline considered\newline}
\begin{minipage}{.95\linewidth}
\textbf{Prompt 9 (MI: 2.230, Acc: 0.498):}
\fbox{ \parbox{\textwidth}{ \fontsize{\figurefont}{\figurefont}\selectfont
\hlx{"I would speak to you privately," Bowen said, casting a glance around at the others milling about.\\\\The worry in her eyes deepened, but she nodded hesitantly and awaited Bowen's directive.\\\\He led her through the great hall, annoyance biting at him when he saw no place where people weren't congregated. He stepped outside the back of the keep, where, finally, he spied an area near the bathhouses, where it was quiet and} \_\_\_\_. \\\\The missing word in the story should be: "\hly{ }
}}\end{minipage} \parbox{\textwidth}{\textbf{Collapsing token sets:} None, all tokens are \newline considered\newline}
\begin{minipage}{.95\linewidth}
\textbf{Prompt 10 (MI: 2.632, Acc: 0.474):}
\fbox{ \parbox{\textwidth}{ \fontsize{\figurefont}{\figurefont}\selectfont
\hlx{"I would speak to you privately," Bowen said, casting a glance around at the others milling about.\\\\The worry in her eyes deepened, but she nodded hesitantly and awaited Bowen's directive.\\\\He led her through the great hall, annoyance biting at him when he saw no place where people weren't congregated. He stepped outside the back of the keep, where, finally, he spied an area near the bathhouses, where it was quiet and} \_\_\_\_. \\Fill in the blank with the missing word or phrase.\\What is the missing word? The missing word is "\hly{ }
}}\end{minipage} \parbox{\textwidth}{\textbf{Collapsing token sets:} None, all tokens are \newline considered\newline}
\begin{minipage}{.95\linewidth}
\textbf{Prompt 11 (MI: 4.549, Acc: 0.470):}
\fbox{ \parbox{\textwidth}{ \fontsize{\figurefont}{\figurefont}\selectfont
It was a cold night. The wind was \_\_\_\_ around the courtyard as I stepped out of the car and into the darkness.\\Word: whistling\\\\\hlx{"I would speak to you privately," Bowen said, casting a glance around at the others milling about.\\\\The worry in her eyes deepened, but she nodded hesitantly and awaited Bowen's directive.\\\\He led her through the great hall, annoyance biting at him when he saw no place where people weren't congregated. He stepped outside the back of the keep, where, finally, he spied an area near the bathhouses, where it was quiet and} \_\_\_\_. \\\\Word:\hly{ }
}}\end{minipage} \parbox{\textwidth}{\textbf{Collapsing token sets:} None, all tokens are \newline considered\newline}
\begin{minipage}{.95\linewidth}
\textbf{Prompt 12 (MI: 2.637, Acc: 0.454):}
\fbox{ \parbox{\textwidth}{ \fontsize{\figurefont}{\figurefont}\selectfont
P1: I'm going to tell you a story, but leave a word out. Once I'm done telling the story, pick the word that best fits in the blank. \\\hlx{"I would speak to you privately," Bowen said, casting a glance around at the others milling about.\\\\The worry in her eyes deepened, but she nodded hesitantly and awaited Bowen's directive.\\\\He led her through the great hall, annoyance biting at him when he saw no place where people weren't congregated. He stepped outside the back of the keep, where, finally, he spied an area near the bathhouses, where it was quiet and} \_\_\_\_. \\P2: The word which fits best is "\hly{ }
}}\end{minipage} \parbox{\textwidth}{\textbf{Collapsing token sets:} None, all tokens are \newline considered\newline}
\begin{minipage}{.95\linewidth}
\textbf{Prompt 13 (MI: 2.476, Acc: 0.434):}
\fbox{ \parbox{\textwidth}{ \fontsize{\figurefont}{\figurefont}\selectfont
\hlx{"I would speak to you privately," Bowen said, casting a glance around at the others milling about.\\\\The worry in her eyes deepened, but she nodded hesitantly and awaited Bowen's directive.\\\\He led her through the great hall, annoyance biting at him when he saw no place where people weren't congregated. He stepped outside the back of the keep, where, finally, he spied an area near the bathhouses, where it was quiet and} \_\_\_\_. \\Fill in the blank with the missing word or phrase to complete the sentence.\\What is the missing word? The missing word is "\hly{ }
}}\end{minipage} \parbox{\textwidth}{\textbf{Collapsing token sets:} None, all tokens are \newline considered\newline}
\begin{minipage}{.95\linewidth}
\textbf{Prompt 14 (MI: 3.043, Acc: 0.432):}
\fbox{ \parbox{\textwidth}{ \fontsize{\figurefont}{\figurefont}\selectfont
Read the following sentences, and try to guess which word goes in the blank.\\\hlx{"I would speak to you privately," Bowen said, casting a glance around at the others milling about.\\\\The worry in her eyes deepened, but she nodded hesitantly and awaited Bowen's directive.\\\\He led her through the great hall, annoyance biting at him when he saw no place where people weren't congregated. He stepped outside the back of the keep, where, finally, he spied an area near the bathhouses, where it was quiet and} \_\_\_\_. \\Answer: "\hly{ }
}}\end{minipage} \parbox{\textwidth}{\textbf{Collapsing token sets:} None, all tokens are \newline considered\newline}
\begin{minipage}{.95\linewidth}
\textbf{Prompt 15 (MI: 2.450, Acc: 0.428):}
\fbox{ \parbox{\textwidth}{ \fontsize{\figurefont}{\figurefont}\selectfont
Fill in blank:\\\\\hlx{"I would speak to you privately," Bowen said, casting a glance around at the others milling about.\\\\The worry in her eyes deepened, but she nodded hesitantly and awaited Bowen's directive.\\\\He led her through the great hall, annoyance biting at him when he saw no place where people weren't congregated. He stepped outside the back of the keep, where, finally, he spied an area near the bathhouses, where it was quiet and} \_\_\_\_. ->\hly{ }
}}\end{minipage} \parbox{\textwidth}{\textbf{Collapsing token sets:} None, all tokens are \newline considered\newline}
\begin{minipage}{.95\linewidth}
\textbf{Prompt 16 (MI: 2.820, Acc: 0.398):}
\fbox{ \parbox{\textwidth}{ \fontsize{\figurefont}{\figurefont}\selectfont
Fill in the blank with the missing word.\\\hlx{"I would speak to you privately," Bowen said, casting a glance around at the others milling about.\\\\The worry in her eyes deepened, but she nodded hesitantly and awaited Bowen's directive.\\\\He led her through the great hall, annoyance biting at him when he saw no place where people weren't congregated. He stepped outside the back of the keep, where, finally, he spied an area near the bathhouses, where it was quiet and} \_\_\_\_. \\Answer: "\hly{ }
}}\end{minipage} \parbox{\textwidth}{\textbf{Collapsing token sets:} None, all tokens are \newline considered\newline}
\begin{minipage}{.95\linewidth}
\textbf{Prompt 17 (MI: 1.931, Acc: 0.376):}
\fbox{ \parbox{\textwidth}{ \fontsize{\figurefont}{\figurefont}\selectfont
\hlx{"I would speak to you privately," Bowen said, casting a glance around at the others milling about.\\\\The worry in her eyes deepened, but she nodded hesitantly and awaited Bowen's directive.\\\\He led her through the great hall, annoyance biting at him when he saw no place where people weren't congregated. He stepped outside the back of the keep, where, finally, he spied an area near the bathhouses, where it was quiet and} \_\_\_\_. \\Which word should we put in the blank to complete the story? Let's use the word "\hly{ }
}}\end{minipage} \parbox{\textwidth}{\textbf{Collapsing token sets:} None, all tokens are \newline considered\newline}
\begin{minipage}{.95\linewidth}
\textbf{Prompt 18 (MI: 2.530, Acc: 0.374):}
\fbox{ \parbox{\textwidth}{ \fontsize{\figurefont}{\figurefont}\selectfont
P1: What word do you think fits best in the following story? \\\hlx{"I would speak to you privately," Bowen said, casting a glance around at the others milling about.\\\\The worry in her eyes deepened, but she nodded hesitantly and awaited Bowen's directive.\\\\He led her through the great hall, annoyance biting at him when he saw no place where people weren't congregated. He stepped outside the back of the keep, where, finally, he spied an area near the bathhouses, where it was quiet and} \_\_\_\_. \\P2: The word which fits best is "\hly{ }
}}\end{minipage} \parbox{\textwidth}{\textbf{Collapsing token sets:} None, all tokens are \newline considered\newline}
\begin{minipage}{.95\linewidth}
\textbf{Prompt 19 (MI: 2.372, Acc: 0.364):}
\fbox{ \parbox{\textwidth}{ \fontsize{\figurefont}{\figurefont}\selectfont
\hlx{"I would speak to you privately," Bowen said, casting a glance around at the others milling about.\\\\The worry in her eyes deepened, but she nodded hesitantly and awaited Bowen's directive.\\\\He led her through the great hall, annoyance biting at him when he saw no place where people weren't congregated. He stepped outside the back of the keep, where, finally, he spied an area near the bathhouses, where it was quiet and} \_\_\_\_. \\Which word fills in the blank best?\\The word that fills in the blank best is "\hly{ }
}}\end{minipage} \parbox{\textwidth}{\textbf{Collapsing token sets:} None, all tokens are \newline considered\newline}
\begin{minipage}{.95\linewidth}
\textbf{Prompt 20 (MI: 2.860, Acc: 0.296):}
\fbox{ \parbox{\textwidth}{ \fontsize{\figurefont}{\figurefont}\selectfont
Pick the best word to replace the blank.\\Story: \hlx{"I would speak to you privately," Bowen said, casting a glance around at the others milling about.\\\\The worry in her eyes deepened, but she nodded hesitantly and awaited Bowen's directive.\\\\He led her through the great hall, annoyance biting at him when he saw no place where people weren't congregated. He stepped outside the back of the keep, where, finally, he spied an area near the bathhouses, where it was quiet and} \_\_\_\_. \\Answer: "\hly{ }
}}\end{minipage} \parbox{\textwidth}{\textbf{Collapsing token sets:} None, all tokens are \newline considered\newline}
\\ \\\subsection{ROCStories}
\begin{minipage}{.95\linewidth}
 \textbf{Prompt 1 (MI: 3.859, Acc: 0.538):}
\fbox{ \parbox{\textwidth}{ \fontsize{\figurefont}{\figurefont}\selectfont
Fill in the blank for the following sentences.\\\\"\hlx{Marissa loved \_\_\_\_\_ pokemon go game. It is the biggest thing right now. She had done so much more walking since she started playing it. She walked all day and evening sometimes. She walked almost 10 miles in two days.}" -> \hlx{"Marissa loved}\hly{ }
}}\end{minipage} \parbox{\textwidth}{\textbf{Collapsing token sets:} None, all tokens are \newline considered\newline}
\begin{minipage}{.95\linewidth}
\textbf{Prompt 2 (MI: 4.427, Acc: 0.524):}
\fbox{ \parbox{\textwidth}{ \fontsize{\figurefont}{\figurefont}\selectfont
Fill in the blank for the following sentences.\\\\"It was a cold night. The wind was \_\_\_\_\_ around the courtyard as I stepped out of the car and into the darkness." -> "It was a cold night. The wind was whistling around the courtyard as I stepped out of the car and into the darkness."\\"\hlx{Marissa loved \_\_\_\_\_ pokemon go game. It is the biggest thing right now. She had done so much more walking since she started playing it. She walked all day and evening sometimes. She walked almost 10 miles in two days.}" -> \hlx{"Marissa loved}\hly{ }
}}\end{minipage} \parbox{\textwidth}{\textbf{Collapsing token sets:} None, all tokens are \newline considered\newline}
\begin{minipage}{.95\linewidth}
\textbf{Prompt 3 (MI: 3.728, Acc: 0.420):}
\fbox{ \parbox{\textwidth}{ \fontsize{\figurefont}{\figurefont}\selectfont
\hlx{Poke GO!}\\\hlx{\\Marissa loved}\hly{ }
}}\end{minipage} \parbox{\textwidth}{\textbf{Collapsing token sets:} None, all tokens are \newline considered\newline}
\begin{minipage}{.95\linewidth}
\textbf{Prompt 4 (MI: 3.670, Acc: 0.356):}
\fbox{ \parbox{\textwidth}{ \fontsize{\figurefont}{\figurefont}\selectfont
Fill in the blank with the missing word or phrase to complete the sentence.\\\\I like to eat \_\_\_\_\_ and jelly sandwiches.\\Answer: peanut butter\\\\\hlx{Marissa loved \_\_\_\_\_ pokemon go game. It is the biggest thing right now. She had done so much more walking since she started playing it. She walked all day and evening sometimes. She walked almost 10 miles in two days.} \\Answer:\hly{ }
}}\end{minipage} \parbox{\textwidth}{\textbf{Collapsing token sets:} None, all tokens are \newline considered\newline}
\begin{minipage}{.95\linewidth}
\textbf{Prompt 5 (MI: 3.904, Acc: 0.310):}
\fbox{ \parbox{\textwidth}{ \fontsize{\figurefont}{\figurefont}\selectfont
Fill in the blank with the missing word or phrase.\\\\Sentence: I like to eat \_\_\_\_\_\_\_ and jelly sandwiches.\\Missing Word/Phrase: peanut butter\\\\Sentence: \hlx{Marissa loved \_\_\_\_\_ pokemon go game. It is the biggest thing right now. She had done so much more walking since she started playing it. She walked all day and evening sometimes. She walked almost 10 miles in two days.}\\Missing Word/Phrase:\hly{ }
}}\end{minipage} \parbox{\textwidth}{\textbf{Collapsing token sets:} None, all tokens are \newline considered\newline}
\newline
\newline
\newline
\newline
\newline
\newline
\begin{minipage}{.95\linewidth}
\textbf{Prompt 6 (MI: 4.167, Acc: 0.298):}
\fbox{ \parbox{\textwidth}{ \fontsize{\figurefont}{\figurefont}\selectfont
P1: I'm going to tell you a story, but leave a word out. Once I'm done telling the story, pick the word that best fits in the blank. \\It was a cold night. The wind was \_\_\_\_\_ around the courtyard as I stepped out of the car and into the darkness.\\P2: whistling\\P1: I'm going to tell you a story, but leave a word out. Once I'm done telling the story, pick the word that best fits in the blank. \\\hlx{Marissa loved \_\_\_\_\_ pokemon go game. It is the biggest thing right now. She had done so much more walking since she started playing it. She walked all day and evening sometimes. She walked almost 10 miles in two days.} \\P2:\hly{ }
}}\end{minipage} \parbox{\textwidth}{\textbf{Collapsing token sets:} None, all tokens are \newline considered\newline}
\begin{minipage}{.95\linewidth}
\textbf{Prompt 7 (MI: 4.066, Acc: 0.290):}
\fbox{ \parbox{\textwidth}{ \fontsize{\figurefont}{\figurefont}\selectfont
P1: I'm going to tell you a story, but leave a word out. Once I'm done telling the story, pick the word that best fits in the blank. \\I like to eat \_\_\_\_\_ and jelly sandwiches.\\P2: peanut butter\\P1: I'm going to tell you a story, but leave a word out. Once I'm done telling the story, pick the word that best fits in the blank. \\\hlx{Marissa loved \_\_\_\_\_ pokemon go game. It is the biggest thing right now. She had done so much more walking since she started playing it. She walked all day and evening sometimes. She walked almost 10 miles in two days.} \\P2:\hly{ }
}}\end{minipage} \parbox{\textwidth}{\textbf{Collapsing token sets:} None, all tokens are \newline considered\newline}
\begin{minipage}{.95\linewidth}
\textbf{Prompt 8 (MI: 3.707, Acc: 0.258):}
\fbox{ \parbox{\textwidth}{ \fontsize{\figurefont}{\figurefont}\selectfont
Guess the word in the blank to complete the story.\\Story: \hlx{Marissa loved \_\_\_\_\_ pokemon go game. It is the biggest thing right now. She had done so much more walking since she started playing it. She walked all day and evening sometimes. She walked almost 10 miles in two days.} \\Answer:\hly{ }
}}\end{minipage} \parbox{\textwidth}{\textbf{Collapsing token sets:} None, all tokens are \newline considered\newline}
\begin{minipage}{.95\linewidth}
\textbf{Prompt 9 (MI: 3.644, Acc: 0.256):}
\fbox{ \parbox{\textwidth}{ \fontsize{\figurefont}{\figurefont}\selectfont
Pick the best word to replace the blank.\\Story: \hlx{Marissa loved \_\_\_\_\_ pokemon go game. It is the biggest thing right now. She had done so much more walking since she started playing it. She walked all day and evening sometimes. She walked almost 10 miles in two days.} \\Answer:\hly{ }
}}\end{minipage} \parbox{\textwidth}{\textbf{Collapsing token sets:} None, all tokens are \newline considered\newline}
\begin{minipage}{.95\linewidth}
\textbf{Prompt 10 (MI: 1.979, Acc: 0.222):}
\fbox{ \parbox{\textwidth}{ \fontsize{\figurefont}{\figurefont}\selectfont
\hlx{Marissa loved \_\_\_\_\_ pokemon go game. It is the biggest thing right now. She had done so much more walking since she started playing it. She walked all day and evening sometimes. She walked almost 10 miles in two days.} \\Fill in the blank with the missing word or phrase.\\What is the missing word? The missing word is "\hly{ }
}}\end{minipage} \parbox{\textwidth}{\textbf{Collapsing token sets:} None, all tokens are \newline considered\newline}
\begin{minipage}{.95\linewidth}
\textbf{Prompt 11 (MI: 3.199, Acc: 0.220):}
\fbox{ \parbox{\textwidth}{ \fontsize{\figurefont}{\figurefont}\selectfont
Fill in the blank with the missing word or phrase.\\\hlx{Marissa loved \_\_\_\_\_ pokemon go game. It is the biggest thing right now. She had done so much more walking since she started playing it. She walked all day and evening sometimes. She walked almost 10 miles in two days.} \\Answer:\hly{ }
}}\end{minipage} \parbox{\textwidth}{\textbf{Collapsing token sets:} None, all tokens are \newline considered\newline}
\begin{minipage}{.95\linewidth}
\textbf{Prompt 12 (MI: 2.013, Acc: 0.214):}
\fbox{ \parbox{\textwidth}{ \fontsize{\figurefont}{\figurefont}\selectfont
\hlx{Marissa loved \_\_\_\_\_ pokemon go game. It is the biggest thing right now. She had done so much more walking since she started playing it. She walked all day and evening sometimes. She walked almost 10 miles in two days.} \\Fill in the blank with the missing word or phrase to complete the sentence.\\What is the missing word? The missing word is "\hly{ }
}}\end{minipage} \parbox{\textwidth}{\textbf{Collapsing token sets:} None, all tokens are \newline considered\newline}
\begin{minipage}{.95\linewidth}
\textbf{Prompt 13 (MI: 3.116, Acc: 0.182):}
\fbox{ \parbox{\textwidth}{ \fontsize{\figurefont}{\figurefont}\selectfont
Read the following sentences, and try to guess which word goes in the blank.\\\hlx{Marissa loved \_\_\_\_\_ pokemon go game. It is the biggest thing right now. She had done so much more walking since she started playing it. She walked all day and evening sometimes. She walked almost 10 miles in two days.} \\Answer:\hly{ }
}}\end{minipage} \parbox{\textwidth}{\textbf{Collapsing token sets:} None, all tokens are \newline considered\newline}
\begin{minipage}{.95\linewidth}
\textbf{Prompt 14 (MI: 1.843, Acc: 0.158):}
\fbox{ \parbox{\textwidth}{ \fontsize{\figurefont}{\figurefont}\selectfont
\hlx{Marissa loved \_\_\_\_\_ pokemon go game. It is the biggest thing right now. She had done so much more walking since she started playing it. She walked all day and evening sometimes. She walked almost 10 miles in two days.} \\\\The missing word in the story should be: "\hly{ }
}}\end{minipage} \parbox{\textwidth}{\textbf{Collapsing token sets:} None, all tokens are \newline considered\newline}
\begin{minipage}{.95\linewidth}
\textbf{Prompt 15 (MI: 2.681, Acc: 0.140):}
\fbox{ \parbox{\textwidth}{ \fontsize{\figurefont}{\figurefont}\selectfont
P1: I'm going to tell you a story, but leave a word out. Once I'm done telling the story, pick the word that best fits in the blank. \\\hlx{Marissa loved \_\_\_\_\_ pokemon go game. It is the biggest thing right now. She had done so much more walking since she started playing it. She walked all day and evening sometimes. She walked almost 10 miles in two days.} \\P2: The word which fits best is "\hly{ }
}}\end{minipage} \parbox{\textwidth}{\textbf{Collapsing token sets:} None, all tokens are \newline considered\newline}
\begin{minipage}{.95\linewidth}
\textbf{Prompt 16 (MI: 2.150, Acc: 0.120):}
\fbox{ \parbox{\textwidth}{ \fontsize{\figurefont}{\figurefont}\selectfont
\hlx{Marissa loved \_\_\_\_\_ pokemon go game. It is the biggest thing right now. She had done so much more walking since she started playing it. She walked all day and evening sometimes. She walked almost 10 miles in two days.} \\Which word should we put in the blank to complete the story? Let's use the word "\hly{ }
}}\end{minipage} \parbox{\textwidth}{\textbf{Collapsing token sets:} None, all tokens are \newline considered\newline}
\begin{minipage}{.95\linewidth}
\textbf{Prompt 17 (MI: 2.634, Acc: 0.088):}
\fbox{ \parbox{\textwidth}{ \fontsize{\figurefont}{\figurefont}\selectfont
\hlx{Marissa loved \_\_\_\_\_ pokemon go game. It is the biggest thing right now. She had done so much more walking since she started playing it. She walked all day and evening sometimes. She walked almost 10 miles in two days.} \\Which word fills in the blank best?\\The word that fills in the blank best is "\hly{ }
}}\end{minipage} \parbox{\textwidth}{\textbf{Collapsing token sets:} None, all tokens are \newline considered\newline}
\begin{minipage}{.95\linewidth}
\textbf{Prompt 18 (MI: 2.637, Acc: 0.086):}
\fbox{ \parbox{\textwidth}{ \fontsize{\figurefont}{\figurefont}\selectfont
P1: What word do you think fits best in the following story? \\\hlx{Marissa loved \_\_\_\_\_ pokemon go game. It is the biggest thing right now. She had done so much more walking since she started playing it. She walked all day and evening sometimes. She walked almost 10 miles in two days.} \\P2: The word which fits best is "\hly{ }
}}\end{minipage} \parbox{\textwidth}{\textbf{Collapsing token sets:} None, all tokens are \newline considered\newline}
\begin{minipage}{.95\linewidth}
\textbf{Prompt 19 (MI: 3.648, Acc: 0.050):}
\fbox{ \parbox{\textwidth}{ \fontsize{\figurefont}{\figurefont}\selectfont
It was a cold night. The wind was \_\_\_\_\_ around the courtyard as I stepped out of the car and into the darkness.\\Word: whistling\\\\\hlx{Marissa loved \_\_\_\_\_ pokemon go game. It is the biggest thing right now. She had done so much more walking since she started playing it. She walked all day and evening sometimes. She walked almost 10 miles in two days.} \\Put the best word in the blank to complete the story. \\Word:\hly{ }
}}\end{minipage} \parbox{\textwidth}{\textbf{Collapsing token sets:} None, all tokens are \newline considered\newline}
\begin{minipage}{.95\linewidth}
\textbf{Prompt 20 (MI: 1.891, Acc: 0.036):}
\fbox{ \parbox{\textwidth}{ \fontsize{\figurefont}{\figurefont}\selectfont
\hlx{Marissa loved \_\_\_\_\_ pokemon go game. It is the biggest thing right now. She had done so much more walking since she started playing it. She walked all day and evening sometimes. She walked almost 10 miles in two days.} \\Choose a word to replace the blank. \\Word: "\hly{ }
}}\end{minipage} \parbox{\textwidth}{\textbf{Collapsing token sets:} None, all tokens are \newline considered\newline}
\newpage
\noindent
\subsection{CoQA}
\begin{minipage}{.95\linewidth}
 \textbf{Prompt 1 (MI: 0.600, Acc: 0.590):}
\fbox{ \parbox{\textwidth}{ \fontsize{\figurefont}{\figurefont}\selectfont
Instructions: For each question below, choose the answer from the answer bank corresponding to the question that best answers the question.\\\\Question 1 Answer Bank: ladybug, bunny, goldfish, leopard, caterpillarQuestion: What animal would be most dangerous for a human to encounter in the wild?\\\\Answer: leopard\\\\Question 2 Answer Bank: \hlx{wrong, pleasure, encouragement, depression, relief}\\\\Question: \hlx{If you're still in love and end up stopping being married to your partner, what emotion are you likely to experience?}\\\\Answer:\hly{ }
}}\end{minipage} \parbox{\textwidth}{\textbf{Collapsing token sets:} \{'A': ['wrong'], \newline 'B': ['pleasure'], 'C': ['encouragement'], \newline 'D': ['depression'], 'E': ['relief']\}\newline}
\begin{minipage}{.95\linewidth}
\textbf{Prompt 2 (MI: 0.233, Acc: 0.546):}
\fbox{ \parbox{\textwidth}{ \fontsize{\figurefont}{\figurefont}\selectfont
Common Sense Quiz Answer Key\\\\Question 1: Where would people not typically go for fun?\\A: theme park\\B: movie theatre\\C: carnival\\D: waste management facility\\E: beach\\Correct Answer: D\\\\Question 2: \hlx{If you're still in love and end up stopping being married to your partner, what emotion are you likely to experience?}\\\hlx{A: wrong\\B: pleasure\\C: encouragement\\D: depression\\E: relief}\\Correct Answer:\hly{ }
}}\end{minipage} \parbox{\textwidth}{\textbf{Collapsing token sets:} ['A', 'B', 'C', 'D', 'E']\newline}
\begin{minipage}{.95\linewidth}
\textbf{Prompt 3 (MI: 0.474, Acc: 0.470):}
\fbox{ \parbox{\textwidth}{ \fontsize{\figurefont}{\figurefont}\selectfont
Given the following questions and choices, pick the choice that corresponds best to the question.\\\\"I'm crossing the river, my feet are wet but my body is dry, where am I?", "bridge, waterfall, valley, pebble, mountain", -> "valley"\\"In what Spanish speaking North American country can you get a great cup of coffee?", "mexico, mildred's coffee shop, diner, kitchen, canteen", -> "mexico"\\"\hlx{If you're still in love and end up stopping being married to your partner, what emotion are you likely to experience?}", "\hlx{wrong, pleasure, encouragement, depression, relief}" -> "\hly{ }
}}\end{minipage} \parbox{\textwidth}{\textbf{Collapsing token sets:} \{'A': ['wrong'], \newline 'B': ['pleasure'], 'C': ['encouragement'], \newline 'D': ['depression'], 'E': ['relief']\}\newline}
\newline
\newline
\newline
\newline
\newline
\newline
\newline
\begin{minipage}{.95\linewidth}
\textbf{Prompt 4 (MI: 0.083, Acc: 0.466):}
\fbox{ \parbox{\textwidth}{ \fontsize{\figurefont}{\figurefont}\selectfont
What would you use to put out a fire?\\A: gasoline\\B: poison\\C: laundry detergent\\D: water\\E: pencil\\Answer: D. water\\\\\hlx{If you're still in love and end up stopping being married to your partner, what emotion are you likely to experience?}\\\hlx{A: wrong\\B: pleasure\\C: encouragement\\D: depression\\E: relief}\\Answer:\hly{ }
}}\end{minipage} \parbox{\textwidth}{\textbf{Collapsing token sets:} ['A', 'B', 'C', 'D', 'E']\newline}
\begin{minipage}{.95\linewidth}
\textbf{Prompt 5 (MI: 0.504, Acc: 0.462):}
\fbox{ \parbox{\textwidth}{ \fontsize{\figurefont}{\figurefont}\selectfont
\# multiple choice quiz questions and answers\\\\qa = ['q': 'What is France?', 'choices': ['state', 'city', 'country', 'continent', 'mountain range'], 'answer': 'country', ], '[q': '\hlx{If you're still in love and end up stopping being married to your partner, what emotion are you likely to experience?}', 'choices': [\hlx{wrong, pleasure, encouragement, depression, relief}], 'answer': '\hly{ }
}}\end{minipage} \parbox{\textwidth}{\textbf{Collapsing token sets:} \{'A': ['wrong'], \newline 'B': ['pleasure'], 'C': ['encouragement'], \newline 'D': ['depression'], 'E': ['relief']\}\newline}
\begin{minipage}{.95\linewidth}
\textbf{Prompt 6 (MI: 0.431, Acc: 0.448):}
\fbox{ \parbox{\textwidth}{ \fontsize{\figurefont}{\figurefont}\selectfont
Given the following questions and choices, pick the choice that corresponds best to the question.\\\\"I'm crossing the river, my feet are wet but my body is dry, where am I?", "bridge, waterfall, valley, pebble, mountain", -> "valley"\\"\hlx{If you're still in love and end up stopping being married to your partner, what emotion are you likely to experience?}", "\hlx{wrong, pleasure, encouragement, depression, relief}" -> "\hly{ }
}}\end{minipage} \parbox{\textwidth}{\textbf{Collapsing token sets:} \{'A': ['wrong'], \newline 'B': ['pleasure'], 'C': ['encouragement'], \newline 'D': ['depression'], 'E': ['relief']\}\newline}
\newpage
\noindent
\begin{minipage}{.95\linewidth}
\textbf{Prompt 7 (MI: 0.417, Acc: 0.428):}
\fbox{ \parbox{\textwidth}{ \fontsize{\figurefont}{\figurefont}\selectfont
Choose the best single answer to the question, and explain your answer.\\\\Question: I'm crossing the river, my feet are wet but my body is dry, where am I?\\Choices: bridge, waterfall, valley, pebble, mountain\\Answer: "valley" is the best answer. While "bridge" also seems to make sense at first, your feet would not be wet if you crossed over a river on a bridge. Meanwhile, if you crossed the river at a valley, the river would be shallow, only getting your feet wet.\\\\Question: In what Spanish speaking North American country can you get a great cup of coffee?\\Choices: mildred's coffee shop, mexico, diner, kitchen, canteen\\Answer: "mexico" is the best answer. It's true that you can get a cup of coffee in a coffee shop or a diner, but the question specifically asks for a Spanish speaking North American country. Mexico is the only country listed, so that must be the correct answer.\\\\Question: \hlx{If you're still in love and end up stopping being married to your partner, what emotion are you likely to experience?}\\Choices: \hlx{wrong, pleasure, encouragement, depression, relief}\\Answer: "\hly{ }
}}\end{minipage} \parbox{\textwidth}{\textbf{Collapsing token sets:} \{'A': ['wrong'], \newline 'B': ['pleasure'], 'C': ['encouragement'], \newline 'D': ['depression'], 'E': ['relief']\}\newline}
\begin{minipage}{.95\linewidth}
\textbf{Prompt 8 (MI: 0.364, Acc: 0.408):}
\fbox{ \parbox{\textwidth}{ \fontsize{\figurefont}{\figurefont}\selectfont
Q: What might a vegan eat for breakfast?\\\\Choices: oats, bacon, sausage, omelet, ham\\\\A: oats\\\\Q: \hlx{If you're still in love and end up stopping being married to your partner, what emotion are you likely to experience?}\\\\Choices: \hlx{wrong, pleasure, encouragement, depression, relief}\\\\A:\hly{ }
}}\end{minipage} \parbox{\textwidth}{\textbf{Collapsing token sets:} \{'A': ['wrong'], \newline 'B': ['pleasure'], 'C': ['encouragement'], \newline 'D': ['depression'], 'E': ['relief']\}\newline}
\begin{minipage}{.95\linewidth}
\textbf{Prompt 9 (MI: 0.410, Acc: 0.408):}
\fbox{ \parbox{\textwidth}{ \fontsize{\figurefont}{\figurefont}\selectfont
What would you use to put out a fire?\\A: gasoline\\B: poison\\C: laundry detergent\\D: water\\E: pencil\\Answer: water\\\\\hlx{If you're still in love and end up stopping being married to your partner, what emotion are you likely to experience?}\\\hlx{A: wrong\\B: pleasure\\C: encouragement\\D: depression\\E: relief}\\Answer:\hly{ }
}}\end{minipage} \parbox{\textwidth}{\textbf{Collapsing token sets:} ['A', 'B', 'C', 'D', 'E']\newline}
\begin{minipage}{.95\linewidth}
\textbf{Prompt 10 (MI: 0.363, Acc: 0.396):}
\fbox{ \parbox{\textwidth}{ \fontsize{\figurefont}{\figurefont}\selectfont
Choose the best single answer to the question, and explain your answer.\\\\Question: I'm crossing the river, my feet are wet but my body is dry, where am I?\\Choices: bridge, waterfall, valley, pebble, mountain\\Answer: "valley" is the best answer. While "bridge" also seems to make sense at first, your feet would not be wet if you crossed over a river on a bridge. Meanwhile, if you crossed the river at a valley, the river would be shallow, only getting your feet wet.\\\\Question: \hlx{If you're still in love and end up stopping being married to your partner, what emotion are you likely to experience?}\\Choices: \hlx{wrong, pleasure, encouragement, depression, relief}\\Answer: "\hly{ }
}}\end{minipage} \parbox{\textwidth}{\textbf{Collapsing token sets:} \{'A': ['wrong'], \newline 'B': ['pleasure'], 'C': ['encouragement'], \newline 'D': ['depression'], 'E': ['relief']\}\newline}
\begin{minipage}{.95\linewidth}
\textbf{Prompt 11 (MI: 0.059, Acc: 0.380):}
\fbox{ \parbox{\textwidth}{ \fontsize{\figurefont}{\figurefont}\selectfont
Common Sense Quiz Answer Key\\\\Question 1: \hlx{If you're still in love and end up stopping being married to your partner, what emotion are you likely to experience?}\\\\\hlx{A: wrong\\B: pleasure\\C: encouragement\\D: depression\\E: relief}\\\\Correct Answer:\hly{ }
}}\end{minipage} \parbox{\textwidth}{\textbf{Collapsing token sets:} ['A', 'B', 'C', 'D', 'E']\newline}
\begin{minipage}{.95\linewidth}
\textbf{Prompt 12 (MI: 0.233, Acc: 0.360):}
\fbox{ \parbox{\textwidth}{ \fontsize{\figurefont}{\figurefont}\selectfont
Given the question, order the options from best answer to the question to worst answer to the question.\\\\Question: I'm crossing the river, my feet are wet but my body is dry, where am I?\\Choices: bridge, waterfall, valley, pebble, mountain\\Answers (in order of best to worst): valley, bridge, waterfall, mountain, pebble\\\\Question: \hlx{If you're still in love and end up stopping being married to your partner, what emotion are you likely to experience?}\\Choices: \hlx{wrong, pleasure, encouragement, depression, relief}\\Answers (in order of best to worst):\hly{ }
}}\end{minipage} \parbox{\textwidth}{\textbf{Collapsing token sets:} \{'A': ['wrong'], \newline 'B': ['pleasure'], 'C': ['encouragement'], \newline 'D': ['depression'], 'E': ['relief']\}\newline}
\begin{minipage}{.95\linewidth}
\textbf{Prompt 13 (MI: 0.255, Acc: 0.360):}
\fbox{ \parbox{\textwidth}{ \fontsize{\figurefont}{\figurefont}\selectfont
Given the question, order the options from best answer to the question to worst answer to the question.\\\\Question: I'm crossing the river, my feet are wet but my body is dry, where am I?\\Choices: bridge, waterfall, valley, pebble, mountain\\Answers (in order of best to worst): valley, bridge, waterfall, mountain, pebble\\\\Question: In what Spanish speaking North American country can you get a great cup of coffee?\\Choices: mildred's coffee shop, mexico, diner, kitchen, canteen\\Answers (in order of best to worst): mexico, mildred's coffee shop, diner, kitchen, canteen\\\\Question: \hlx{If you're still in love and end up stopping being married to your partner, what emotion are you likely to experience?}\\Choices: \hlx{wrong, pleasure, encouragement, depression, relief}\\Answers (in order of best to worst):\hly{ }
}}\end{minipage} \parbox{\textwidth}{\textbf{Collapsing token sets:} \{'A': ['wrong'], \newline 'B': ['pleasure'], 'C': ['encouragement'], \newline 'D': ['depression'], 'E': ['relief']\}\newline}
\begin{minipage}{.95\linewidth}
\textbf{Prompt 14 (MI: 0.222, Acc: 0.354):}
\fbox{ \parbox{\textwidth}{ \fontsize{\figurefont}{\figurefont}\selectfont
Q: \hlx{If you're still in love and end up stopping being married to your partner, what emotion are you likely to experience?}\\\\Choices: \hlx{wrong, pleasure, encouragement, depression, relief}\\\\A:\hly{ }
}}\end{minipage} \parbox{\textwidth}{\textbf{Collapsing token sets:} \{'A': ['wrong'], \newline 'B': ['pleasure'], 'C': ['encouragement'], \newline 'D': ['depression'], 'E': ['relief']\}\newline}
\begin{minipage}{.95\linewidth}
\textbf{Prompt 15 (MI: 0.246, Acc: 0.342):}
\fbox{ \parbox{\textwidth}{ \fontsize{\figurefont}{\figurefont}\selectfont
Teacher: I'm going to ask you a common sense question.\\\\Student: Alright.\\\\Teacher: \hlx{If you're still in love and end up stopping being married to your partner, what emotion are you likely to experience?}\\\\Student: What are the possible answers?\\\\Teacher: The answer is either "wrong," "pleasure," "encouragement," "depression," or "relief."\\\\Student: I know the right answer - it's "\hly{ }
}}\end{minipage} \parbox{\textwidth}{\textbf{Collapsing token sets:} \{'A': ['wrong'], \newline 'B': ['pleasure'], 'C': ['encouragement'], \newline 'D': ['depression'], 'E': ['relief']\}\newline}
\begin{minipage}{.95\linewidth}
\textbf{Prompt 16 (MI: 0.376, Acc: 0.336):}
\fbox{ \parbox{\textwidth}{ \fontsize{\figurefont}{\figurefont}\selectfont
questions,choices,answers\\"What is France?","[state,city,country,continent,mountain range]",country\\"\hlx{If you're still in love and end up stopping being married to your partner, what emotion are you likely to experience?}","[wrong,pleasure,encouragement,depression,relief]",\hly{ }
}}\end{minipage} \parbox{\textwidth}{\textbf{Collapsing token sets:} \{'A': ['wrong'], \newline 'B': ['pleasure'], 'C': ['encouragement'], \newline 'D': ['depression'], 'E': ['relief']\}\newline}
\begin{minipage}{.95\linewidth}
\textbf{Prompt 17 (MI: 0.265, Acc: 0.276):}
\fbox{ \parbox{\textwidth}{ \fontsize{\figurefont}{\figurefont}\selectfont
Me: I watched the most recent episode of the "Is It Really Common Sense" game show yesterday night.\\Friend: Oh, how was it?\\Me: It was good. I remember one of the questions.\\Friend: What was the question?\\Me: \hlx{If you're still in love and end up stopping being married to your partner, what emotion are you likely to experience?}\\Friend: What were the options?\\Me: wrong, pleasure, encouragement, depression, or relief\\Friend: Did the contestant get the answer right?\\Me: Yep!\\Friend: Which of the options was correct?\\Me: The correct answer was\hly{ }
}}\end{minipage} \parbox{\textwidth}{\textbf{Collapsing token sets:} \{'A': ['wrong'], \newline 'B': ['pleasure'], 'C': ['encouragement'], \newline 'D': ['depression'], 'E': ['relief']\}\newline}
\begin{minipage}{.95\linewidth}
\textbf{Prompt 18 (MI: 0.197, Acc: 0.248):}
\fbox{ \parbox{\textwidth}{ \fontsize{\figurefont}{\figurefont}\selectfont
Given the question, order the options from best answer to the question to worst answer to the question.\\\\Question: \hlx{If you're still in love and end up stopping being married to your partner, what emotion are you likely to experience?}\\Choices: \hlx{wrong, pleasure, encouragement, depression, relief}\\Answers (in order of best to worst):\hly{ }
}}\end{minipage} \parbox{\textwidth}{\textbf{Collapsing token sets:} \{'A': ['wrong'], \newline 'B': ['pleasure'], 'C': ['encouragement'], \newline 'D': ['depression'], 'E': ['relief']\}\newline}
\begin{minipage}{.95\linewidth}
\textbf{Prompt 19 (MI: 0.013, Acc: 0.234):}
\fbox{ \parbox{\textwidth}{ \fontsize{\figurefont}{\figurefont}\selectfont
\hlx{If you're still in love and end up stopping being married to your partner, what emotion are you likely to experience?}\\\\\hlx{A: wrong\\B: pleasure\\C: encouragement\\D: depression\\E: relief}\\\\Answer:\hly{ }
}}\end{minipage} \parbox{\textwidth}{\textbf{Collapsing token sets:} ['A', 'B', 'C', 'D', 'E']\newline}
\newpage
\noindent
\begin{minipage}{.95\linewidth}
\textbf{Prompt 20 (MI: 0.241, Acc: 0.228):}
\fbox{ \parbox{\textwidth}{ \fontsize{\figurefont}{\figurefont}\selectfont
Teacher: I'm going to ask you a common sense question.\\\\Student: Alright.\\\\Teacher: What would you not expect to read about in a book on the founding of the United States?\\\\Student: What are the possible answers?\\\\Teacher: The answer is either "george washington," "declaration of independence," "boston tea party," "star spangled banner," or "vampire assassins."\\\\Student: I know the right answer - it's "vampire assassins."\\\\Teacher: That's right! Here's another common sense question for you. \hlx{If you're still in love and end up stopping being married to your partner, what emotion are you likely to experience?}\\\\Student: What are the possible answers?\\\\Teacher: The answer is either "wrong," "pleasure," "encouragement," "depression," or "relief."\\\\Student: I know the right answer - it's "\hly{ }
}}\end{minipage} \parbox{\textwidth}{\textbf{Collapsing token sets:} \{'A': ['wrong'], \newline 'B': ['pleasure'], 'C': ['encouragement'], \newline 'D': ['depression'], 'E': ['relief']\}\newline}
\\ \\\subsection{IMDB}
\begin{minipage}{.95\linewidth}
 \textbf{Prompt 1 (MI: 0.175, Acc: 0.944):}
\fbox{ \parbox{\textwidth}{ \fontsize{\figurefont}{\figurefont}\selectfont
P1: How was the movie? \\P2: \hlx{John Cassavetes is on the run from the law. He is at the bottom of the heap. He sees Negro Sidney Poitier as his equal and they quickly become friends, forming a sort of alliance against a bully of a foreman played by Jack Warden.\\\\As someone who has worked in a warehouse myself when I was younger, I can tell you that the warehouse fights, complete with tumbling packing cases and flailing grappling hooks are as realistic as it gets. I've been in fights like these myself, although no one got killed.\\\\The introduction of Sidney Poitier's widow is a variation on Shakespeare's Shylock "Do I not bleed?" This is an anti racist film, which, at the time, was much needed.\\\\All the three principle characters - Warden, Cassavetes and Poitier - are superb, with Warden the most outstanding of the three.} \\P1: Would you say your review of the movie is negative or positive? \\P2: I would say my review review of the movie is\hly{ }
}}\end{minipage} \parbox{\textwidth}{\textbf{Collapsing token sets:} \{'positive': ['positive'], \newline 'negative': ['negative']\}\newline}
\newline
\newline
\newline
\newline
\newline
\newline
\newline
\newline
\newline
\begin{minipage}{.95\linewidth}
\textbf{Prompt 2 (MI: 0.306, Acc: 0.920):}
\fbox{ \parbox{\textwidth}{ \fontsize{\figurefont}{\figurefont}\selectfont
P1: Could you give me a review of the movie you just saw? \\P2: Sure, \hlx{John Cassavetes is on the run from the law. He is at the bottom of the heap. He sees Negro Sidney Poitier as his equal and they quickly become friends, forming a sort of alliance against a bully of a foreman played by Jack Warden.\\\\As someone who has worked in a warehouse myself when I was younger, I can tell you that the warehouse fights, complete with tumbling packing cases and flailing grappling hooks are as realistic as it gets. I've been in fights like these myself, although no one got killed.\\\\The introduction of Sidney Poitier's widow is a variation on Shakespeare's Shylock "Do I not bleed?" This is an anti racist film, which, at the time, was much needed.\\\\All the three principle characters - Warden, Cassavetes and Poitier - are superb, with Warden the most outstanding of the three.} \\P1: So, overall, would you give it a positive or negative review? \\P2: I would give it a\hly{ }
}}\end{minipage} \parbox{\textwidth}{\textbf{Collapsing token sets:} \{'positive': ['positive'], \newline 'negative': ['negative']\}\newline}
\begin{minipage}{.95\linewidth}
\textbf{Prompt 3 (MI: 0.154, Acc: 0.904):}
\fbox{ \parbox{\textwidth}{ \fontsize{\figurefont}{\figurefont}\selectfont
Considering this movie review, determine its sentiment.\\\\Review:  \hlx{John Cassavetes is on the run from the law. He is at the bottom of the heap. He sees Negro Sidney Poitier as his equal and they quickly become friends, forming a sort of alliance against a bully of a foreman played by Jack Warden.\\\\As someone who has worked in a warehouse myself when I was younger, I can tell you that the warehouse fights, complete with tumbling packing cases and flailing grappling hooks are as realistic as it gets. I've been in fights like these myself, although no one got killed.\\\\The introduction of Sidney Poitier's widow is a variation on Shakespeare's Shylock "Do I not bleed?" This is an anti racist film, which, at the time, was much needed.\\\\All the three principle characters - Warden, Cassavetes and Poitier - are superb, with Warden the most outstanding of the three.}\\\\In general, was the sentiment positive or negative The sentiment was\hly{ }
}}\end{minipage} \parbox{\textwidth}{\textbf{Collapsing token sets:} \{'positive': ['positive'], \newline 'negative': ['negative']\}\newline}
\newpage
\noindent
\begin{minipage}{.95\linewidth}
\textbf{Prompt 4 (MI: 0.260, Acc: 0.898):}
\fbox{ \parbox{\textwidth}{ \fontsize{\figurefont}{\figurefont}\selectfont
P1: How was the movie? \\P2: \hlx{John Cassavetes is on the run from the law. He is at the bottom of the heap. He sees Negro Sidney Poitier as his equal and they quickly become friends, forming a sort of alliance against a bully of a foreman played by Jack Warden.\\\\As someone who has worked in a warehouse myself when I was younger, I can tell you that the warehouse fights, complete with tumbling packing cases and flailing grappling hooks are as realistic as it gets. I've been in fights like these myself, although no one got killed.\\\\The introduction of Sidney Poitier's widow is a variation on Shakespeare's Shylock "Do I not bleed?" This is an anti racist film, which, at the time, was much needed.\\\\All the three principle characters - Warden, Cassavetes and Poitier - are superb, with Warden the most outstanding of the three.} \\P1: Would you say your review of the movie is positive or negative? \\P2: I would say my review of the movie is\hly{ }
}}\end{minipage} \parbox{\textwidth}{\textbf{Collapsing token sets:} \{'positive': ['positive'], \newline 'negative': ['negative']\}\newline}
\begin{minipage}{.95\linewidth}
\textbf{Prompt 5 (MI: 0.237, Acc: 0.888):}
\fbox{ \parbox{\textwidth}{ \fontsize{\figurefont}{\figurefont}\selectfont
After reading the following review, classify it as negative or positive. \\\\Review: \hlx{John Cassavetes is on the run from the law. He is at the bottom of the heap. He sees Negro Sidney Poitier as his equal and they quickly become friends, forming a sort of alliance against a bully of a foreman played by Jack Warden.\\\\As someone who has worked in a warehouse myself when I was younger, I can tell you that the warehouse fights, complete with tumbling packing cases and flailing grappling hooks are as realistic as it gets. I've been in fights like these myself, although no one got killed.\\\\The introduction of Sidney Poitier's widow is a variation on Shakespeare's Shylock "Do I not bleed?" This is an anti racist film, which, at the time, was much needed.\\\\All the three principle characters - Warden, Cassavetes and Poitier - are superb, with Warden the most outstanding of the three.} \\\\Classification:\hly{ }
}}\end{minipage} \parbox{\textwidth}{\textbf{Collapsing token sets:} \{'positive': ['positive'], \newline 'negative': ['negative']\}\newline}
\begin{minipage}{.95\linewidth}
\textbf{Prompt 6 (MI: 0.151, Acc: 0.886):}
\fbox{ \parbox{\textwidth}{ \fontsize{\figurefont}{\figurefont}\selectfont
Read the following movie review to determine the review's sentiment.\\\\\hlx{John Cassavetes is on the run from the law. He is at the bottom of the heap. He sees Negro Sidney Poitier as his equal and they quickly become friends, forming a sort of alliance against a bully of a foreman played by Jack Warden.\\\\As someone who has worked in a warehouse myself when I was younger, I can tell you that the warehouse fights, complete with tumbling packing cases and flailing grappling hooks are as realistic as it gets. I've been in fights like these myself, although no one got killed.\\\\The introduction of Sidney Poitier's widow is a variation on Shakespeare's Shylock "Do I not bleed?" This is an anti racist film, which, at the time, was much needed.\\\\All the three principle characters - Warden, Cassavetes and Poitier - are superb, with Warden the most outstanding of the three.}\\\\In general, was the sentiment positive or negative? The sentiment was\hly{ }
}}\end{minipage} \parbox{\textwidth}{\textbf{Collapsing token sets:} \{'positive': ['positive'], \newline 'negative': ['negative']\}\newline}
\begin{minipage}{.95\linewidth}
\textbf{Prompt 7 (MI: 0.086, Acc: 0.886):}
\fbox{ \parbox{\textwidth}{ \fontsize{\figurefont}{\figurefont}\selectfont
Considering this movie review, determine its sentiment.\\\\Review:\\"""\\\hlx{John Cassavetes is on the run from the law. He is at the bottom of the heap. He sees Negro Sidney Poitier as his equal and they quickly become friends, forming a sort of alliance against a bully of a foreman played by Jack Warden.\\\\As someone who has worked in a warehouse myself when I was younger, I can tell you that the warehouse fights, complete with tumbling packing cases and flailing grappling hooks are as realistic as it gets. I've been in fights like these myself, although no one got killed.\\\\The introduction of Sidney Poitier's widow is a variation on Shakespeare's Shylock "Do I not bleed?" This is an anti racist film, which, at the time, was much needed.\\\\All the three principle characters - Warden, Cassavetes and Poitier - are superb, with Warden the most outstanding of the three.}\\"""\\In general, what was the sentiment of the review? The sentiment was\hly{ }
}}\end{minipage} \parbox{\textwidth}{\textbf{Collapsing token sets:} \{'positive': ['positive'], \newline 'negative': ['negative']\}\newline}
\begin{minipage}{.95\linewidth}
\textbf{Prompt 8 (MI: 0.274, Acc: 0.858):}
\fbox{ \parbox{\textwidth}{ \fontsize{\figurefont}{\figurefont}\selectfont
Yesterday I went to see a movie. \hlx{John Cassavetes is on the run from the law. He is at the bottom of the heap. He sees Negro Sidney Poitier as his equal and they quickly become friends, forming a sort of alliance against a bully of a foreman played by Jack Warden.\\\\As someone who has worked in a warehouse myself when I was younger, I can tell you that the warehouse fights, complete with tumbling packing cases and flailing grappling hooks are as realistic as it gets. I've been in fights like these myself, although no one got killed.\\\\The introduction of Sidney Poitier's widow is a variation on Shakespeare's Shylock "Do I not bleed?" This is an anti racist film, which, at the time, was much needed.\\\\All the three principle characters - Warden, Cassavetes and Poitier - are superb, with Warden the most outstanding of the three.} Between positive and negative, I would say the movie was\hly{ }
}}\end{minipage} \parbox{\textwidth}{\textbf{Collapsing token sets:} \{'positive': ['positive'], \newline 'negative': ['negative']\}\newline}
\begin{minipage}{.95\linewidth}
\textbf{Prompt 9 (MI: 0.026, Acc: 0.852):}
\fbox{ \parbox{\textwidth}{ \fontsize{\figurefont}{\figurefont}\selectfont
Q: Is the sentiment of the following movie review negative or positive?\\"""\\\hlx{John Cassavetes is on the run from the law. He is at the bottom of the heap. He sees Negro Sidney Poitier as his equal and they quickly become friends, forming a sort of alliance against a bully of a foreman played by Jack Warden.\\\\As someone who has worked in a warehouse myself when I was younger, I can tell you that the warehouse fights, complete with tumbling packing cases and flailing grappling hooks are as realistic as it gets. I've been in fights like these myself, although no one got killed.\\\\The introduction of Sidney Poitier's widow is a variation on Shakespeare's Shylock "Do I not bleed?" This is an anti racist film, which, at the time, was much needed.\\\\All the three principle characters - Warden, Cassavetes and Poitier - are superb, with Warden the most outstanding of the three.}\\"""\\A: The sentiment of the movie review was\hly{ }
}}\end{minipage} \parbox{\textwidth}{\textbf{Collapsing token sets:} \{'positive': ['positive'], \newline 'negative': ['negative']\}\newline}
\newpage
\noindent
\begin{minipage}{.95\linewidth}
\textbf{Prompt 10 (MI: 0.119, Acc: 0.842):}
\fbox{ \parbox{\textwidth}{ \fontsize{\figurefont}{\figurefont}\selectfont
Read the following movie review to determine the review's sentiment.\\\\\hlx{John Cassavetes is on the run from the law. He is at the bottom of the heap. He sees Negro Sidney Poitier as his equal and they quickly become friends, forming a sort of alliance against a bully of a foreman played by Jack Warden.\\\\As someone who has worked in a warehouse myself when I was younger, I can tell you that the warehouse fights, complete with tumbling packing cases and flailing grappling hooks are as realistic as it gets. I've been in fights like these myself, although no one got killed.\\\\The introduction of Sidney Poitier's widow is a variation on Shakespeare's Shylock "Do I not bleed?" This is an anti racist film, which, at the time, was much needed.\\\\All the three principle characters - Warden, Cassavetes and Poitier - are superb, with Warden the most outstanding of the three.}\\\\In general, was the sentiment negative or positive? The sentiment was\hly{ }
}}\end{minipage} \parbox{\textwidth}{\textbf{Collapsing token sets:} \{'positive': ['positive'], \newline 'negative': ['negative']\}\newline}
\begin{minipage}{.95\linewidth}
\textbf{Prompt 11 (MI: 0.162, Acc: 0.824):}
\fbox{ \parbox{\textwidth}{ \fontsize{\figurefont}{\figurefont}\selectfont
Q: Is the sentiment of the following movie review positive or negative?\\\hlx{John Cassavetes is on the run from the law. He is at the bottom of the heap. He sees Negro Sidney Poitier as his equal and they quickly become friends, forming a sort of alliance against a bully of a foreman played by Jack Warden.\\\\As someone who has worked in a warehouse myself when I was younger, I can tell you that the warehouse fights, complete with tumbling packing cases and flailing grappling hooks are as realistic as it gets. I've been in fights like these myself, although no one got killed.\\\\The introduction of Sidney Poitier's widow is a variation on Shakespeare's Shylock "Do I not bleed?" This is an anti racist film, which, at the time, was much needed.\\\\All the three principle characters - Warden, Cassavetes and Poitier - are superb, with Warden the most outstanding of the three.} \\A (positive or negative):\hly{ }
}}\end{minipage} \parbox{\textwidth}{\textbf{Collapsing token sets:} \{'positive': ['positive'], \newline 'negative': ['negative']\}\newline}
\newpage
\noindent
\begin{minipage}{.95\linewidth}
\textbf{Prompt 12 (MI: 0.101, Acc: 0.822):}
\fbox{ \parbox{\textwidth}{ \fontsize{\figurefont}{\figurefont}\selectfont
Q: Is the sentiment of the following movie review negative or positive?\\\hlx{John Cassavetes is on the run from the law. He is at the bottom of the heap. He sees Negro Sidney Poitier as his equal and they quickly become friends, forming a sort of alliance against a bully of a foreman played by Jack Warden.\\\\As someone who has worked in a warehouse myself when I was younger, I can tell you that the warehouse fights, complete with tumbling packing cases and flailing grappling hooks are as realistic as it gets. I've been in fights like these myself, although no one got killed.\\\\The introduction of Sidney Poitier's widow is a variation on Shakespeare's Shylock "Do I not bleed?" This is an anti racist film, which, at the time, was much needed.\\\\All the three principle characters - Warden, Cassavetes and Poitier - are superb, with Warden the most outstanding of the three.} \\A (negative or positive):\hly{ }
}}\end{minipage} \parbox{\textwidth}{\textbf{Collapsing token sets:} \{'positive': ['positive'], \newline 'negative': ['negative']\}\newline}
\begin{minipage}{.95\linewidth}
\textbf{Prompt 13 (MI: 0.084, Acc: 0.810):}
\fbox{ \parbox{\textwidth}{ \fontsize{\figurefont}{\figurefont}\selectfont
Considering this movie review, determine its sentiment.\\\\Review:\\"""\\\hlx{John Cassavetes is on the run from the law. He is at the bottom of the heap. He sees Negro Sidney Poitier as his equal and they quickly become friends, forming a sort of alliance against a bully of a foreman played by Jack Warden.\\\\As someone who has worked in a warehouse myself when I was younger, I can tell you that the warehouse fights, complete with tumbling packing cases and flailing grappling hooks are as realistic as it gets. I've been in fights like these myself, although no one got killed.\\\\The introduction of Sidney Poitier's widow is a variation on Shakespeare's Shylock "Do I not bleed?" This is an anti racist film, which, at the time, was much needed.\\\\All the three principle characters - Warden, Cassavetes and Poitier - are superb, with Warden the most outstanding of the three.}\\"""\\In general, was the sentiment positive or negative? The sentiment was\hly{ }
}}\end{minipage} \parbox{\textwidth}{\textbf{Collapsing token sets:} \{'positive': ['positive'], \newline 'negative': ['negative']\}\newline}
\newpage
\noindent
\begin{minipage}{.95\linewidth}
\textbf{Prompt 14 (MI: 0.201, Acc: 0.798):}
\fbox{ \parbox{\textwidth}{ \fontsize{\figurefont}{\figurefont}\selectfont
P1: Could you give me a review of the movie you just saw? \\P2: Sure, \hlx{John Cassavetes is on the run from the law. He is at the bottom of the heap. He sees Negro Sidney Poitier as his equal and they quickly become friends, forming a sort of alliance against a bully of a foreman played by Jack Warden.\\\\As someone who has worked in a warehouse myself when I was younger, I can tell you that the warehouse fights, complete with tumbling packing cases and flailing grappling hooks are as realistic as it gets. I've been in fights like these myself, although no one got killed.\\\\The introduction of Sidney Poitier's widow is a variation on Shakespeare's Shylock "Do I not bleed?" This is an anti racist film, which, at the time, was much needed.\\\\All the three principle characters - Warden, Cassavetes and Poitier - are superb, with Warden the most outstanding of the three.} \\P1: So overall was the sentiment of the movie negative or positive? \\P2: I would give it a\hly{ }
}}\end{minipage} \parbox{\textwidth}{\textbf{Collapsing token sets:} \{'positive': ['positive'], \newline 'negative': ['negative']\}\newline}
\begin{minipage}{.95\linewidth}
\textbf{Prompt 15 (MI: 0.234, Acc: 0.786):}
\fbox{ \parbox{\textwidth}{ \fontsize{\figurefont}{\figurefont}\selectfont
After reading the following review, classify it as positive or negative. \\\\Review: \hlx{John Cassavetes is on the run from the law. He is at the bottom of the heap. He sees Negro Sidney Poitier as his equal and they quickly become friends, forming a sort of alliance against a bully of a foreman played by Jack Warden.\\\\As someone who has worked in a warehouse myself when I was younger, I can tell you that the warehouse fights, complete with tumbling packing cases and flailing grappling hooks are as realistic as it gets. I've been in fights like these myself, although no one got killed.\\\\The introduction of Sidney Poitier's widow is a variation on Shakespeare's Shylock "Do I not bleed?" This is an anti racist film, which, at the time, was much needed.\\\\All the three principle characters - Warden, Cassavetes and Poitier - are superb, with Warden the most outstanding of the three.} \\\\Classification:\hly{ }
}}\end{minipage} \parbox{\textwidth}{\textbf{Collapsing token sets:} \{'positive': ['positive'], \newline 'negative': ['negative']\}\newline}
\newpage
\noindent
\begin{minipage}{.95\linewidth}
\textbf{Prompt 16 (MI: 0.042, Acc: 0.628):}
\fbox{ \parbox{\textwidth}{ \fontsize{\figurefont}{\figurefont}\selectfont
Q: Is the sentiment of the following movie review positive or negative?\\"""\\\hlx{John Cassavetes is on the run from the law. He is at the bottom of the heap. He sees Negro Sidney Poitier as his equal and they quickly become friends, forming a sort of alliance against a bully of a foreman played by Jack Warden.\\\\As someone who has worked in a warehouse myself when I was younger, I can tell you that the warehouse fights, complete with tumbling packing cases and flailing grappling hooks are as realistic as it gets. I've been in fights like these myself, although no one got killed.\\\\The introduction of Sidney Poitier's widow is a variation on Shakespeare's Shylock "Do I not bleed?" This is an anti racist film, which, at the time, was much needed.\\\\All the three principle characters - Warden, Cassavetes and Poitier - are superb, with Warden the most outstanding of the three.}\\"""\\A: The sentiment of the movie review was\hly{ }
}}\end{minipage} \parbox{\textwidth}{\textbf{Collapsing token sets:} \{'positive': ['positive'], \newline 'negative': ['negative']\}\newline}
\begin{minipage}{.95\linewidth}
\textbf{Prompt 17 (MI: 0.021, Acc: 0.486):}
\fbox{ \parbox{\textwidth}{ \fontsize{\figurefont}{\figurefont}\selectfont
\hlx{John Cassavetes is on the run from the law. He is at the bottom of the heap. He sees Negro Sidney Poitier as his equal and they quickly become friends, forming a sort of alliance against a bully of a foreman played by Jack Warden.\\\\As someone who has worked in a warehouse myself when I was younger, I can tell you that the warehouse fights, complete with tumbling packing cases and flailing grappling hooks are as realistic as it gets. I've been in fights like these myself, although no one got killed.\\\\The introduction of Sidney Poitier's widow is a variation on Shakespeare's Shylock "Do I not bleed?" This is an anti racist film, which, at the time, was much needed.\\\\All the three principle characters - Warden, Cassavetes and Poitier - are superb, with Warden the most outstanding of the three.}\\\\Was the previous review negative or positive? The previous review was\hly{ }
}}\end{minipage} \parbox{\textwidth}{\textbf{Collapsing token sets:} \{'positive': ['positive'], \newline 'negative': ['negative']\}\newline}
\newpage
\noindent
\begin{minipage}{.95\linewidth}
\textbf{Prompt 18 (MI: 0.016, Acc: 0.484):}
\fbox{ \parbox{\textwidth}{ \fontsize{\figurefont}{\figurefont}\selectfont
\hlx{John Cassavetes is on the run from the law. He is at the bottom of the heap. He sees Negro Sidney Poitier as his equal and they quickly become friends, forming a sort of alliance against a bully of a foreman played by Jack Warden.\\\\As someone who has worked in a warehouse myself when I was younger, I can tell you that the warehouse fights, complete with tumbling packing cases and flailing grappling hooks are as realistic as it gets. I've been in fights like these myself, although no one got killed.\\\\The introduction of Sidney Poitier's widow is a variation on Shakespeare's Shylock "Do I not bleed?" This is an anti racist film, which, at the time, was much needed.\\\\All the three principle characters - Warden, Cassavetes and Poitier - are superb, with Warden the most outstanding of the three.}\\\\Was the previous review positive or negative? The previous review was\hly{ }
}}\end{minipage} \parbox{\textwidth}{\textbf{Collapsing token sets:} \{'positive': ['positive'], \newline 'negative': ['negative']\}\newline}
\begin{minipage}{.95\linewidth}
\textbf{Prompt 19 (MI: 0.019, Acc: 0.462):}
\fbox{ \parbox{\textwidth}{ \fontsize{\figurefont}{\figurefont}\selectfont
\hlx{John Cassavetes is on the run from the law. He is at the bottom of the heap. He sees Negro Sidney Poitier as his equal and they quickly become friends, forming a sort of alliance against a bully of a foreman played by Jack Warden.\\\\As someone who has worked in a warehouse myself when I was younger, I can tell you that the warehouse fights, complete with tumbling packing cases and flailing grappling hooks are as realistic as it gets. I've been in fights like these myself, although no one got killed.\\\\The introduction of Sidney Poitier's widow is a variation on Shakespeare's Shylock "Do I not bleed?" This is an anti racist film, which, at the time, was much needed.\\\\All the three principle characters - Warden, Cassavetes and Poitier - are superb, with Warden the most outstanding of the three.}\\\\Was the sentiment of previous review positive or negative? The previous review was\hly{ }
}}\end{minipage} \parbox{\textwidth}{\textbf{Collapsing token sets:} \{'positive': ['positive'], \newline 'negative': ['negative']\}\newline}
\begin{minipage}{.95\linewidth}
\textbf{Prompt 20 (MI: 0.017, Acc: 0.450):}
\fbox{ \parbox{\textwidth}{ \fontsize{\figurefont}{\figurefont}\selectfont
\hlx{John Cassavetes is on the run from the law. He is at the bottom of the heap. He sees Negro Sidney Poitier as his equal and they quickly become friends, forming a sort of alliance against a bully of a foreman played by Jack Warden.\\\\As someone who has worked in a warehouse myself when I was younger, I can tell you that the warehouse fights, complete with tumbling packing cases and flailing grappling hooks are as realistic as it gets. I've been in fights like these myself, although no one got killed.\\\\The introduction of Sidney Poitier's widow is a variation on Shakespeare's Shylock "Do I not bleed?" This is an anti racist film, which, at the time, was much needed.\\\\All the three principle characters - Warden, Cassavetes and Poitier - are superb, with Warden the most outstanding of the three.}\\\\Was the sentiment of previous review negative or positive? The previous review was\hly{ }
}}\end{minipage} \parbox{\textwidth}{\textbf{Collapsing token sets:} \{'positive': ['positive'], \newline 'negative': ['negative']\}\newline}
\\ \\\subsection{BoolQ}
\begin{minipage}{.95\linewidth}
 \textbf{Prompt 1 (MI: 0.077, Acc: 0.778):}
\fbox{ \parbox{\textwidth}{ \fontsize{\figurefont}{\figurefont}\selectfont
Given the passage and question, please answer the question with yes or no.\\\\'''Turn on red -- In Canada, left turn on red light from a one-way road into a one-way road is permitted except in some areas of Quebec, New Brunswick, and Prince Edward Island. Left turn on red light from a two-way road into a one-way road is permitted in British Columbia but only if the driver turns onto the closest lane and yields to pedestrians and cross traffic.''', '''Can you turn left on red in canada?''' -> '''Yes'''\\\\'''\hlx{Pyruvic acid -- Pyruvic acid (CHCOCOOH) is the simplest of the alpha-keto acids, with a carboxylic acid and a ketone functional group. Pyruvate (/paruvet/), the conjugate base, CHCOCOO, is a key intermediate in several metabolic pathways.}''', '''\hlx{Is pyruvic acid and pyruvate the same thing?}''' -> '''\hly{ }
}}\end{minipage} \parbox{\textwidth}{\textbf{Collapsing token sets:} \{'True': ['yes'], \newline 'False': ['no']\}\newline}
\begin{minipage}{.95\linewidth}
\textbf{Prompt 2 (MI: 0.090, Acc: 0.752):}
\fbox{ \parbox{\textwidth}{ \fontsize{\figurefont}{\figurefont}\selectfont
Passage: "Turn on red -- In Canada, left turn on red light from a one-way road into a one-way road is permitted except in some areas of Quebec, New Brunswick, and Prince Edward Island. Left turn on red light from a two-way road into a one-way road is permitted in British Columbia but only if the driver turns onto the closest lane and yields to pedestrians and cross traffic."\\Question: "Can you turn left on red in canada?"\\Answer: "Yes"\\\\Passage: "\hlx{Pyruvic acid -- Pyruvic acid (CHCOCOOH) is the simplest of the alpha-keto acids, with a carboxylic acid and a ketone functional group. Pyruvate (/paruvet/), the conjugate base, CHCOCOO, is a key intermediate in several metabolic pathways.}"\\Question: "\hlx{Is pyruvic acid and pyruvate the same thing?}"\\Answer: "\hly{ }
}}\end{minipage} \parbox{\textwidth}{\textbf{Collapsing token sets:} \{'True': ['yes'], \newline 'False': ['no']\}\newline}
\newline
\newline
\newline
\newline
\newline
\newline
\begin{minipage}{.95\linewidth}
\textbf{Prompt 3 (MI: 0.055, Acc: 0.750):}
\fbox{ \parbox{\textwidth}{ \fontsize{\figurefont}{\figurefont}\selectfont
Given the passage and question, please answer the question with yes or no.\\\\'''Turn on red -- In Canada, left turn on red light from a one-way road into a one-way road is permitted except in some areas of Quebec, New Brunswick, and Prince Edward Island. Left turn on red light from a two-way road into a one-way road is permitted in British Columbia but only if the driver turns onto the closest lane and yields to pedestrians and cross traffic.''', '''Can you turn left on red in canada?''' -> '''Yes'''\\\\'''Lord Voldemort -- Lord Voldemort ( known as Tom Marvolo Riddle) is a fictional character and the main antagonist in J.K. Rowling's series of Harry Potter novels. Voldemort first appeared in Harry Potter and the Philosopher's Stone, which was released in 1997. Voldemort appears either in person or in flashbacks in each book and its film adaptation in the series, except the third, Harry Potter and the Prisoner of Azkaban, where he is only mentioned.''', '''Are tom riddle and lord voldemort the same person?''' -> '''Yes'''\\\\'''Clerks -- Clerks is a 1994 American independent black-and-white comedy film written, directed and co-produced by Kevin Smith. Starring Brian O'Halloran as Dante Hicks and Jeff Anderson as Randal Graves, it presents a day in the lives of two store clerks and their acquaintances.''', '''Is the movie clerks in colors?''' -> '''No'''\\\\'''\hlx{Pyruvic acid -- Pyruvic acid (CHCOCOOH) is the simplest of the alpha-keto acids, with a carboxylic acid and a ketone functional group. Pyruvate (/paruvet/), the conjugate base, CHCOCOO, is a key intermediate in several metabolic pathways.}''', '''\hlx{Is pyruvic acid and pyruvate the same thing?}''' -> '''\hly{ }
}}\end{minipage} \parbox{\textwidth}{\textbf{Collapsing token sets:} \{'True': ['yes'], \newline 'False': ['no']\}\newline}
\begin{minipage}{.95\linewidth}
\textbf{Prompt 4 (MI: 0.076, Acc: 0.740):}
\fbox{ \parbox{\textwidth}{ \fontsize{\figurefont}{\figurefont}\selectfont
Passage: "Turn on red -- In Canada, left turn on red light from a one-way road into a one-way road is permitted except in some areas of Quebec, New Brunswick, and Prince Edward Island. Left turn on red light from a two-way road into a one-way road is permitted in British Columbia but only if the driver turns onto the closest lane and yields to pedestrians and cross traffic."\\Question: "Can you turn left on red in canada?"\\Answer: "Yes"\\\\Passage: "Lord Voldemort -- Lord Voldemort ( known as Tom Marvolo Riddle) is a fictional character and the main antagonist in J.K. Rowling's series of Harry Potter novels. Voldemort first appeared in Harry Potter and the Philosopher's Stone, which was released in 1997. Voldemort appears either in person or in flashbacks in each book and its film adaptation in the series, except the third, Harry Potter and the Prisoner of Azkaban, where he is only mentioned."\\Question: "Are tom riddle and lord voldemort the same person?"\\Answer: "Yes"\\\\Passage: "Clerks -- Clerks is a 1994 American independent black-and-white comedy film written, directed and co-produced by Kevin Smith. Starring Brian O'Halloran as Dante Hicks and Jeff Anderson as Randal Graves, it presents a day in the lives of two store clerks and their acquaintances."\\Question: "Is the movie clerks in colors?"\\Answer: "No"\\\\Passage: "\hlx{Pyruvic acid -- Pyruvic acid (CHCOCOOH) is the simplest of the alpha-keto acids, with a carboxylic acid and a ketone functional group. Pyruvate (/paruvet/), the conjugate base, CHCOCOO, is a key intermediate in several metabolic pathways.}"\\Question: "\hlx{Is pyruvic acid and pyruvate the same thing?}"\\Answer: "\hly{ }
}}\end{minipage} \parbox{\textwidth}{\textbf{Collapsing token sets:} \{'True': ['yes'], \newline 'False': ['no']\}\newline}
\begin{minipage}{.95\linewidth}
\textbf{Prompt 5 (MI: 0.037, Acc: 0.740):}
\fbox{ \parbox{\textwidth}{ \fontsize{\figurefont}{\figurefont}\selectfont
Given the passage and question, please answer the question with yes or no.\\\\'''Turn on red -- In Canada, left turn on red light from a one-way road into a one-way road is permitted except in some areas of Quebec, New Brunswick, and Prince Edward Island. Left turn on red light from a two-way road into a one-way road is permitted in British Columbia but only if the driver turns onto the closest lane and yields to pedestrians and cross traffic.''', '''Can you turn left on red in canada?''' -> '''Yes'''\\\\'''Lord Voldemort -- Lord Voldemort ( known as Tom Marvolo Riddle) is a fictional character and the main antagonist in J.K. Rowling's series of Harry Potter novels. Voldemort first appeared in Harry Potter and the Philosopher's Stone, which was released in 1997. Voldemort appears either in person or in flashbacks in each book and its film adaptation in the series, except the third, Harry Potter and the Prisoner of Azkaban, where he is only mentioned.''', '''Are tom riddle and lord voldemort the same person?''' -> '''Yes'''\\\\'''\hlx{Pyruvic acid -- Pyruvic acid (CHCOCOOH) is the simplest of the alpha-keto acids, with a carboxylic acid and a ketone functional group. Pyruvate (/paruvet/), the conjugate base, CHCOCOO, is a key intermediate in several metabolic pathways.}''', '''\hlx{Is pyruvic acid and pyruvate the same thing?}''' -> '''\hly{ }
}}\end{minipage} \parbox{\textwidth}{\textbf{Collapsing token sets:} \{'True': ['yes'], \newline 'False': ['no']\}\newline}
\begin{minipage}{.95\linewidth}
\textbf{Prompt 6 (MI: 0.068, Acc: 0.702):}
\fbox{ \parbox{\textwidth}{ \fontsize{\figurefont}{\figurefont}\selectfont
"\hlx{Pyruvic acid -- Pyruvic acid (CHCOCOOH) is the simplest of the alpha-keto acids, with a carboxylic acid and a ketone functional group. Pyruvate (/paruvet/), the conjugate base, CHCOCOO, is a key intermediate in several metabolic pathways.}"\\\\For the question: "\hlx{Is pyruvic acid and pyruvate the same thing?}"\\I would answer: "\hly{ }
}}\end{minipage} \parbox{\textwidth}{\textbf{Collapsing token sets:} \{'True': ['yes'], \newline 'False': ['no']\}\newline}
\begin{minipage}{.95\linewidth}
\textbf{Prompt 7 (MI: 0.039, Acc: 0.698):}
\fbox{ \parbox{\textwidth}{ \fontsize{\figurefont}{\figurefont}\selectfont
"\hlx{Pyruvic acid -- Pyruvic acid (CHCOCOOH) is the simplest of the alpha-keto acids, with a carboxylic acid and a ketone functional group. Pyruvate (/paruvet/), the conjugate base, CHCOCOO, is a key intermediate in several metabolic pathways.}"\\\\When picking between yes or no For the question: "\hlx{Is pyruvic acid and pyruvate the same thing?}"\\I would answer: "\hly{ }
}}\end{minipage} \parbox{\textwidth}{\textbf{Collapsing token sets:} \{'True': ['yes'], \newline 'False': ['no']\}\newline}
\begin{minipage}{.95\linewidth}
\textbf{Prompt 8 (MI: 0.034, Acc: 0.698):}
\fbox{ \parbox{\textwidth}{ \fontsize{\figurefont}{\figurefont}\selectfont
ANSWER KEY\\\\Please read the following passage with the following question in mind: "\hlx{Is pyruvic acid and pyruvate the same thing?}"\\\\\hlx{Pyruvic acid -- Pyruvic acid (CHCOCOOH) is the simplest of the alpha-keto acids, with a carboxylic acid and a ketone functional group. Pyruvate (/paruvet/), the conjugate base, CHCOCOO, is a key intermediate in several metabolic pathways.}\\\\\hlx{Is pyruvic acid and pyruvate the same thing?}\\Answer key: "\hly{ }
}}\end{minipage} \parbox{\textwidth}{\textbf{Collapsing token sets:} \{'True': ['yes'], \newline 'False': ['no']\}\newline}
\begin{minipage}{.95\linewidth}
\textbf{Prompt 9 (MI: 0.055, Acc: 0.688):}
\fbox{ \parbox{\textwidth}{ \fontsize{\figurefont}{\figurefont}\selectfont
Passage: "Turn on red -- In Canada, left turn on red light from a one-way road into a one-way road is permitted except in some areas of Quebec, New Brunswick, and Prince Edward Island. Left turn on red light from a two-way road into a one-way road is permitted in British Columbia but only if the driver turns onto the closest lane and yields to pedestrians and cross traffic."\\Question: "Can you turn left on red in canada?"\\Answer: "Yes"\\\\Passage: "Lord Voldemort -- Lord Voldemort ( known as Tom Marvolo Riddle) is a fictional character and the main antagonist in J.K. Rowling's series of Harry Potter novels. Voldemort first appeared in Harry Potter and the Philosopher's Stone, which was released in 1997. Voldemort appears either in person or in flashbacks in each book and its film adaptation in the series, except the third, Harry Potter and the Prisoner of Azkaban, where he is only mentioned."\\Question: "Are tom riddle and lord voldemort the same person?"\\Answer: "Yes"\\\\Passage: "\hlx{Pyruvic acid -- Pyruvic acid (CHCOCOOH) is the simplest of the alpha-keto acids, with a carboxylic acid and a ketone functional group. Pyruvate (/paruvet/), the conjugate base, CHCOCOO, is a key intermediate in several metabolic pathways.}"\\Question: "\hlx{Is pyruvic acid and pyruvate the same thing?}"\\Answer: "\hly{ }
}}\end{minipage} \parbox{\textwidth}{\textbf{Collapsing token sets:} \{'True': ['yes'], \newline 'False': ['no']\}\newline}
\begin{minipage}{.95\linewidth}
\textbf{Prompt 10 (MI: 0.052, Acc: 0.682):}
\fbox{ \parbox{\textwidth}{ \fontsize{\figurefont}{\figurefont}\selectfont
"\hlx{Pyruvic acid -- Pyruvic acid (CHCOCOOH) is the simplest of the alpha-keto acids, with a carboxylic acid and a ketone functional group. Pyruvate (/paruvet/), the conjugate base, CHCOCOO, is a key intermediate in several metabolic pathways.}"\\\\For the question: "\hlx{Is pyruvic acid and pyruvate the same thing?}"\\My answer would be: "\hly{ }
}}\end{minipage} \parbox{\textwidth}{\textbf{Collapsing token sets:} \{'True': ['yes'], \newline 'False': ['no']\}\newline}
\begin{minipage}{.95\linewidth}
\textbf{Prompt 11 (MI: 0.026, Acc: 0.682):}
\fbox{ \parbox{\textwidth}{ \fontsize{\figurefont}{\figurefont}\selectfont
Given the passage and question, please answer the question with yes or no.\\\\'''\hlx{Pyruvic acid -- Pyruvic acid (CHCOCOOH) is the simplest of the alpha-keto acids, with a carboxylic acid and a ketone functional group. Pyruvate (/paruvet/), the conjugate base, CHCOCOO, is a key intermediate in several metabolic pathways.}''', '''\hlx{Is pyruvic acid and pyruvate the same thing?}''' -> '''\hly{ }
}}\end{minipage} \parbox{\textwidth}{\textbf{Collapsing token sets:} \{'True': ['yes'], \newline 'False': ['no']\}\newline}
\begin{minipage}{.95\linewidth}
\textbf{Prompt 12 (MI: 0.016, Acc: 0.680):}
\fbox{ \parbox{\textwidth}{ \fontsize{\figurefont}{\figurefont}\selectfont
"\hlx{Pyruvic acid -- Pyruvic acid (CHCOCOOH) is the simplest of the alpha-keto acids, with a carboxylic acid and a ketone functional group. Pyruvate (/paruvet/), the conjugate base, CHCOCOO, is a key intermediate in several metabolic pathways.}"\\\\When picking between "true" or "false", For the question: "\hlx{Is pyruvic acid and pyruvate the same thing?}"\\My answer would be: "\hly{ }
}}\end{minipage} \parbox{\textwidth}{\textbf{Collapsing token sets:} \{'True': ['true'], \newline 'False': ['false']\}\newline}
\begin{minipage}{.95\linewidth}
\textbf{Prompt 13 (MI: 0.074, Acc: 0.674):}
\fbox{ \parbox{\textwidth}{ \fontsize{\figurefont}{\figurefont}\selectfont
Please read the following passage with the following question in mind: "\hlx{Is pyruvic acid and pyruvate the same thing?}"\\\\\hlx{Pyruvic acid -- Pyruvic acid (CHCOCOOH) is the simplest of the alpha-keto acids, with a carboxylic acid and a ketone functional group. Pyruvate (/paruvet/), the conjugate base, CHCOCOO, is a key intermediate in several metabolic pathways.}\\\\\hlx{Is pyruvic acid and pyruvate the same thing?}\\Answer: "\hly{ }
}}\end{minipage} \parbox{\textwidth}{\textbf{Collapsing token sets:} \{'True': ['yes'], \newline 'False': ['no']\}\newline}
\begin{minipage}{.95\linewidth}
\textbf{Prompt 14 (MI: 0.050, Acc: 0.668):}
\fbox{ \parbox{\textwidth}{ \fontsize{\figurefont}{\figurefont}\selectfont
Read the following passage: "\hlx{Pyruvic acid -- Pyruvic acid (CHCOCOOH) is the simplest of the alpha-keto acids, with a carboxylic acid and a ketone functional group. Pyruvate (/paruvet/), the conjugate base, CHCOCOO, is a key intermediate in several metabolic pathways.}"\\\\Given this question: "\hlx{Is pyruvic acid and pyruvate the same thing?}"\\I would answer: "\hly{ }
}}\end{minipage} \parbox{\textwidth}{\textbf{Collapsing token sets:} \{'True': ['yes'], \newline 'False': ['no']\}\newline}
\begin{minipage}{.95\linewidth}
\textbf{Prompt 15 (MI: 0.058, Acc: 0.646):}
\fbox{ \parbox{\textwidth}{ \fontsize{\figurefont}{\figurefont}\selectfont
Read the following passage: "\hlx{Pyruvic acid -- Pyruvic acid (CHCOCOOH) is the simplest of the alpha-keto acids, with a carboxylic acid and a ketone functional group. Pyruvate (/paruvet/), the conjugate base, CHCOCOO, is a key intermediate in several metabolic pathways.}"\\\\Given this question: "\hlx{Is pyruvic acid and pyruvate the same thing?}"\\I would respond: "\hly{ }
}}\end{minipage} \parbox{\textwidth}{\textbf{Collapsing token sets:} \{'True': ['yes'], \newline 'False': ['no']\}\newline}
\begin{minipage}{.95\linewidth}
\textbf{Prompt 16 (MI: 0.027, Acc: 0.634):}
\fbox{ \parbox{\textwidth}{ \fontsize{\figurefont}{\figurefont}\selectfont
Based on the passage: "\hlx{Pyruvic acid -- Pyruvic acid (CHCOCOOH) is the simplest of the alpha-keto acids, with a carboxylic acid and a ketone functional group. Pyruvate (/paruvet/), the conjugate base, CHCOCOO, is a key intermediate in several metabolic pathways.}"\\\\And answering the question: "\hlx{Is pyruvic acid and pyruvate the same thing?}"\\By choosing yes or no\\My answer would be: "\hly{ }
}}\end{minipage} \parbox{\textwidth}{\textbf{Collapsing token sets:} \{'True': ['yes'], \newline 'False': ['no']\}\newline}
\newpage
\noindent
\begin{minipage}{.95\linewidth}
\textbf{Prompt 17 (MI: 0.013, Acc: 0.522):}
\fbox{ \parbox{\textwidth}{ \fontsize{\figurefont}{\figurefont}\selectfont
Read the following passage: "\hlx{Pyruvic acid -- Pyruvic acid (CHCOCOOH) is the simplest of the alpha-keto acids, with a carboxylic acid and a ketone functional group. Pyruvate (/paruvet/), the conjugate base, CHCOCOO, is a key intermediate in several metabolic pathways.}\\\\Given this question: "\hlx{Is pyruvic acid and pyruvate the same thing?}"\\\\If asked to choose "true" or "false", My answer would be: "\hly{ }
}}\end{minipage} \parbox{\textwidth}{\textbf{Collapsing token sets:} \{'True': ['true'], \newline 'False': ['false']\}\newline}
\begin{minipage}{.95\linewidth}
\textbf{Prompt 18 (MI: 0.020, Acc: 0.518):}
\fbox{ \parbox{\textwidth}{ \fontsize{\figurefont}{\figurefont}\selectfont
Read the following passage: "\hlx{Pyruvic acid -- Pyruvic acid (CHCOCOOH) is the simplest of the alpha-keto acids, with a carboxylic acid and a ketone functional group. Pyruvate (/paruvet/), the conjugate base, CHCOCOO, is a key intermediate in several metabolic pathways.}"\\\\Given this question: "\hlx{Is pyruvic acid and pyruvate the same thing?}"\\If asked to choose yes or no, My answer would be: "\hly{ }
}}\end{minipage} \parbox{\textwidth}{\textbf{Collapsing token sets:} \{'True': ['yes'], \newline 'False': ['no']\}\newline}
\begin{minipage}{.95\linewidth}
\textbf{Prompt 19 (MI: 0.013, Acc: 0.452):}
\fbox{ \parbox{\textwidth}{ \fontsize{\figurefont}{\figurefont}\selectfont
Read the following passage: "\hlx{Pyruvic acid -- Pyruvic acid (CHCOCOOH) is the simplest of the alpha-keto acids, with a carboxylic acid and a ketone functional group. Pyruvate (/paruvet/), the conjugate base, CHCOCOO, is a key intermediate in several metabolic pathways.}"\\\\Given this question: "\hlx{Is pyruvic acid and pyruvate the same thing?}"\\If asked to choose "true" or "false", I would answer: "\hly{ }
}}\end{minipage} \parbox{\textwidth}{\textbf{Collapsing token sets:} \{'True': ['true'], \newline 'False': ['false']\}\newline}
\begin{minipage}{.95\linewidth}
\textbf{Prompt 20 (MI: 0.022, Acc: 0.438):}
\fbox{ \parbox{\textwidth}{ \fontsize{\figurefont}{\figurefont}\selectfont
Read the following passage: "\hlx{Pyruvic acid -- Pyruvic acid (CHCOCOOH) is the simplest of the alpha-keto acids, with a carboxylic acid and a ketone functional group. Pyruvate (/paruvet/), the conjugate base, CHCOCOO, is a key intermediate in several metabolic pathways.}"\\\\Given this question: "\hlx{Is pyruvic acid and pyruvate the same thing?}"\\If asked to choose yes or no, I would answer: "\hly{ }
}}\end{minipage} \parbox{\textwidth}{\textbf{Collapsing token sets:} \{'True': ['yes'], \newline 'False': ['no']\}\newline}
\\ \\\subsection{COPA}
\begin{minipage}{.95\linewidth}
 \textbf{Prompt 1 (MI: 0.044, Acc: 0.782):}
\fbox{ \parbox{\textwidth}{ \fontsize{\figurefont}{\figurefont}\selectfont
For the following premises, choose the alternative that is either a cause or result of the premise, and justify your answer.\\\\Premise: The man broke his toe. What was the CAUSE of this?\\Alternative 1: He got a hole in his sock. \\Alternative 2: He dropped a hammer on his foot.\\Answer: Alternative 2. Getting a hole in your sock would not break your toe, unless there is additional information. Dropping a hammer (which is a heavy object), on the other hand, would almost certaintly break your toe. Thus, the best answer is Alternative 2.\\\\Premise: I tipped the bottle. What happened as a RESULT?\\Alternative 1: The liquid in the bottle froze.\\Alternative 2: The liquid in the bottle poured out.\\Answer: Alternative 2. Tipping a bottle causes liquid to fall out, not to freeze. Freezing is caused by being placed in a cold place. Pouring out (Alternative 2) is correct because it makes the most sense.\\\\Premise: I knocked on my neighbor's door. What happened as a RESULT?\\Alternative 1: My neighbor invited me in.\\Alternative 2: My neighbor left his house.\\Answer: Alternative 1. When you knock on a neighbor's door, it is likely that if they are home they will answer and invite you in. It does not make much sense, however, that a neighbor would leave their house without explanation. Therefore, Alternative 1 is the best result of the premise.\\\\Premise: \hlx{My foot went numb.} What happened as a RESULT?\\Alternative 1: \hlx{I put my shoes on.}\\Alternative 2: \hlx{I shook my foot.}\\Answer: Alternative\hly{ }
}}\end{minipage} \parbox{\textwidth}{\textbf{Collapsing token sets:} \{'1': ['1'], '2': ['2']\}\newline}
\begin{minipage}{.95\linewidth}
\textbf{Prompt 2 (MI: 0.034, Acc: 0.762):}
\fbox{ \parbox{\textwidth}{ \fontsize{\figurefont}{\figurefont}\selectfont
The Choice Of Plausible Alternatives (COPA) evaluation provides researchers with a tool for assessing progress in open-domain commonsense causal reasoning. COPA consists of 1000 questions, split equally into development and test sets of 500 questions each. Each question is composed of a premise and two alternatives, where the task is to select the alternative that more plausibly has a causal relation with the premise. The correct alternative is randomized so that the expected performance of randomly guessing is 50\%.\\\\Examples\\\\Premise: The man broke his toe. What was the CAUSE of this?\\Alternative 1: He got a hole in his sock. \\Alternative 2: He dropped a hammer on his foot.\\Answer: Alternative 2\\\\Premise: I tipped the bottle. What happened as a RESULT?\\Alternative 1: The liquid in the bottle froze.\\Alternative 2: The liquid in the bottle poured out.\\Answer: Alternative 2\\\\Premise: I knocked on my neighbor's door. What happened as a RESULT?\\Alternative 1: My neighbor invited me in.\\Alternative 2: My neighbor left his house.\\Answer: Alternative 1\\\\Premise: \hlx{My foot went numb.} What happened as a RESULT?\\Alternative 1: \hlx{I put my shoes on.}\\Alternative 2: \hlx{I shook my foot.}\\Answer: Alternative\hly{ }
}}\end{minipage} \parbox{\textwidth}{\textbf{Collapsing token sets:} \{'1': ['1'], '2': ['2']\}\newline}
\begin{minipage}{.95\linewidth}
\textbf{Prompt 3 (MI: 0.003, Acc: 0.628):}
\fbox{ \parbox{\textwidth}{ \fontsize{\figurefont}{\figurefont}\selectfont
What is the effect of the following premise: "\hlx{My foot went numb.}"\\\\Choice 1. \hlx{I put my shoes on.}\\Choice 2. \hlx{I shook my foot.}\\Answer: Choice\hly{ }
}}\end{minipage} \parbox{\textwidth}{\textbf{Collapsing token sets:} \{'1': ['1'], '2': ['2']\}\newline}
\begin{minipage}{.95\linewidth}
\textbf{Prompt 4 (MI: 0.002, Acc: 0.612):}
\fbox{ \parbox{\textwidth}{ \fontsize{\figurefont}{\figurefont}\selectfont
Solve the following COPA task by choosing the sentence which makes the most sense after the premise.\\\\Premise: \hlx{My foot went numb.}\\Choice 1. \hlx{I put my shoes on.}\\Choice 2. \hlx{I shook my foot.}\\Answer: Choice\hly{ }
}}\end{minipage} \parbox{\textwidth}{\textbf{Collapsing token sets:} \{'1': ['1'], '2': ['2']\}\newline}
\begin{minipage}{.95\linewidth}
\textbf{Prompt 5 (MI: 0.003, Acc: 0.550):}
\fbox{ \parbox{\textwidth}{ \fontsize{\figurefont}{\figurefont}\selectfont
If asked to pick between choice 1 ("\hlx{I put my shoes on.}") or choice 2 ("\hlx{I shook my foot.}") to see what the effect of this premise ("\hlx{My foot went numb.}") was, I would say: "choice\hly{ }
}}\end{minipage} \parbox{\textwidth}{\textbf{Collapsing token sets:} \{'1': ['1'], '2': ['2']\}\newline}
\begin{minipage}{.95\linewidth}
\textbf{Prompt 6 (MI: 0.010, Acc: 0.540):}
\fbox{ \parbox{\textwidth}{ \fontsize{\figurefont}{\figurefont}\selectfont
Solve the following COPA tasks by choosing the sentence which makes the most sense after the premise.\\\\Premise: The man broke his toe.\\Choice 1. He got a hole in his sock.\\Choice 2. He dropped a hammer on his foot.\\Answer: Choice 2.\\\\Premise: \hlx{My foot went numb.}\\Choice 1. \hlx{I put my shoes on.}\\Choice 2. \hlx{I shook my foot.}\\Answer: Choice\hly{ }
}}\end{minipage} \parbox{\textwidth}{\textbf{Collapsing token sets:} \{'1': ['1'], '2': ['2']\}\newline}
\begin{minipage}{.95\linewidth}
\textbf{Prompt 7 (MI: 0.002, Acc: 0.532):}
\fbox{ \parbox{\textwidth}{ \fontsize{\figurefont}{\figurefont}\selectfont
What is the effect of the following premise: "\hlx{My foot went numb.}"\\\\If asked to choose between Choice 1: "\hlx{I put my shoes on.}" or Choice 2: "\hlx{I shook my foot.}"\\My answer would be: Choice\hly{ }
}}\end{minipage} \parbox{\textwidth}{\textbf{Collapsing token sets:} \{'1': ['1'], '2': ['2']\}\newline}
\begin{minipage}{.95\linewidth}
\textbf{Prompt 8 (MI: 0.006, Acc: 0.530):}
\fbox{ \parbox{\textwidth}{ \fontsize{\figurefont}{\figurefont}\selectfont
I will give you a premise and you will choose either sentence 1) or 2) which is the better plausible alternative.\\Premise: \hlx{My foot went numb.}\\1) \hlx{I put my shoes on.}\\2) \hlx{I shook my foot.}\\The most plausible alternative is: Sentence\hly{ }
}}\end{minipage} \parbox{\textwidth}{\textbf{Collapsing token sets:} \{'1': ['1'], '2': ['2']\}\newline}
\begin{minipage}{.95\linewidth}
\textbf{Prompt 9 (MI: 0.018, Acc: 0.524):}
\fbox{ \parbox{\textwidth}{ \fontsize{\figurefont}{\figurefont}\selectfont
Read the following premise and answer by choosing "effect1" or "effect2"\\Premise: "\hlx{My foot went numb.}"\\effect1: "\hlx{I put my shoes on.}"\\effect2: "\hlx{I shook my foot.}"\\Answer: "effect\hly{ }
}}\end{minipage} \parbox{\textwidth}{\textbf{Collapsing token sets:} \{'1': ['1'], '2': ['2']\}\newline}
\begin{minipage}{.95\linewidth}
\textbf{Prompt 10 (MI: 0.008, Acc: 0.520):}
\fbox{ \parbox{\textwidth}{ \fontsize{\figurefont}{\figurefont}\selectfont
Read the following premise and pick "effect2" or "effect1"\\Premise: "\hlx{My foot went numb.}"\\effect1: "\hlx{I put my shoes on.}"\\effect2: "\hlx{I shook my foot.}"\\Answer: "effect\hly{ }
}}\end{minipage} \parbox{\textwidth}{\textbf{Collapsing token sets:} \{'1': ['1'], '2': ['2']\}\newline}
\begin{minipage}{.95\linewidth}
\textbf{Prompt 11 (MI: 0.003, Acc: 0.516):}
\fbox{ \parbox{\textwidth}{ \fontsize{\figurefont}{\figurefont}\selectfont
Based on this premise: "\hlx{My foot went numb.}"\\\\If asked to choose between\\Choice 1: "\hlx{I put my shoes on.}"\\or\\Choice 2: "\hlx{I shook my foot.}"\\My answer would be: Choice\hly{ }
}}\end{minipage} \parbox{\textwidth}{\textbf{Collapsing token sets:} \{'1': ['1'], '2': ['2']\}\newline}
\begin{minipage}{.95\linewidth}
\textbf{Prompt 12 (MI: 0.008, Acc: 0.510):}
\fbox{ \parbox{\textwidth}{ \fontsize{\figurefont}{\figurefont}\selectfont
Which one of these stories makes the most sense?\\Story 1: \hlx{My foot went numb.} \hlx{I put my shoes on.}\\Story 2: \hlx{My foot went numb.} \hlx{I shook my foot.}\\Answer: Story\hly{ }
}}\end{minipage} \parbox{\textwidth}{\textbf{Collapsing token sets:} \{'1': ['1'], '2': ['2']\}\newline}
\begin{minipage}{.95\linewidth}
\textbf{Prompt 13 (MI: 0.003, Acc: 0.506):}
\fbox{ \parbox{\textwidth}{ \fontsize{\figurefont}{\figurefont}\selectfont
P1: Here's a premise: "The man broke his toe."\\Which sentence provides the better alternative?\\1. "He got a hole in his sock", or \\2. "He dropped a hammer on his foot." \\P2: The better alternative is sentence\\\\P1: Here's a premise: "\hlx{My foot went numb.}".Which sentence provides the better alternative? 1. "I put my shoes on", or 2. "\hlx{I shook my foot.}"P2: The better alternative is sentence\hly{ }
}}\end{minipage} \parbox{\textwidth}{\textbf{Collapsing token sets:} \{'1': ['1'], '2': ['2']\}\newline}
\begin{minipage}{.95\linewidth}
\textbf{Prompt 14 (MI: 0.003, Acc: 0.504):}
\fbox{ \parbox{\textwidth}{ \fontsize{\figurefont}{\figurefont}\selectfont
Based on this premise: "\hlx{My foot went numb.}"\\\\If asked to pick between\\Choice 1: "\hlx{I put my shoes on.}" or Choice 2: "\hlx{I shook my foot.}" to get the effect\\ of the predeciding sentence, I would say: "Choice\hly{ }
}}\end{minipage} \parbox{\textwidth}{\textbf{Collapsing token sets:} \{'1': ['1'], '2': ['2']\}\newline}
\begin{minipage}{.95\linewidth}
\textbf{Prompt 15 (MI: 0.036, Acc: 0.502):}
\fbox{ \parbox{\textwidth}{ \fontsize{\figurefont}{\figurefont}\selectfont
I am going to tell you two stories, one of them will make sense and the other will not.\\Story 1: \hlx{My foot went numb.} \hlx{I put my shoes on.}\\Story 2: \hlx{My foot went numb.} \hlx{I shook my foot.}\\The story that makes sense is Story\hly{ }
}}\end{minipage} \parbox{\textwidth}{\textbf{Collapsing token sets:} \{'1': ['1'], '2': ['2']\}\newline}
\begin{minipage}{.95\linewidth}
\textbf{Prompt 16 (MI: 0.009, Acc: 0.502):}
\fbox{ \parbox{\textwidth}{ \fontsize{\figurefont}{\figurefont}\selectfont
\hlx{My foot went numb.}\\Which of the following alternatives is most plausible for the previous sentence?\\Sentence 1) \hlx{I put my shoes on.}\\Sentence 2) \hlx{I shook my foot.}\\The most plausible alternative is sentence\hly{ }
}}\end{minipage} \parbox{\textwidth}{\textbf{Collapsing token sets:} \{'1': ['1'], '2': ['2']\}\newline}
\begin{minipage}{.95\linewidth}
\textbf{Prompt 17 (MI: 0.006, Acc: 0.500):}
\fbox{ \parbox{\textwidth}{ \fontsize{\figurefont}{\figurefont}\selectfont
I will give you a premise and you will choose either sentence 1) or 2) which is the better plausible alternative.\\\\Premise: The man broke his toe. \\1) He got a hole in his sock.\\2) He dropped a hammer on his foot. \\The most plausible alternative is: Sentence 2).\\\\I will give you a premise and you will choose either sentence 1) or 2) which is the better plausible alternative.\\Premise: \hlx{My foot went numb.}\\1) \hlx{I put my shoes on.}\\2) \hlx{I shook my foot.}\\The most plausible alternative is: Sentence\hly{ }
}}\end{minipage} \parbox{\textwidth}{\textbf{Collapsing token sets:} \{'1': ['1'], '2': ['2']\}\newline}
\begin{minipage}{.95\linewidth}
\textbf{Prompt 18 (MI: 0.003, Acc: 0.500):}
\fbox{ \parbox{\textwidth}{ \fontsize{\figurefont}{\figurefont}\selectfont
P1: Here's a premise: \hlx{My foot went numb.}.Which sentence provides the better alternative? 1. "I put my shoes on", or 2. "\hlx{I shook my foot.}"P2: The better alternative is sentence\hly{ }
}}\end{minipage} \parbox{\textwidth}{\textbf{Collapsing token sets:} \{'1': ['1'], '2': ['2']\}\newline}
\begin{minipage}{.95\linewidth}
\textbf{Prompt 19 (MI: 0.019, Acc: 0.500):}
\fbox{ \parbox{\textwidth}{ \fontsize{\figurefont}{\figurefont}\selectfont
"The man broke his toe."\\Which of the following alternatives is most plausible for the previous sentence?\\\\Sentence 1) He got a hole in his sock.\\Sentence 2) He dropped a hammer on his foot. \\The most plausible alternative is sentence 2).\\"\hlx{My foot went numb.}"\\Which of the following alternatives is most plausible for the previous sentence?\\Sentence 1) \hlx{I put my shoes on.}\\Sentence 2) \hlx{I shook my foot.}\\The most plausible alternative is sentence\hly{ }
}}\end{minipage} \parbox{\textwidth}{\textbf{Collapsing token sets:} \{'1': ['1'], '2': ['2']\}\newline}
\begin{minipage}{.95\linewidth}
\textbf{Prompt 20 (MI: 0.001, Acc: 0.496):}
\fbox{ \parbox{\textwidth}{ \fontsize{\figurefont}{\figurefont}\selectfont
I want to figure out which effect of this sentence is more probably: "\hlx{My foot went numb.}"\\Choice 1: "\hlx{I put my shoes on.}" or Choice 2: "\hlx{I shook my foot.}"\\I would say: "Choice\hly{ }
}}\end{minipage} \parbox{\textwidth}{\textbf{Collapsing token sets:} \{'1': ['1'], '2': ['2']\}\newline}
\\ \\\subsection{WiC}
\begin{minipage}{.95\linewidth}
 \textbf{Prompt 1 (MI: 0.036, Acc: 0.520):}
\fbox{ \parbox{\textwidth}{ \fontsize{\figurefont}{\figurefont}\selectfont
Classify whether the following two sentences' use of the word has the same meaning or not.\\\\Word: bright\\Usage 1: He is a bright child\\Usage 2: The sun is very bright today\\Meaning: different\\\\Word: \hlx{didacticism}\\Usage 1: \hlx{The didacticism of the 19th century gave birth to many great museums.}\\Usage 2: \hlx{The didacticism expected in books for the young.}\\Meaning:\hly{ }
}}\end{minipage} \parbox{\textwidth}{\textbf{Collapsing token sets:} \{'True': ['same'], \newline 'False': ['different']\}\newline}
\begin{minipage}{.95\linewidth}
\textbf{Prompt 2 (MI: 0.006, Acc: 0.512):}
\fbox{ \parbox{\textwidth}{ \fontsize{\figurefont}{\figurefont}\selectfont
"\hlx{The didacticism of the 19th century gave birth to many great museums.}"\\"\hlx{The didacticism expected in books for the young.}"\\\\True or false, the word \hlx{didacticism} has the same meaning.\\Answer:\hly{ }
}}\end{minipage} \parbox{\textwidth}{\textbf{Collapsing token sets:} \{'True': ['true'], \newline 'False': ['false']\}\newline}
\begin{minipage}{.95\linewidth}
\textbf{Prompt 3 (MI: 0.025, Acc: 0.506):}
\fbox{ \parbox{\textwidth}{ \fontsize{\figurefont}{\figurefont}\selectfont
Depending on its context, an ambiguous word can refer to multiple, potentially unrelated, meanings. Mainstream static word embeddings, such as Word2vec and GloVe, are unable to reflect this dynamic semantic nature. Contextualised word embeddings are an attempt at addressing this limitation by computing dynamic representations for words which can adapt based on context. A system's task on the WiC dataset is to identify the intended meaning of words. WiC is framed as a binary classification task. Each instance in WiC has a target word w, either a verb or a noun, for which two contexts are provided. Each of these contexts triggers a specific meaning of w. The task is to identify if the occurrences of w in the two contexts correspond to the same meaning or not. In fact, the dataset can also be viewed as an application of Word Sense Disambiguation in practise.\\WiC features multiple interesting characteristics:\\\\* It is suitable for evaluating a wide range of applications, including contextualized word and sense representation and Word Sense Disambiguation;\\* It is framed asa binary classification dataset, in which, unlike Stanford Contextual Word Similarity (SCWS), identical words are paired with each other (in different contexts); hence, a context-insensitive word embedding model would perform similarly to a random baseline;\\* It is constructed using high quality annotations curated by experts.\\\\Examples from the dataset:\\Context-1 // Context-2 // Target // Label\\There's a lot of trash on the bed of the river // I keep a glass of water on my bed when I sleep // bed // Different\\Air pollution // Open a window and let in some air // air // Same\\\hlx{The didacticism of the 19th century gave birth to many great museums.} // \hlx{The didacticism expected in books for the young.} // \hlx{didacticism} //\hly{ }
}}\end{minipage} \parbox{\textwidth}{\textbf{Collapsing token sets:} \{'True': ['same'], \newline 'False': ['different']\}\newline}
\begin{minipage}{.95\linewidth}
\textbf{Prompt 4 (MI: 0.007, Acc: 0.504):}
\fbox{ \parbox{\textwidth}{ \fontsize{\figurefont}{\figurefont}\selectfont
"\hlx{The didacticism of the 19th century gave birth to many great museums.}"\\"\hlx{The didacticism expected in books for the young.}"\\\\True or False, the word "\hlx{didacticism}" has the same meaning.\\Answer:\hly{ }
}}\end{minipage} \parbox{\textwidth}{\textbf{Collapsing token sets:} \{'True': ['true'], \newline 'False': ['false']\}\newline}
\begin{minipage}{.95\linewidth}
\textbf{Prompt 5 (MI: 0.006, Acc: 0.504):}
\fbox{ \parbox{\textwidth}{ \fontsize{\figurefont}{\figurefont}\selectfont
Q: What does 2 + 2 equal?\\A: 4\\\\Q: Does the word "\hlx{didacticism}" have the same meaning in the following sentences? "\hlx{The didacticism of the 19th century gave birth to many great museums.}"; "\hlx{The didacticism expected in books for the young.}"\\A:\hly{ }
}}\end{minipage} \parbox{\textwidth}{\textbf{Collapsing token sets:} \{'True': ['yes'], \newline 'False': ['no']\}\newline}
\begin{minipage}{.95\linewidth}
\textbf{Prompt 6 (MI: 0.007, Acc: 0.496):}
\fbox{ \parbox{\textwidth}{ \fontsize{\figurefont}{\figurefont}\selectfont
Q: What year did America first land on the moon?\\A: 1969\\\\Q: Does the word "\hlx{didacticism}" have the same meaning in the following sentences? "\hlx{The didacticism of the 19th century gave birth to many great museums.}"; "\hlx{The didacticism expected in books for the young.}"\\A:\hly{ }
}}\end{minipage} \parbox{\textwidth}{\textbf{Collapsing token sets:} \{'True': ['yes'], \newline 'False': ['no']\}\newline}
\begin{minipage}{.95\linewidth}
\textbf{Prompt 7 (MI: 0.004, Acc: 0.496):}
\fbox{ \parbox{\textwidth}{ \fontsize{\figurefont}{\figurefont}\selectfont
I am going to answer true or false questions about whether a word that appears in two sentences has the same meaning or not.\\\\True or False, the word "\hlx{didacticism}" has the same meaning in the following sentences.\\\\Sentence 1: \hlx{The didacticism of the 19th century gave birth to many great museums.}\\Sentence 2: \hlx{The didacticism expected in books for the young.}\\Answer:\hly{ }
}}\end{minipage} \parbox{\textwidth}{\textbf{Collapsing token sets:} \{'True': ['true'], \newline 'False': ['false']\}\newline}
\begin{minipage}{.95\linewidth}
\textbf{Prompt 8 (MI: 0.006, Acc: 0.494):}
\fbox{ \parbox{\textwidth}{ \fontsize{\figurefont}{\figurefont}\selectfont
Classify whether the following two sentences' use of the word has the same meaning or not.\\\\Word: bright\\Usage 1: He is a bright child\\Usage 2: The sun is very bright today\\Meaning: different\\\\Word: air\\Usage 1: Utah has too much air pollution.\\Usage 2: Open a window and let in some air.\\Meaning: same\\\\Word: cool\\Usage 1: Her pants are cool.\\Usage 2: Let your food cool.\\Meaning: different\\\\Word: \hlx{didacticism}\\Usage 1: \hlx{The didacticism of the 19th century gave birth to many great museums.}\\Usage 2: \hlx{The didacticism expected in books for the young.}\\Meaning:\hly{ }
}}\end{minipage} \parbox{\textwidth}{\textbf{Collapsing token sets:} \{'True': ['same'], \newline 'False': ['different']\}\newline}
\begin{minipage}{.95\linewidth}
\textbf{Prompt 9 (MI: 0.007, Acc: 0.494):}
\fbox{ \parbox{\textwidth}{ \fontsize{\figurefont}{\figurefont}\selectfont
Q: What does 2 + 2 equal?\\A: 4\\\\Q: If you are 60 inches tall how tall are you in feet?\\A: 5 feet\\\\Q: Does the word "\hlx{didacticism}" have the same meaning in the following sentences? "\hlx{The didacticism of the 19th century gave birth to many great museums.}"; "\hlx{The didacticism expected in books for the young.}"\\A:\hly{ }
}}\end{minipage} \parbox{\textwidth}{\textbf{Collapsing token sets:} \{'True': ['yes'], \newline 'False': ['no']\}\newline}
\begin{minipage}{.95\linewidth}
\textbf{Prompt 10 (MI: 0.004, Acc: 0.494):}
\fbox{ \parbox{\textwidth}{ \fontsize{\figurefont}{\figurefont}\selectfont
True or False, the word "\hlx{didacticism}" has the same meaning in the following sentences.\\\\Sentence 1: "\hlx{The didacticism of the 19th century gave birth to many great museums.}"\\Sentence 2: "\hlx{The didacticism expected in books for the young.}"\\\\Answer:\hly{ }
}}\end{minipage} \parbox{\textwidth}{\textbf{Collapsing token sets:} \{'True': ['true'], \newline 'False': ['false']\}\newline}
\begin{minipage}{.95\linewidth}
\textbf{Prompt 11 (MI: 0.009, Acc: 0.494):}
\fbox{ \parbox{\textwidth}{ \fontsize{\figurefont}{\figurefont}\selectfont
In the sentences "\hlx{The didacticism of the 19th century gave birth to many great museums.}" and "\hlx{The didacticism expected in books for the young.}", true or false, the statement "the word \hlx{didacticism} has the same meaning" is\hly{ }
}}\end{minipage} \parbox{\textwidth}{\textbf{Collapsing token sets:} \{'True': ['true'], \newline 'False': ['false']\}\newline}
\begin{minipage}{.95\linewidth}
\textbf{Prompt 12 (MI: 0.008, Acc: 0.492):}
\fbox{ \parbox{\textwidth}{ \fontsize{\figurefont}{\figurefont}\selectfont
Q: What year did America first land on the moon?\\A: 1969\\\\Q: What is the average height in America?\\A: 5 feet 9 inches\\\\Q: Does the word "\hlx{didacticism}" have the same meaning in the following sentences? "\hlx{The didacticism of the 19th century gave birth to many great museums.}"; "\hlx{The didacticism expected in books for the young.}"\\A:\hly{ }
}}\end{minipage} \parbox{\textwidth}{\textbf{Collapsing token sets:} \{'True': ['yes'], \newline 'False': ['no']\}\newline}
\begin{minipage}{.95\linewidth}
\textbf{Prompt 13 (MI: 0.017, Acc: 0.492):}
\fbox{ \parbox{\textwidth}{ \fontsize{\figurefont}{\figurefont}\selectfont
"\hlx{The didacticism of the 19th century gave birth to many great museums.}"\\"\hlx{The didacticism expected in books for the young.}"\\\\"True" or "False", the word \hlx{didacticism} has the same meaning.\\Answer: "\hly{ }
}}\end{minipage} \parbox{\textwidth}{\textbf{Collapsing token sets:} \{'True': ['true'], \newline 'False': ['false']\}\newline}
\begin{minipage}{.95\linewidth}
\textbf{Prompt 14 (MI: 0.017, Acc: 0.488):}
\fbox{ \parbox{\textwidth}{ \fontsize{\figurefont}{\figurefont}\selectfont
\hlx{The didacticism of the 19th century gave birth to many great museums.} // \hlx{The didacticism expected in books for the young.}\\Choose "yes" or "no". Does the word \hlx{didacticism} have the same meaning in the previous sentences? "\hly{ }
}}\end{minipage} \parbox{\textwidth}{\textbf{Collapsing token sets:} \{'True': ['yes'], \newline 'False': ['no']\}\newline}
\begin{minipage}{.95\linewidth}
\textbf{Prompt 15 (MI: 0.008, Acc: 0.488):}
\fbox{ \parbox{\textwidth}{ \fontsize{\figurefont}{\figurefont}\selectfont
In the sentences "\hlx{The didacticism of the 19th century gave birth to many great museums.}" and "\hlx{The didacticism expected in books for the young.}" and choosing "true" or "false", the statement "the word \hlx{didacticism} has the same meaning" is "\hly{ }
}}\end{minipage} \parbox{\textwidth}{\textbf{Collapsing token sets:} \{'True': ['true'], \newline 'False': ['false']\}\newline}
\begin{minipage}{.95\linewidth}
\textbf{Prompt 16 (MI: 0.031, Acc: 0.486):}
\fbox{ \parbox{\textwidth}{ \fontsize{\figurefont}{\figurefont}\selectfont
In linguistics, a word sense is one of the meanings of a word. Words are in two sets: a large set with multiple meanings (word senses) and a small set with only one meaning (word sense). For example, a dictionary may have over 50 different senses of the word "play", each of these having a different meaning based on the context of the word's usage in a sentence, as follows:\\\\"We went to see the play Romeo and Juliet at the theater."\\"The coach devised a great play that put the visiting team on the defensive."\\"The children went out to play in the park."\\In each sentence we associate a different meaning of the word "play" based on hints the rest of the sentence gives us.\\\\People and computers, as they read words, must use a process called word-sense disambiguation[1][2] to find the correct meaning of a word. This process uses context to narrow the possible senses down to the probable ones. The context includes such things as the ideas conveyed by adjacent words and nearby phrases, the known or probable purpose and register of the conversation or document, and the orientation (time and place) implied or expressed. The disambiguation is thus context-sensitive.\\\\Advanced semantic analysis has resulted in a sub-distinction. A word sense corresponds either neatly to a seme (the smallest possible unit of meaning) or a sememe (larger unit of meaning), and polysemy of a word of phrase is the property of having multiple semes or sememes and thus multiple senses.\\\\The following are examples of two sentences where the meaning of the word is either the same or different.\\\\Examples:\\There's a lot of trash on the bed of the river // I keep a glass of water on my bed when I sleep // bed // Different\\Air pollution // Open a window and let in some air // air // Same\\\hlx{The didacticism of the 19th century gave birth to many great museums.} // \hlx{The didacticism expected in books for the young.} // \hlx{didacticism} //\hly{ }
}}\end{minipage} \parbox{\textwidth}{\textbf{Collapsing token sets:} \{'True': ['same'], \newline 'False': ['different']\}\newline}
\begin{minipage}{.95\linewidth}
\textbf{Prompt 17 (MI: 0.007, Acc: 0.466):}
\fbox{ \parbox{\textwidth}{ \fontsize{\figurefont}{\figurefont}\selectfont
Classify whether the following two sentences' use of the word has the same meaning or not.\\\\Word: bright\\Usage 1: He is a bright child\\Usage 2: The sun is very bright today\\Meaning: different\\\\Word: air\\Usage 1: Utah has too much air pollution.\\Usage 2: Open a window and let in some air.\\Meaning: same\\\\Word: cool\\Usage 1: Her pants are cool.\\Usage 2: Let your food cool.\\Meaning: different\\\\Word: fight\\Usage 1: My wife and I had a fight.\\Usage 2: I fight for my freedom.\\Meaning: same\\\\Word: \hlx{didacticism}\\Usage 1: \hlx{The didacticism of the 19th century gave birth to many great museums.}\\Usage 2: \hlx{The didacticism expected in books for the young.}\\Meaning:\hly{ }
}}\end{minipage} \parbox{\textwidth}{\textbf{Collapsing token sets:} \{'True': ['same'], \newline 'False': ['different']\}\newline}
\begin{minipage}{.95\linewidth}
\textbf{Prompt 18 (MI: 0.010, Acc: 0.460):}
\fbox{ \parbox{\textwidth}{ \fontsize{\figurefont}{\figurefont}\selectfont
Classify whether the following two sentences' use of the word has the same meaning or not.\\\\Word: bright\\Usage 1: He is a bright child\\Usage 2: The sun is very bright today\\Meaning: different\\\\Word: air\\Usage 1: Utah has too much air pollution.\\Usage 2: Open a window and let in some air.\\Meaning: same\\\\Word: \hlx{didacticism}\\Usage 1: \hlx{The didacticism of the 19th century gave birth to many great museums.}\\Usage 2: \hlx{The didacticism expected in books for the young.}\\Meaning:\hly{ }
}}\end{minipage} \parbox{\textwidth}{\textbf{Collapsing token sets:} \{'True': ['same'], \newline 'False': ['different']\}\newline}
\begin{minipage}{.95\linewidth}
\textbf{Prompt 19 (MI: 0.007, Acc: 0.460):}
\fbox{ \parbox{\textwidth}{ \fontsize{\figurefont}{\figurefont}\selectfont
Q: Is the United States in South America?\\A: No\\\\Q: Does the word "\hlx{didacticism}" have the same meaning in the following sentences? "\hlx{The didacticism of the 19th century gave birth to many great museums.}"; "\hlx{The didacticism expected in books for the young.}"\\A:\hly{ }
}}\end{minipage} \parbox{\textwidth}{\textbf{Collapsing token sets:} \{'True': ['yes'], \newline 'False': ['no']\}\newline}
\begin{minipage}{.95\linewidth}
\textbf{Prompt 20 (MI: 0.004, Acc: 0.440):}
\fbox{ \parbox{\textwidth}{ \fontsize{\figurefont}{\figurefont}\selectfont
Q: Is the United States in South America?\\A: No\\\\Q: Is the following sentence missing a comma? Before leaving I ate breakfast.\\A: Yes\\\\Q: Does the word "\hlx{didacticism}" have the same meaning in the following sentences? "\hlx{The didacticism of the 19th century gave birth to many great museums.}"; "\hlx{The didacticism expected in books for the young.}"\\A:\hly{ }
}}\end{minipage} \parbox{\textwidth}{\textbf{Collapsing token sets:} \{'True': ['yes'], \newline 'False': ['no']\}\newline}
\\ \\

\end{document}